\pdfoutput=1
\documentclass[11pt]{article}

\usepackage[preprint]{acl}

\usepackage{times}
\usepackage{latexsym}

\usepackage{booktabs}
\usepackage{longtable}
\usepackage{tabularx}
\newcolumntype{L}{>{\raggedright\arraybackslash}X}

\usepackage{pdflscape}   % only if you want landscape for properties
\definecolor{coralpink}{rgb}{0.97, 0.51, 0.47}
\definecolor{babyblueeyes}{rgb}{0.63, 0.79, 0.95}

\newcommand{\dataset}{CulturalGround\xspace}
\newcommand{\model}{CulturalPangea\xspace}

\usepackage[T1]{fontenc}
\usepackage[utf8]{inputenc}

\usepackage{microtype}
\usepackage{xspace}
\usepackage{inconsolata}

\usepackage{graphicx}
\usepackage{multirow}
\usepackage{array}
\usepackage{amsmath}
\usepackage{booktabs}

\usepackage{color-edits}
\addauthor{jn}{blue}
\addauthor{ys}{pink}
\addauthor{gn}{magenta}

\usepackage{xcolor}
\usepackage{tcolorbox}
\tcbuselibrary{breakable}
\tcbuselibrary{skins}

\usepackage{cuted}
\usepackage{capt-of} 

\usepackage{xcolor}
\usepackage{colortbl}

\usepackage{caption}
\usepackage{subfig}
\usepackage{subcaption}        % <— must come after caption

\usepackage{adjustbox,booktabs,graphicx,array}
\renewcommand{\arraystretch}{1}

\usepackage{CJKutf8}
\newcommand{\jp}[1]{{\begin{CJK}{UTF8}{min}#1\end{CJK}}}

\title{
  Grounding Multilingual Multimodal LLMs With Cultural Knowledge
}

\author{
  Jean de Dieu Nyandwi \quad Yueqi Song \quad Simran Khanuja \quad Graham Neubig \\
    \texttt{\{jeandedi, yueqis, skhanuja, gneubig\}@andrew.cmu.edu} \\
  Carnegie Mellon University \\ \\
  \url{https://neulab.github.io/CulturalGround/}
}

\begin{document}
\maketitle

\begin{abstract}
Multimodal Large Language Models excel in high-resource settings, but often misinterpret long-tail cultural entities and underperform in low-resource languages. To address this gap, we propose a data-centric approach that directly grounds MLLMs in cultural knowledge. Leveraging a large scale knowledge graph from Wikidata, we collect images that represent culturally significant entities, and generate synthetic multilingual visual question answering data. The resulting dataset, \dataset, comprises 22 million high-quality, culturally-rich VQA pairs spanning 42 countries and 39 languages. We train an open-source MLLM \model on \dataset, interleaving standard multilingual instruction-tuning data to preserve general abilities. \model achieves state-of-the-art performance among open models on various culture-focused multilingual multimodal benchmarks, outperforming prior models by an average of \textbf{+5.0\%} without degrading results on mainstream vision–language tasks. Our findings show that our targeted, culturally grounded approach could substantially narrow the cultural gap in MLLMs and offer a practical path towards globally inclusive multimodal systems.
\end{abstract}

% \vspace{0.5cm} % Small spacing

\begin{figure*}[t!]
\centering
\includegraphics[width=\textwidth]{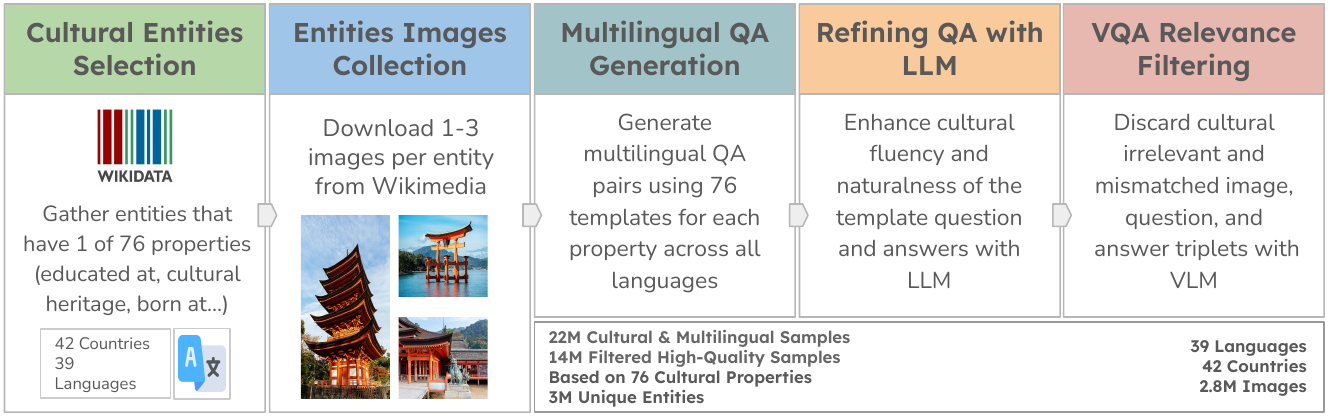}
\caption{Our data curation pipeline involves gathering culturally relevant entities from the Wikidata knowledge base, creating several questions and answers about each entity, rephrasing them using an LLM, and filtering low-quality samples using a VLM.}
\label{fig:data_pipeline}
\end{figure*}

\section{Introduction}

Despite being trained on billions of image–text pairs, today's multimodal large language models (MLLMs) remain biased towards English and Western-centric training data~\citep{ramaswamy2023geode,vayani2024alm,yue2024pangea,liu2025culturevlm}. 
As a result, MLLMs often excel on high-resource languages, but even state-of-the-art models overlook or misinterpret non-Western cultural cues, especially long-tail entities~\citep{liu2021visually,blasi-etal-2022-systematic,ahia-etal-2023-languages,alkhamissi-etal-2024-investigating,romero2024cvqa,ananthram2024see}.
Simply translating English data or increasing the size of the training corpora does not solve this problem. Translated data remain ``Anglo-centric'', and naively scaling up training corpora will not change the underlying biased distribution of the data~\citep{yu-etal-2022-beyond,tao2024cultural,gallegos2024bias}. 

Recent efforts emphasize the need for targeted, multicultural data curation to bridge this gap~\citep{yue2024pangea,liu2025culturevlm}. MLLMs can only learn the knowledge they perceive: imbuing models with cultural understanding thus requires training data that explicitly incorporates diverse cultural entities, images, and linguistic contexts~\citep{hershcovich-etal-2022-challenges,li2024culturellm,cahyawijaya2025crowdsource}.
However, for long-tail entities, these data remain few and far between.

In this paper, we propose a novel approach to ground multilingual MLLMs with cultural knowledge from large-scale knowledge bases. 
We introduce a scalable pipeline for constructing culturally grounded multilingual multimodal data, curating rich visio-linguistic training data centered on regional cultural entities.

Our pipeline, as shown in \autoref{fig:data_pipeline}, consists of the following steps: 
\textbf{1) Cultural Concept Selection:} we first select culturally significant concepts from the large-scale knowledge resource Wikidata\footnote{\url{https://www.wikidata.org/}}, leveraging its structured, multilingual knowledge graph~\citep{vrandevcic2014wikidata}; 
\textbf{2) Image Collection:} for each selected entity, we retrieve images from both Wikidata and Wikimedia Commons\footnote{\url{https://commons.wikimedia.org/}}, increasing coverage and diversity beyond any single source; 
\textbf{3) Multilingual Factual VQA Generation:} utilizing a set of structured, language-specific templates, we generate multilingual factual Visual Question Answering (VQA) data based on each entity’s Wikidata properties (e.g., \textit{occupation}, \textit{religion}) in 39 languages;
\textbf{4) VQA Refinement:} We refine these template-based VQAs with an LLM, prompting it to improve fluency, contextual richness, and cultural naturalness, without leaking the entity identity in the question;
\textbf{5) Filtering:} To guarantee accurate cultural knowledge grounding, we use MLLMs to discard mismatched or culturally irrelevant VQA instances.
Our pipeline results in \dataset, a high-quality multilingual multimodal dataset comprising 22 million VQA examples, explicitly curated to reflect diversity in cultural entities.

To evaluate the effectiveness of \dataset in imbuing models with cultural knowledge, we conducted experiments to train a multilingual MLLM \model on \dataset. In experiments, \model outperforms previous open-source MLLMs on multiple culture-focused vision–language benchmarks on average of \textbf{+5.0} across all benchmarks. On evaluation sets such as CVQA~\citep{romero2024cvqa} and ALM-Bench~\citep{vayani2024alm}, which specifically probe cross-cultural and multilingual understanding, \model achieves state-of-the-art results among open models. Crucially, these gains in cultural competency do not come at the expense of general capability: our model retains competitive performance on standard vision–language tasks. These findings demonstrate that a data-centric strategy, carefully curating multilingual, culturally rich examples can substantially bridge the cultural gap in MLLMs.
\section{Dataset}

\begin{figure*}[]
  \centering
  \begin{minipage}[c]{1.0\linewidth}
    \includegraphics[width=\linewidth]{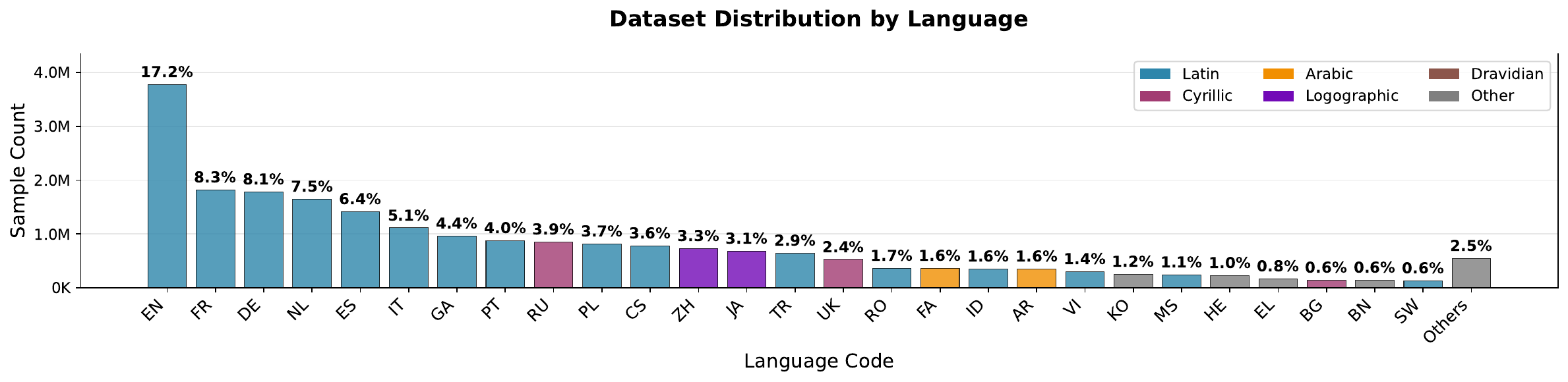}
  \end{minipage}%
  \caption{The distribution of samples across languages highlighting percentages of each language in \dataset.}
  \label{fig:language_dist_plot}
\end{figure*}

In this section, we describe our proposed data generation pipeline from \autoref{fig:data_pipeline} in detail.

\subsection{Problem Definition}
Our goal is to construct a culturally grounded multimodal dataset suitable for training multilingual vision-language models. The core challenge is ensuring both cultural relevance and factual accuracy across diverse regions and languages.

\textbf{Inputs.} Our construction process takes four primary inputs:
\begin{itemize}
\item A structured knowledge base \(\mathcal{K}\) (specifically Wikidata) containing entities \(E\) and their factual relationships
\item A set of regions for which we would like to create data \(R = \{r_1, \ldots, r_m\}\), each identified by unique IDs within \(\mathcal{K}\)
\item A collection of target languages \(L = \{l_1, \ldots, l_k\}\) representing the linguistic communities we would like to cover.
\item A set of culturally meaningful properties \(P = \{p_1, \ldots, p_n\}\) (e.g., \emph{place of birth}, \emph{occupation}) that indicate cultural relevance.
\end{itemize}

\textbf{Output.} We produce a dataset \(D\) consisting of triplets \((i_e, q_{e,p}^{(l)}, a_{e,p}^{(l)})\) where each triplet contains:
\begin{itemize}
\item \(i_e\): An image depicting entity \(e\)
\item \(q_{e,p}^{(l)}\): A natural language question about property \(p\) of entity \(e\) in language \(l\)
\item \(a_{e,p}^{(l)}\): The corresponding factual answer in the same language \(l\)
\end{itemize}

\subsection{Cultural Entity Selection}

The first step toward constructing \(D\) is to identify a culturally representative subset of entities \(E' \subseteq E\) from the knowledge base \(\mathcal{K}\).
For every target region \(r \in R\) we map \(r\) to its unique Wikidata identifier (QID).\footnote{A QID is a unique identifier beginning with ``Q’’ followed by digits, e.g.\ \url{https://www.wikidata.org/wiki/Q43649390}.}
% We then keep every entity \(e\) that is connected to \emph{any} \(r\) by at least one culturally meaningful property \(p \in P\):
% \[
% \small
% E' \;=\; \bigl\{\,e\in E \mid \exists\,r\in R,\,p\in P \text{ with } (e,p,r)\in\mathcal{K}\bigr\}
% \]
\[
% \small
\begin{aligned}
E' \;=\; \bigl\{\, e \in E \mid\; & \exists\, r \in R,\, p \in P \text{ with } \\
& (e,p,r) \in \mathcal{K} \,\bigr\}
\end{aligned}
\]

To guarantee multilingual coverage, we further require each \(e\in E'\) to have a label or description in at least one language \(l\in L\).
This intersectional filtering step results in a culturally rich and linguistically diverse set of entities spanning people, places, foods, artifacts, and institutions (full statistics can be seen in Figures \ref{fig:culturalground_samples}, \ref{fig:entity_categories}, and \ref{fig:language_diversity_distribution} in the appendix). 

\subsection{Entity Images Collection}

For every \(e\in E'\) we assemble a set of images
\(
\mathcal{I}_e = \{i_{e,1},\dots,i_{e,t_e}\}
\)
that will later appear in triplets \((i_e,q_{e,p}^{(l)},a_{e,p}^{(l)})\).
If \(\mathcal{K}\) links an image via property {\tt P18}, we include that as \(i_{e,1}\).

To broaden visual coverage, we also obtain additional images from the Wikimedia Commons category page when available \citep{srinivasan2021wit}. Commons categories often contain dozens of photographs or illustrations related to a given entity (for example, a notable person’s category might include portraits from different events, while a landmark’s category might contain images from various angles or time periods). This step enriches \model’s training data with visual variability: the same cultural entity might appear in different settings or styles across its image set.

\subsection{Multilingual Question-Answer Generation}
\label{subsec:multilingual_qa}

We generate factual QA pairs across languages by instantiating language-specific templates with knowledge-base facts. For every culturally selected entity \(e \in E'\), property \(p \in P\)\footnote{To avoid over-representation of popular entities with many Wikidata properties, we cap the number of properties used per entity at the country-specific median.} with value \(v\) such that \((e,p,v)\in\mathcal{K}\), and target language \(l \in L\) for which \(e\) is labelled in \(l\), we create a question–answer pair \((q_{e,p}^{(l)}, a_{e,p}^{(l)})\).

\paragraph{Property-level QA.} Let \(T_p^{(l)}\) be the question template for property \(p\) in language \(l\), and let \(\mathrm{name}_l(\cdot)\) and \(\mathrm{desc}_l(\cdot)\) denote the language-\(l\) label and (optional) short description returned by \(\mathcal{K}\). We instantiate:
\[q_{e,p}^{(l)} = T_p^{(l)}\left(\text{relevant reference to } e\right)\]
\[a_{e,p}^{(l)} = A_p^{(l)}(\mathrm{name}_l(e), v)\]

where \(A_p^{(l)}\) is the answer template for property \(p\) in language \(l\) and \(v\) is the property's value.

\emph{Example (English):} \(p=\) \emph{place of birth} \(\Rightarrow\) \(q\): "Where was this person born?"; \(a\): "\{entity\_name\} was born in \{property\_value\}." \(\rightarrow\) "Albert Einstein was born in Ulm, Germany."

\paragraph{Entity-level QA.} To capture identity and brief context, we include one entity-level pair per \(e\) using template \(T_{\mathrm{id}}^{(l)}\):

\[q_{e,\mathrm{id}}^{(l)} = T_{\mathrm{id}}^{(l)}\left(\text{relevant reference}\right)\]

\[a_{e,\mathrm{id}}^{(l)} = A_{\mathrm{id}}^{(l)}\left(\mathrm{name}_l(e), \mathrm{desc}_l(e)\right)\].

\emph{Example (English):} \(q\): "What is the entity shown in the image?"; \(a\): "\{entity\_name\}, \{entity\_description\}." \(\rightarrow\) "The Taj Mahal, a 17th-century mausoleum in India."

Each QA is initially generated in ``fill-in-the-blank'' style~\citep{jiang2020x} by inserting the entity's Wikidata facts into the QA templates. Together, the property-level questions and the general entity-level questions form a diverse initial factual QA set for each entity, covering both specific facts and broader contextual information.
We then match the corresponding images of the entity to the questions generated for this entity, creating multiple candidate (image, question, answer) triplets \((i_{e,j}, q_{e,p}^{(l)}, a_{e,p}^{(l)})\) per entity. 
These templated pairs ensure wide coverage of factual attributes, but often lack linguistic variation and contextual nuance, motivating a subsequent refinement step.

\subsection{Refining Multilingual VQA Data for Cultural Fluency}

While the template-based QA generation ensures factual correctness, the resulting text is often formulaic or grammatically awkward, lacking the rich context of a human-written question or answer. We therefore employ a large language model (LLM) to rewrite and polish each question–answer pair \(\bigl(q^{(l)}_{e,p}, a^{(l)}_{e,p}\bigr)\) for greater fluency and cultural naturalness.
\[
(q'^{(l)}_{e,p},\, a'^{(l)}_{e,p})
\;=\;
\mathsf{Refine}^{(l)}\!\bigl(q^{(l)}_{e,p},\, a^{(l)}_{e,p}\bigr).
\]

In particular, we prompt strong open-source models (such as Qwen2.5-72B~\citep{qwen2025qwen25technicalreport} and Gemma3-27B~\citep{team2025gemma}) with the templated pair and specific instructions on how to improve it. The LLM is instructed to avoid any direct mention of the entity’s name or identity in the question (ensuring that the refined question \(q'^{(l)}_{e,p}\) remains answerable only via the image, with no textual leakage of the entity name) and to incorporate subtle contextual cues. For example, the model might say “in your country” or “this individual in India” rather than explicitly naming a person, or describe a shrine by its religion and notable features instead of giving its proper name.

We further direct the LLM to enrich each refined answer \(a'^{(l)}_{e,p}\) with culturally relevant details while preserving the core factual content, often by adding a brief background or significance to the fact. For instance, a short answer like “He is the Prime Minister of India.” can be expanded to “He is the 14th Prime Minister of India, serving as the country’s highest political leader.” Likewise, a generic question such as “Which public office does this person hold?” might be rewritten as “What is the highest political office that this individual currently holds in India?”, adding clarity and specificity.

This step significantly improved the coherence and readability of the QA pairs, confirming that the refinement yields questions and answers that are more natural and informative yet still anchored in the original cultural context (see \autoref{fig:culturalground_samples}) for examples).

\subsection{Image-Text Relevance Filtering}
\label{subsec:image_text_filtering}

After assembling the images with the rewritten QA pairs, we apply a final filtering step to ensure that each image is \emph{meaningfully} relevant to its paired question and answer. Not every image retrieved from Wikimedia Commons precisely depicts the intended entity, and some (re)written QAs may not align with specific images (especially when multiple images exist per entity).

Concretely, for a candidate triplet \((i, q'^{(l)}, a'^{(l)})\) in language \(l\), the VLM issues a binary alignment judgment
\[
y\bigl(i, q'^{(l)}, a'^{(l)}\bigr) \in \{\texttt{true}, \texttt{false}\},
\]
indicating whether the image both (i) \emph{visually grounds} the entity/scene implicated by the QA and (ii) \emph{matches the cultural context} described in \(q'^{(l)}\) and \(a'^{(l)}\) (e.g., region, heritage, religious or architectural attributes). We retain the triplet if \(y(\cdot)=\texttt{true}\); otherwise it is discarded. We use Qwen-2.5VL(32B, 72B)-Instruct and Gemma-3(12B, 27B)-IT models for filtering, selecting models based on their strength for specific regions and languages.

This selective filtering ensures that \textbf{\model}'s training data maintains a high level of image--text relevance, improving supervision quality. As a result, the final dataset primarily contains triplets where the image truly depicts the entity in question or closely relates to the QA content, which is crucial for grounding the model’s understanding in visual evidence.

\subsection{Dataset Statistics and Languages}

Following the above curation steps, we compiled a large multilingual multimodal dataset of QA-image pairs. The resulting dataset comprises about 22M million high-quality image-question-answer triplets spanning 39 languages and 42 regions. As shown in \autoref{fig:language_dist_plot}, \autoref{fig:countries-langs}, \autoref{fig:language_diversity_distribution}, and \autoref{fig:entity_connectivity_distribution}, the dataset statistics demonstrate our focus on cultural diversity and long-tail entities.
% \begin{table}[t]
%   \centering
%     \small
%   \begin{tabular}{lc}
%     \toprule
%     % \textbf{Stages} & \textbf{Finetuning} \\
%     % \midrule
%     \multicolumn{2}{c}{\textbf{Training Data}} \\
%     \midrule
%     \dataset(OE) & 13M sampled from 22M \\
%     \dataset(MCQs) & 5M sampled from 8M \\
%     % \dataset-Entities & 200K \\
%     % Pangeans & English, 1.6M \\
%     % Pangeans & Multilingual, 1M  \\
%     % M3LS  &  100K \\
%     \midrule
%     \multicolumn{2}{c}{\textbf{Model}} \\
%     \midrule
%     Vision Encoder & CLIP-ViT-14 \\
%     LLM & Qwen2-Instruct \\
%     Base Model & Pangea-7B \\
%     Trainable Parts & Connector, LLM \\
%     \midrule
%     \multicolumn{2}{c}{\textbf{Training}} \\
%     \midrule
%     Batch Size & 128 \\
%     LR: $\{\theta_{\text{proj}},\theta_{\text{LLM}}\}$ & $5 \times 10^{-6}$ \\
%     Epoch & 1 \\
%     GPU Hours (H100) & 189 \\
%     \bottomrule
%   \end{tabular}
%   \caption{\model's training configurations. \gncomment{In general, it's nice to put the tables on the top of the page not ``here'' because it doesn't break up the text.}}
%   \label{tab:training_config}
% \end{table}

\begin{table*}[t]
  \centering
  \small
  \begin{tabular}{lc@{\hspace{2em}}lc@{\hspace{2em}}lc}
    \toprule
    \multicolumn{2}{c@{\hspace{2em}}}{\textbf{Training Data}} & \multicolumn{2}{c@{\hspace{2em}}}{\textbf{Model}} & \multicolumn{2}{c}{\textbf{Training}} \\
    \midrule
    Total \dataset(OE) & 22M & Vision Encoder & CLIP-ViT-14 & Batch Size & 128 \\
    Sampled  & 13M & LLM & Qwen2-Instruct & Learning Rate & $5 \times 10^{-6}$ \\
    Total \dataset(MCQs) & 8M & Base Model & Pangea-7B & Epoch & 1 \\
    Sampled  & 5M & Trainable Parts & Connector, LLM & GPU Hours (H100) & 720 \\
    \bottomrule
  \end{tabular}
  \caption{\model's training configurations.}
  \label{tab:training_config}
\end{table*}

\section{Experimental Setup}
\subsection{Training}
As a base model, we start from Pangea-7B \citep{yue2024pangea}, which couples a CLIP-based vision encoder~\citep{radford2021learning} with a Qwen2-7B autoregressive language model~\citep{bai2025qwen2}. We keep the vision encoder frozen and fine-tune only the connector and the language model on our culturally grounded data. Key training details include:
\begin{itemize}
    \item \textbf{Dataset}: We train on 13M open-ended VQA pairs sampled from \dataset 21M total samples, covering 39 culturally diverse languages and cultures. To improve model robustness, we also create 8M multiple-choices VQA samples grounded on collected cultural entities, and sample 5M to use in training. Multiple-choices question pipeline is expanded in Appendix \ref{app:mcqs_pipeline} and sampling details can be found in Appendix \ref{app:sampling}.
    \item \textbf{Multilingual mixture}: To preserve the base model’s broad multilingual grounding and avoid catastrophic forgetting, we interleave 5.8M samples of PangeaInstruct English and multilingual data with the \model samples during fine-tuning. We also include a filtered 90K samples of M3LS~\citep{verma2023large} to improve entity recognition. The details for M3LS entity linking data is further discussed in Appendix \ref{app:m3ls}.
    \item \textbf{Optimization}: We train with a small learning rate of 5e-6. This choice is motivated by recent observations that large vision–language models benefit from lower learning rates for cross-lingual transfer~\citep{steiner2024paligemma}. We train with a cosine decay schedule and a brief warm-up, tuning all connector and LLM parameters. Further training settings are highlighted in \autoref{tab:training_config}. We also experiment with checkpoints merging and we discuss this in Section \ref{section:merging}.
\end{itemize}

Together, these settings allow for \model to better retain its pretrained abilities while adapting to the culturally enriched data.

\subsection{Evaluation Protocol}

We evaluate \model's multilingual and cultural competence on benchmarks from PangeaBench and related tasks. In particular, we use the following multimodal benchmarks: CVQA~\citep{romero2024cvqa}, MaRVL~\citep{liu2021marvl}, XM100~\citep{thapliyal2022crossmodal}, ALM-Bench~\citep{vayani2024alllanguages}, MERLIN~\citep{merlin2024}, MaXM~\citep{changpinyo2024pangeabench}, M3Exam~\citep{zhang2023m3exam}. \autoref{tab:cultural_datasets} shows an overview of the benchmarks we evaluate on.

For baselines, we compare our model against other recent state of the art multimodal LLMs such as Llava-Next-7B~\citep{liu2024llavanext}, Molmo-7B-D~\citep{deitke2024molmo}, Llama3.2-11B~\citep{grattafiori2024llama}, and mBLIP~\citep{geigle2023mblip}, PaliGemma-3B\citep{beyer2024paligemma}, AyaVision-8B~\citep{dash2025aya}, and Pangea~\citep{yue2024pangea}.

\section{Results and Analysis}

\begin{table*}[t!]
  \centering
  \small
  \setlength{\tabcolsep}{4pt}
  \begin{tabular}{lcccccccc}
    \toprule
    & \multicolumn{3}{c}{\textbf{Cultural Understanding}} 
    & \multicolumn{1}{c}{\textbf{Entity Recognition}} 
    & \multicolumn{2}{c}{\textbf{Multilingual VQA}} 
    & \multicolumn{1}{c}{\textbf{Captioning}} 
    & \multicolumn{1}{c}{\textbf{Average}} \\
    \cmidrule(lr){2-4} \cmidrule(lr){5-5} \cmidrule(lr){6-7} \cmidrule(lr){8-8} \cmidrule(lr){9-9}
    Models
      & CVQA
      & MARVL
      & ALM
      & MERLIN
      & MAXM
      & M3EXAM
      & XM100 
      & ALL \\
    \midrule
    Llava-Next-7B               & 40.9 & 50.9 & 42.4 & 34.1 & 21.4 & 28.4 & 15.5 & 33.4 \\
    Molmo-7B-D                  & 58.7 & 54.9 & 49.1 & 42.9 & 37.5 & 39.1 &  6.0 & 41.2 \\
    Llama3.2-11B                & \textbf{69.6} & 58.1 & 56.6 & 49.1 & 43.9 & 36.6 &  5.8 & 45.7 \\
    PaliGemma-3B                & 42.5 & 52.2 & 35.7 & 13.1 & 19.9 & 25.6 &  0.6 & 27.1 \\
    mBLIP-mT0-XL                & 37.5 & 66.7 & 36.9 & 15.8 & 36.8 & 25.0 &  6.8 & 32.2 \\
    AyaVision-8B                & 50.8 & 64.5 & 55.1 & 55.3 & 52.1 & 41.7 & 10.0 & 47.1 \\
    Pangea-7B                   & 56.9 & \underline{78.7} & \underline{59.9} & \underline{66.0} & \underline{53.3} & \underline{42.0} & \underline{29.7} & \underline{55.3} \\
    % \model(162K)    & 58.2 & 80.2 & 63.1 & 80.2 & 53.1 & 45.0 & 36.6 & 59.5 \\
    \model    & \underline{59.1} & \textbf{80.3} & \textbf{63.5} & \textbf{81.1} & \textbf{53.9} & \textbf{46.7} & \textbf{36.9} & \textbf{60.3} \\
    \textbf{$\Delta$ over Pangea}               &  +2.2 &  +1.6 &  +3.6 & +15.1 &  +0.6 &  +4.7 &  +7.2 &  +5.0 \\
    \bottomrule
  \end{tabular}
  \caption{Multilingual performance comparison across models on cultural understanding (CVQA, MARVL, ALM), entity recognition (Merlin), multilingual VQA (MAXM, M3Exam), and captioning (XM100). For CVQA, we evaluate on 31 country-language pairs, 38 languages in ALMBench and for other benchmarks, we use all languages available in the respective datasets. The best-performing model on each dataset is in \textbf{bold} and the second best is \underline{underlined}.}
  \label{tab:model_performance}
\end{table*}

\subsection{Cultural Understanding Benchmarks}
In~\autoref{tab:model_performance} we show performance on culturally focused multimodal benchmarks including CVQA, XM100, ALMBench, MaRVL, and MERLIN.
We can see that \model achieves state-of-the-art accuracy on nearly all of these datasets and various question types(\autoref{fig:almbench_question_types}, substantially improving over the base Pangea-7B and other competitive open models.
These results verify that culturally grounded training data significantly enhances culture-specific visual reasoning.
Overall, \textsc{\model} substantially advances the state of the art in culturally informed multimodal understanding.

\subsection{General Multilingual and English Performance}

Despite our attempts to specifically improve accuracy on culturally relevant phenomena, \model maintains strong general multimodal and multilingual capabilities. On M3Exam for example, \model ouperforms Pangea-7B by +4.7 points and matches performance on MaXM, while surpassing most open models. Furthermore, \model outperforms Pangea-7B on XM100(multilingual captioning benchmark) by a large margin.

\begin{figure}[t]
    \centering
    \includegraphics[width=1.0\linewidth]{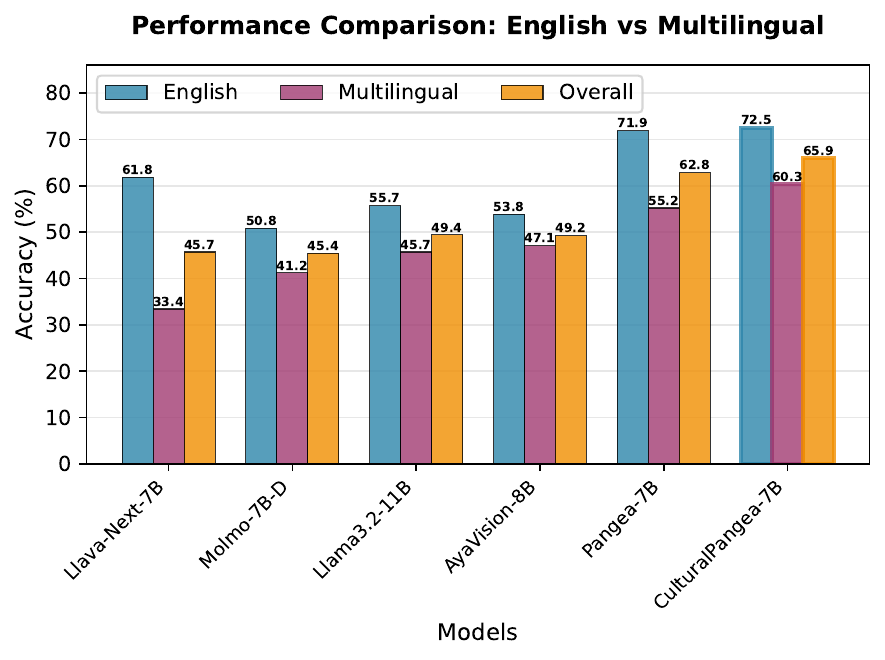}
    \caption{Overall performance comparison in english and multilingual}
    \label{fig:english_multilingual}
\end{figure}

\model maintains excellent English performance not only in cultural settings, but also in general tasks as shown in \autoref{fig:english_multilingual}. Balancing different cultures, languages, while maintaining English and general skills is challenging~\citep{chuang2025metaclip2worldwidescaling, pouget2024no} and we attribute this to our data curation pipeline that gathers entities in many regions and different languages in each region.

\subsection{Analysis and Discussion}
\begin{table*}[t]
\tiny
\centering
\begin{tabular}{@{}lrrrrrrrrrrrrrrrrrr@{}}
\toprule
\textbf{Model} & \textbf{SC} & \textbf{AS} & \textbf{EG} & \textbf{YO} & \textbf{GU} & \textbf{BH} & \textbf{LA} & \textbf{SI} & \textbf{SA} & \textbf{DA} & \textbf{GL} & \textbf{AF} & \textbf{IC} & \textbf{AZ} & \textbf{SH} & \textbf{SK} & \textbf{FI} & \textbf{Avg} \\
\midrule
Pangea-7B      & 28.3 & 40.5 & 63.6 & 21.5 & 35.1 & 49.7 & 19.2 & 37.0 & 64.4 & 59.9 & 65.3 & 58.8 & 45.3 & 51.1 & 26.2 & 38.4 & 41.7 & 45.0 \\
CulturalPangea-7B & 39.4 & 50.9 & 68.3 & 25.8 & 39.1 & 53.2 & 22.0 & 39.6 & 66.8 & 62.2 & 66.9 & 60.3 & 46.5 & 52.2 & 26.8 & 39.0 & 42.1 & 48.3 \\
\midrule
\textbf{Gain}  & \textbf{+11.1} & \textbf{+10.4} & \textbf{+4.7} & \textbf{+4.3} & \textbf{+4.0} & \textbf{+3.5} & \textbf{+2.9} & \textbf{+2.6} & \textbf{+2.4} & \textbf{+2.3} & \textbf{+1.5} & \textbf{+1.5} & \textbf{+1.3} & \textbf{+1.1} & \textbf{+0.7} & \textbf{+0.6} & \textbf{+0.4} & \textbf{+3.3} \\
\bottomrule
\end{tabular}
\caption{Cross-Cultural/Lingual Performance on ALM-Bench. Language codes: SC=Scots Gaelic, AS=Assamese, EG=Egyptian Arabic, YO=Yoruba, GU=Gujarati, BH=Bhojpuri, LA=Lao, SI=Sindhi, SA=Saudi Arabic, DA=Danish, GL=Galician, AF=Afrikaans, IC=Icelandic, AZ=Azerbaijani, SH=Shona, SK=Sanskrit, FI=Filipino.}
\label{tab:crosslingual-performance}
\end{table*}

\paragraph{Cross-Cultural and Cross-Lingual Transfer.}
\model demonstrates cross-cultural and lingual transfer on languages that are not in \dataset. To analyze transfer behaviors across cultures and languages, we compare performance with baseline on 17 languages from ALMBench. As shown in \autoref{tab:crosslingual-performance}, our model consistently shows improvements over Pangea-7B. This trend suggests that the model effectively transfers knowledge to languages with limited training data, alleviating the typical drop-off seen in low-resource settings.

% \begin{figure}[h]
%   \centering
%   \includegraphics[width=\linewidth]{graphicss/cross_lingual_performance_almbench.pdf}
%   \caption{Cross-cultural and cross-lingual accuracy comparison on ALMBench.}
%   \label{fig:cross-lingual}
% \end{figure}
We hypothesize that these cross-lingual improvements may be a result of \model’s multilingual, culturally grounded supervision strategy. During training, most entities are accompanied by QA examples in multiple languages, which encourages the model to align semantic representations across languages. This design can enable knowledge learned from one language to be readily applied to other languages. As a result, \model leverages training signals from diverse languages to significantly boost accuracy on underrepresented languages, explaining its across-the-board gains on ALMBench.

\paragraph{Cultural Data Scaling and General Skill Preservation.}

\begin{figure*}[t]
  \centering
  \includegraphics[width=\linewidth]{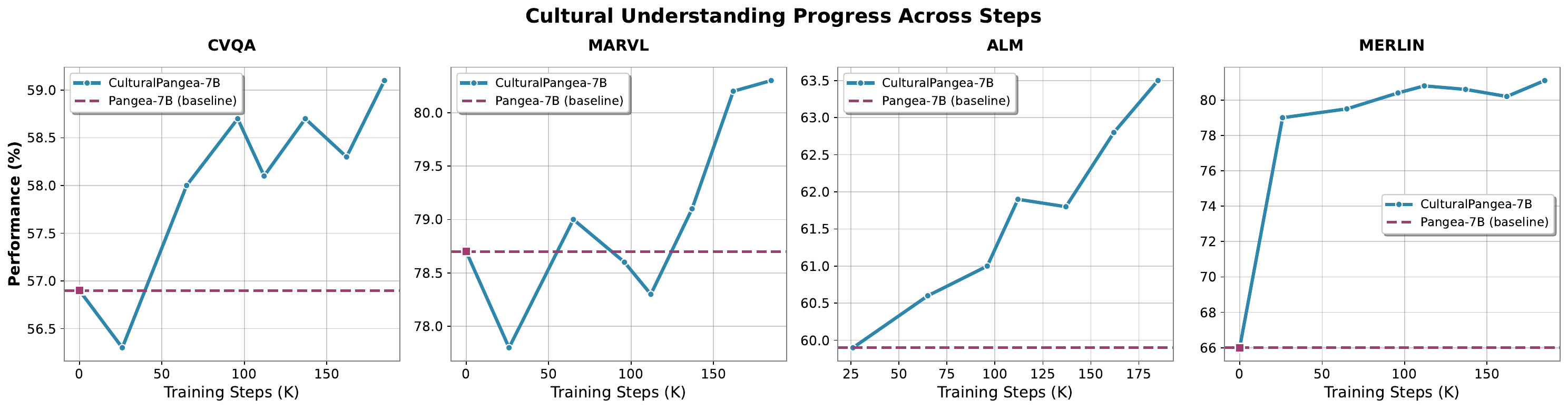}
  \caption{Training performance curves on four culture‐centric benchmarks
  (CVQA, MaRVL, ALM‐Bench, and MERLIN) show accuracy steadily rising as
  \model is trained on more culturally grounded data versus the baseline model.
  Higher training step counts—\emph{i.e.}, greater exposure to the
  \dataset consistently translate into improved accuracy.}
  \label{fig:scaling_cultural_data}
\end{figure*}

\begin{figure*}[t]
  \centering
  \includegraphics[width=\linewidth]{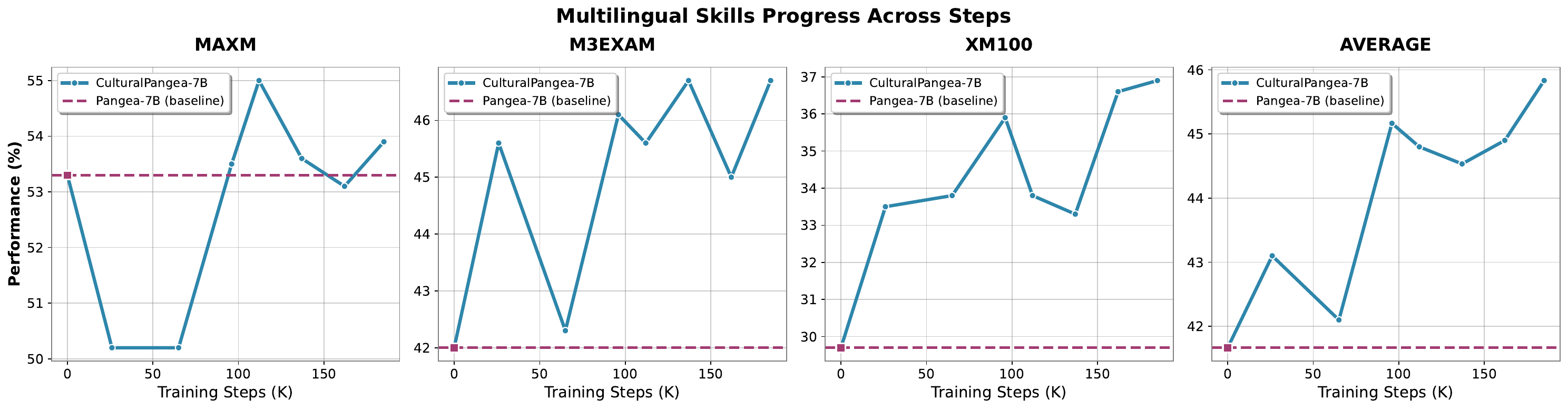}
  \caption{Training performance curves on three general multilingual benchmarks and their overall average show accuracy improvements as \model is exposed on more data, compared to the Pangea‑7B.}
  \label{fig:multilingual_skills}
\end{figure*}

As additional culturally grounded data is introduced during training, \textsc{CulturalPangea}'s performance steadily improves on all culture-sensitive benchmarks (\autoref{fig:scaling_cultural_data}), while its general multilingual vision–language proficiency is concurrently preserved and even enhanced, as shown in \autoref{fig:multilingual_skills}. This outcome indicates that our interleaved training strategy successfully avoided catastrophic forgetting by continuously mixing standard multilingual examples into the cultural fine-tuning process. In essence, this approach parallels replay-based continual learning, wherein revisiting earlier tasks helps maintain broad competence~\citep{kirkpatrick2017overcoming, rolnick2019experience}. 

Consistent with this, we observed a characteristic ``dip-and-recovery'' trajectory in the model's general performance during training: an initial drop when new cultural data was first introduced, followed by a rebound that ultimately exceeded the original baseline. Specifically, on general multilingual benchmarks like \textbf{M3Exam} and \textbf{MaXM}, performance temporarily decreased by 2-3\% in early training steps before recovering to surpass baseline scores by +4.7 and +0.6 points respectively. These dynamics mirror the ``stability gap'' phenomenon reported in incremental learning~\citep{de2022continual, caccia2021new}. 

By the end of training, \model not only acquires remarkably stronger culturally grounded capabilities (averaging +5.0\% improvement across cultural benchmarks) but also achieves slightly higher overall VQA accuracy than the baseline, exemplifying a difficult-to-achieve equilibrium between new specialization and retained generalization, and highlighting the promise of \dataset and training approach.

\begin{figure}[t]
    \centering
    \includegraphics[width=\linewidth]{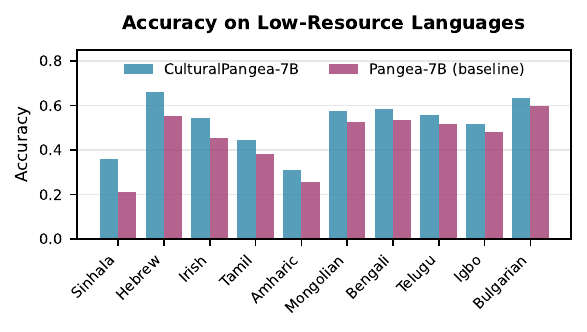}
    \caption{\model achieves the largest gains on underrepresented languages, demonstrating effective scaling to the long tail of languages.}
    \label{fig:lowres-gains}
\end{figure}

\paragraph{Does \dataset Help Low-Resource Languages?}
\model demonstrates substantial gains on ALMBench for languages with limited training data, with improvements most pronounced for resource-poor languages within \dataset. As shown in Figure~\ref{fig:lowres-gains-fullset}, we observe absolute accuracy gains of 15.0\% points on Sinhala, 10.9\% on Hebrew, and 9.1\% on Irish. Additional low-resource languages exhibit consistent improvements: Tamil (+6.3), Amharic (+5.3), Bengali (+4.8), and Telugu (+4.2). These gains occur without sacrificing performance elsewhere—nearly all languages improve, with only Norwegian (-0.5) showing negligible regressions. The largest improvements in traditionally underrepresented languages indicate that culturally-aware grounding effectively scales to the long tail of languages, enhancing multilingual inclusivity.

\begin{figure}[t]
\centering
\includegraphics[width=1.0\linewidth]{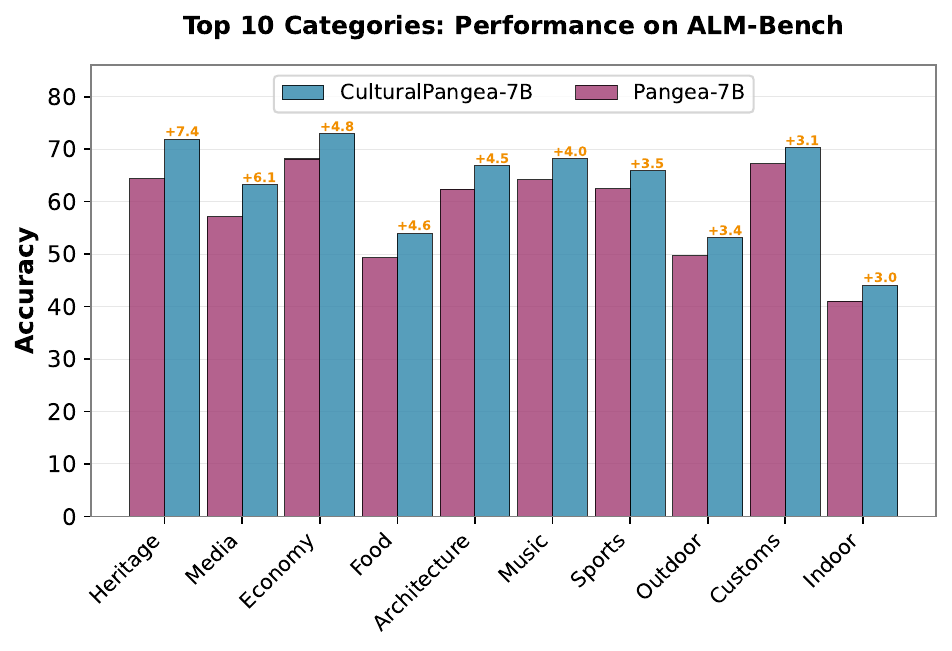}
\caption{Cultural domains in ALMBench such as heritage and food show largest gains over general domains.}
\label{fig:top_10domains}
\end{figure}

\paragraph{Which Cultural Domains Benefit Most?}
The model's gains vary substantially across cultural domains, with culturally rich categories showing the largest improvements. As shown in \autoref{fig:top_10domains} and \autoref{fig:all_domains}, \textbf{Heritage} achieves the highest relative improvement at 11.5\% (from 64.4\% to 71.8\%), followed by \textbf{Media} at 10.6\% and \textbf{Food} at 9.2\%. Other culturally salient domains like \textbf{Architecture} (7.1\%), \textbf{Economy} (7.0\%), and \textbf{Music} (6.2\%) also demonstrate substantial gains. These results indicate that domains requiring broad cultural knowledge and context benefit most from our approach.
In contrast, generic visual domains show minimal improvement or slight regression. \textbf{Sketch} decreases by 0.6\%, while \textbf{Meme} improves only marginally at 2.31\%. Similarly, \textbf{Festivals} (0.73\%) and \textbf{Religion} (2.08\%) show limited gains. This pattern confirms that improvements concentrate in genuine cultural understanding areas, while abstract visual content or highly localized traditions remain challenging. The largest accuracy gains occur in well-represented cultural domains, reinforcing the value of targeting cultural knowledge in model training.

% \begin{figure*}[t]
%   \centering
%   \includegraphics[width=\linewidth]{graphicss/full_categories_alm.pdf}
%   \caption{Absolute accuracy gains of CulturalPangea over the baseline across 18 ALM‑Bench cultural domains.  Improvements cluster in culture‑rich categories (Media, Heritage, Music), while Sketch and Meme offer minimal or negative change.}
%   \label{fig:all_domains}
% \end{figure*}

\begin{figure}[t]
    \centering
    \includegraphics[width=\linewidth]{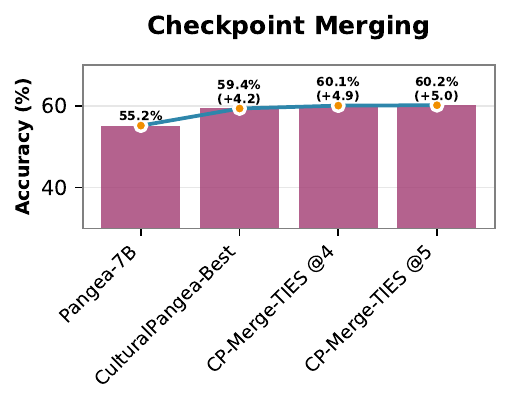}
    \caption{Accuracy improvements from merging checkpoints. CP stands for CulturalPangea.}
    \label{fig:checkpoint-avg}
\end{figure}

\paragraph{Performance Gains via Checkpoint Merging}
\label{section:merging}
Following~\citep{dash2025aya, team2025kwai, li2025model}, we merge 5 strong \textsc{CulturePangea} checkpoints from different training stages using the \textbf{TIES} method, which recovers complementary model strengths often lost during continual training. Although linear and DARE‑TIES variants show comparable results, TIES yields the highest average accuracy. As shown in \autoref{fig:checkpoint-avg}, using our strongest early checkpoint as the base outperforms using the original \textsc{Pangea‑7B}, evidence that the mixed‑data regime had already mitigated catastrophic forgetting. The merged model improves mean accuracy by roughly +0.8 points over the best model, illustrating the value of checkpoint combination.

\section{Related Work}
\subsection{Knowledge Bases for Cultural Grounding}
Integrating structured knowledge bases into language models has proven effective for enriching factual and cultural understanding in text-only NLP. For example, works such as ~\citep{liu2020k, wang2021kepler, kim2022soda, Yang2024SyntheticCP} have attempted to inject world facts in language model representations to provide domain-specific information via knowledge bases. These knowledge-enhanced models achieve better factual consistency and recall, but they operate purely in the textual domain.

Extending knowledge-base grounding to multimodal settings remains relatively rare. Prior work on knowledge-driven VQA has mostly used KBs in answering knowledge-intensive visual questions and focusing on general encyclopedic facts rather than culturally specific knowledge~\citep{deng2025muka,ding2022mukea}. In contrast, our work uniquely leverages a structured knowledge graph to generate culturally diverse VQA pairs. To our knowledge, no previous study has used KBs to curate a multilingual multimodal QA dataset centered on cultural concepts, a gap our proposed pipeline addresses.

\subsection{General-Purpose Multimodal LLMs}
The last few years have seen rapid progress in general-purpose multimodal large language models. Open-source systems like LLaVA~\cite{liu2023visual} and its successors~\citep{liu2024llavanext,liu2023improvedllava} connect vision encoders with LLMs and are trained on visual instruction dataset, achieving impressive results on a wide range of visual tasks. More recent open models such as MoLMo~\citep{deitke2024molmo}, Qwen2.5-VL~\citep{bai2025qwen2}, Phi-3~\citep{abdin2024phi}, further close the gap to proprietary systems by streamlining multimodal training at scale and investing in high-quality datasets. However, it is important to note that none of these general models are explicitly optimized for cultural understanding. Their training data tend to be dominated by English and Western imagery which can lead to blind spots on region-specific content. In summary, general-purpose MLLMs provide a powerful foundation but still exhibit a cultural bias due to their data, leaving room for specialization in multicultural knowledge.

\subsection{Multilingual and Culture-Aware Multimodal LLMs}
To address these gaps, recent efforts have targeted multilingual and cultural understanding in multimodal models by using large-scale multilingual visual instruction datasets, translating existing datasets into a wide array of languages, and using synthetic multilingual instruction pipeline~\citep{yue2024pangea,dash2025aya}. Those works achieve strong performance and indicate a growing focus on true multilingual competence in VLMs.

However, multilingual capability alone does not guarantee cultural understanding~\citep{pouget2024no}. A model trained on many languages may still suffer on culture-specific knowledge, especially if its data are translated or Western-centric. There is increasing recognition that targeted cultural data is needed beyond naive multilinguality. True cultural competence requires exposure to culture-specific content in both text and imagery. Our work follows this principle: rather than relying on pure translated captions, we curate QA pairs grounded in each culture’s unique knowledge and visual context. This approach goes beyond prior multilingual setups by explicitly injecting structured cultural information into the multimodal training data, aiming to build models that are not only linguistically multilingual but also culturally knowledgeable.

\section{Conclusion}
We present a data-centric approach for mining cultural grounded multimodal data from public knowledge bases. \model, a model trained on the resulting dataset demonstrates the effectiveness of the approach and outperforms prior open-source MLLMs on numerous cultural benchmarks such as CVQA, ALMBench, XM100, and MERLIN while preserving general and multilingual vision-language skills. Our findings show that deliberately curating culturally rich data is essential for creating more inclusive multimodal LLMs.

% \newpage
\section{Limitations}
\paragraph{Language and Cultural Coverage}
While \dataset covers 39 languages, its scope remains limited with respect to the full spectrum of world languages and cultures. Future work could potentially incorporate a larger set of languages and cultures to increase diversity and coverage. In addition, transfer learning techniques or multilingual adapters could be explored to improve cross-cultural generalization to languages and cultures not explicitly represented in existing training data.

\paragraph{Potential Bias in Data Distribution}
Despite our effort to promote linguistic and cultural diversity, the dataset might still reflect the underlying biases in global knowledge bases. As shown in \autoref{tab:dataset_stats}, there is still imbalance in the distribution of images, number of unique entities, and number of samples across languages and regions. Countries with higher Gross Domestic Product\footnote{\url{https://en.wikipedia.org/wiki/Gross_domestic_product}} (GDP), such as Japan, tend to have higher number of images, unique entities, and language samples. Higher-resourced languages, such as English, show similar patterns. These imbalance might still skew \dataset, leading \model to perform better on well-represented languages and cultures. Addressing such imbalance remains a challenging yet important direction for future work.

\paragraph{Coverage of Cultural Knowledge}
Our work primarily focuses on grounding MLLMs with factual and entity-centric cultural knowledge, such as occupation and religion, via Wikidata. While this design enables structured scalable data generation, it does not represent the full spectrum of ``cultural knowledge''. Other forms of cultural knowledge, such as social norms, dialects, and implicit values, are not well represented in our dataset. Future work could potentially explore methods to incorporate other dimensions of cultural knowledge training data in a scalable way.

\section{Acknowledgements}

This work was supported in part by a grant from DSTA Singapore.

\bibliography{custom}

% \appendix
% \clearpage
\appendix
\label{sec:appendix}
\section{Additional Analysis}

\subsection{Multimodal Entity Recognition}

We evaluate \textsc{\model} on MERLIN, a benchmark for multilingual multimodal entity recognition and linking. MERLIN is a challenging testbed constructed from news articles paired with images, featuring over 7,000 named entity mentions (linked to 2,500 Wikidata entities) across 5 languages (Hindi, Japanese, Indonesian, Vietnamese, Tamil). This benchmark specifically targets scenarios where textual context alone can be ambiguous, and visual context provides crucial disambiguation an important evaluation for culturally diverse, multilingual settings. On this benchmark shown in \autoref{tab:all_merlin_results}, our model achieves strong results, significantly outperforming the base Pangea-7B model on open-ended entity recognition (i.e., freely identifying the correct entity from multimodal context).
\paragraph{Evaluation Methodology} We employ four evaluation metrics with increasing levels of tolerance to comprehensively assess model performance. The strictest metric, \textbf{Exact Match}, requires predictions to exactly match the target entity name, demanding precise recall of Wikipedia titles. More tolerant metrics better capture semantic understanding and practical utility. The \textbf{+Alias} metric accepts predictions that match any known alias of the target entity, where aliases are sourced from the Wikidata entity's "Also known as" field. For instance, when the target is "Narendra Modi," a model receives credit for predicting any of the common variants: "Modi," "Narendra Bhai," "Narendra Damodardas Modi," "Narendrabhai Damodardas Modi," "Narendrabhai," "Modiji," "Modi Ji," or "NaMo." This reflects real-world usage where entities are referenced through multiple names and honorifics.
The \textbf{+Target} metric accepts cases where the target entity name appears anywhere within the prediction text, accommodating longer descriptive outputs. For example, if the target is "Taj Mahal" and the model predicts "The famous Taj Mahal monument in Agra," this would be considered correct. We deliberately avoid checking if the prediction appears in the target (prediction-in-target) to prevent false positives that could arise from generic terms. Finally, \textbf{All Methods} combines exact matching, alias matching, and target-in-prediction checking to provide the most comprehensive evaluation of model understanding.

\begin{figure*}[t]
    \centering
    \includegraphics[width=\linewidth]{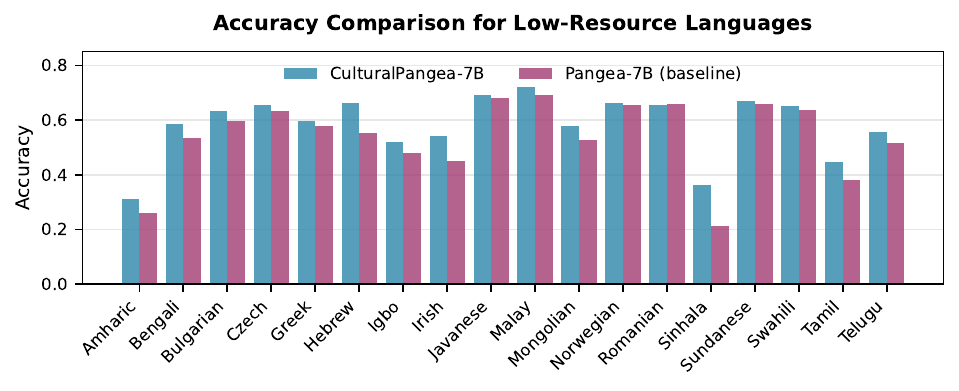}
    \caption{Accuracy improvements on ALM-Bench for low-resource languages with CulturalGround. CulturalPangea achieves the largest gains on underrepresented languages (e.g., Sinhala $+13.54$), demonstrating that CulturalGround effectively scales to the long tail of languages. Performance on other languages remains stable or improves slightly, with only Norwegian and Romanian showing negligible negative changes.}
    \label{fig:lowres-gains-fullset}
\end{figure*}

\begin{figure*}[t]
  \centering
  \includegraphics[width=\linewidth]{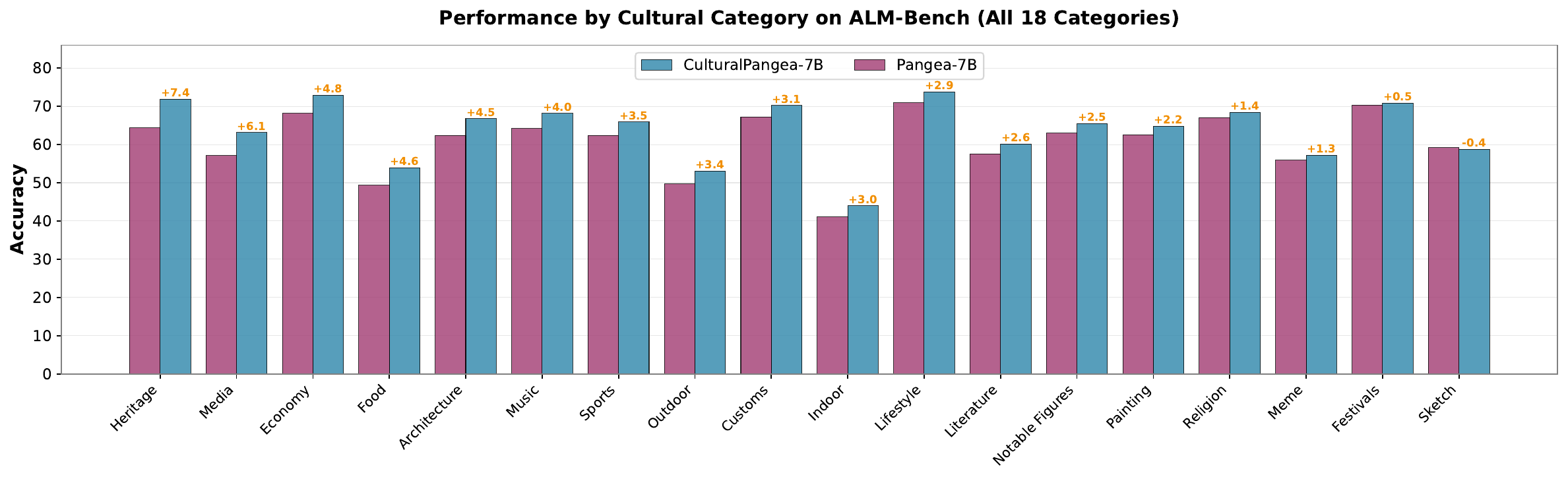}
  \caption{Absolute accuracy gains of CulturalPangea over the baseline across 18 ALM‑Bench cultural domains.  Improvements cluster in culture‑rich categories (Media, Heritage, Music), while Sketch and Meme offer minimal or negative change.}
  \label{fig:all_domains}
\end{figure*}

\begin{table*}[t]
\centering
\scriptsize
\begin{tabular}{@{}llcccll@{}}
\toprule
\textbf{Dataset} & \textbf{Lang/Regions} & \textbf{\#Samples} & \textbf{\#Images} & \textbf{Q Types} & \textbf{Eval Metric} & \textbf{Description} \\
\midrule
\multicolumn{7}{@{}l@{}}{\textbf{Benchmark / Evaluation Datasets}} \\
\midrule
ALM-Bench (2024) & 100L/73 countries & 22.7K & 2.7K & MCQ\&OE & LLM-Judge & Global cultural knowledge across 19 domains \\
% PEARL (2024) & 1L/Arab world & 6,867 & 6.9k & Both & Accuracy & Arabic cultural content (food, festivals, etc.) \\
CVQA (2024) & 31L/30 countries & 10K & 4.5K & MCQ\&OE & Accuracy & Under-represented countries, local experts \\
% FoodieQA (2024) & 2L/China & 700 & 389 & MCQ & Accuracy & Fine-grained Chinese regional cuisines \\
% xGQA (2022) & 7L/General & 12,578 & 398 & MCQ & Accuracy & Multilingual GQA (no cultural focus) \\
MaXM (2023) & 7L/General & 2.1K & 335 & OE & Relaxed-Acc & Translation-based VQA (no cultural focus) \\
M3EXAM (2023) & 9L/General & 12.3K & 2.8K & MCQ & Accuracy & Multilingual exams (no cultural focus) \\
XM100 (2022) & 36L/Global & 3.6K & 100 & Caption & CIDEr & Multilingual captioning \\
% EVJVQA (2023) & 3L/Vietnam & 33k+ & 5k & Both & Accuracy & Vietnamese cultural items and scenes \\
% CulturalVQA (2024) & 1L/11 countries & 2,378 & 2,328 & OE & BLEU/CIDEr & Cultural traditions and practices \\
% SEA-VQA (2024) & 1L/SE Asia & 1,999 & 515 & MCQ & Accuracy & 53 under-represented SE Asian cultures \\
MERLIN (2024) & 5L/4 countries & 7.1K & 4.2K & OE & Accuracy & Multilingual and cultural entity recognition \\
% WorldCuisines (eval) & 30L/189 countries & 72k & 6,045 & Global culinary & MCQ & Accuracy & World cuisines identification \\
\midrule
\multicolumn{7}{@{}l@{}}{\textbf{Training Datasets}} \\
\midrule
WorldCuisines-train(2024) & 30L/189 countries & 1.08M & 6K & MCQ\&OE & -- & Large-scale global cuisine training data \\
IndiFoodVQA (2024) & 1L/India & 16.7K & 16K & MCQ & -- & Knowledge-infused Indian food VQA \\
SEA-VL(2025) & 6L/11 SE Asia & 1.28M & 1.28M & Caption & -- & Authentic SEA cultural visual scenes \& short captions \\
PEARL-train(2025) & 1L/19 Arab countries & 309K & 12.6K & MCQ\&OE & -- & Arabic cultural instruction dataset \\
\midrule
% \rowcolor{gray!20}
\textbf{CulturalGround (2025)} & \textbf{39L/42 regions} & \textbf{30M} & \textbf{2.8M} & \textbf{MCQ\&OE} & \textbf{--} & \textbf{Massive multilingual diverse cultural VQA} \\
\bottomrule
\end{tabular}
\caption{Comparison of culturally diverse vision-language datasets. Our CulturalGround dataset (highlighted) provides the largest-scale multilingual cultural training data. We compare with WorldCuisines-train~\citep{winata2024worldcuisines}, IndieFoodVQA~\citep{agarwaletal2024indifoodvqa}, SEA-VL~\citep{cahyawijaya2025crowdsource}, and PEARL-train~\citep{alwajih2025pearl}}
\label{tab:cultural_datasets}
\end{table*}

\subsection{Performance Gains from Merging Different Checkpoints}

\begin{figure*}[t]
    \centering
    \includegraphics[width=\linewidth]{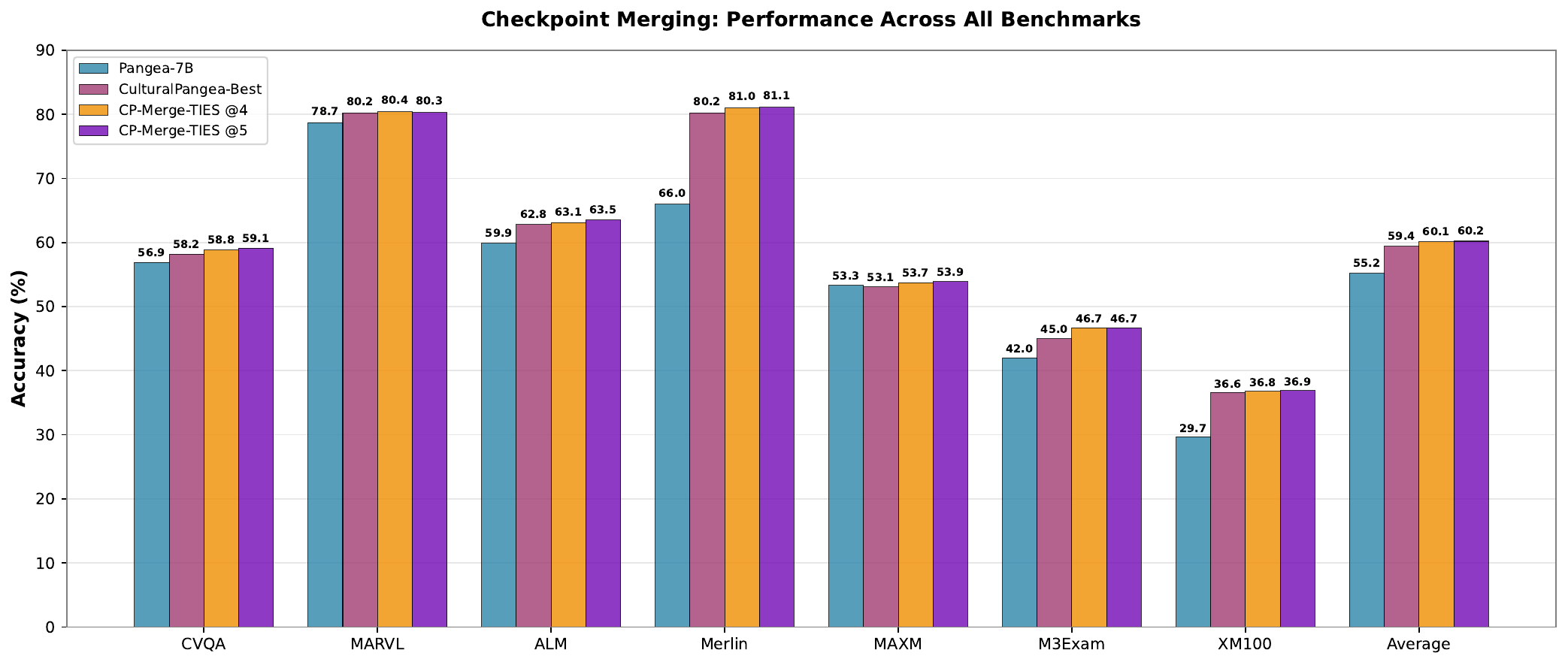}
    \caption{Accuracy improvements from merging checkpoints across all evaluation benchmarks. CK stands for CulturalPangea.}
    \label{fig:checkpoint-all-benchs}
\end{figure*}

We experiment with various checkpoint merging methods such as linear~\citep{wortsman2022model}, TIES~\citep{yadav2023ties}, DARE-TIES~\citep{yu2024language}. All methods improve performance over our best model and baseline and we don't rule out that neither is the best as also observed in~\citep{li2025model}. Performance across the board improves with number of checkpoints up to a certain ceiling, where the choice of merging methods, number of checkpoints, and weight assigned to each checkpoint only makes marginal difference. \autoref{fig:checkpoint-all-benchs} shows the comparison and gains between Pangea-7B(baseline), \model(best checkpoint), and merged model with 4, and 5th latest checkpoints.

\begin{figure}[t]
\centering
\includegraphics[width=0.9\linewidth]{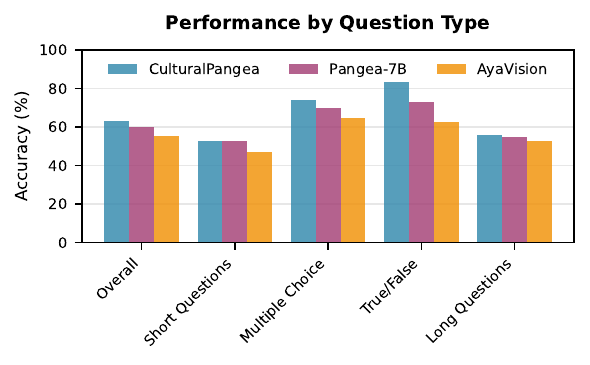}
\caption{Performance on different types of questions in ALMBench: multiple choices, open-ended, and true/false.}
\label{fig:almbench_question_types}
\end{figure}

% \section{When CulturalPangea Succeeds (and Fails)}

\section{Additional Details on Data Curation}

\subsection{Cultural Relevant Wikidata Properties and Templates QA}

\autoref{tab:properties_list} presents a comprehensive catalog of culturally relevant Wikidata properties employed in our entity extraction process. This table includes the specific properties used to identify cultural entities within Wikidata's knowledge graph, along with the corresponding template questions and template answers that form the foundation of our multilingual VQA generation pipeline.

\subsection{Refining Multilingual QA Data for Cultural Fluency}

The prompt used in refining and rephrasing multilingual QA data is shown in \autoref{fig:rephrasing_prompt}.
\begin{figure*}[t]
\definecolor{highlightcolor}{RGB}{46, 134, 171}
\definecolor{framecolor}{RGB}{162, 59, 114}
\resizebox{\textwidth}{!}{%
\begin{tcolorbox}[
  colback=highlightcolor!15, 
  colframe=framecolor, 
  boxrule=1.0pt, 
  title=Prompt for Improving Fluency in Cultural Templates QAs,
  label=cultural-entity-vqa-template,
  breakable,
  width=\textwidth
]

\textbf{System Prompt:} \\
You are a cultural expert specializing in creating high-quality, culturally sensitive questions and answers about diverse entities from around the world. Your goal is to create natural-sounding questions and factually accurate answers that respect cultural nuances while maintaining value about important properties of entities such as location, category, administrative territory, and other key attributes. \\

\textbf{User Prompt:} \\
Given this entity and context in \{language\_name\}: \\
\textit{Entity:} \{label\} \\
\textit{Description:} \{description\} \\
\textit{Entity Region(Country):} \{region\} \\
\textit{Original Question:} \{question\} \\
\textit{Original Answer:} \{answer\} \\

\textit{Task:} Create both a natural question and an answer for a visual question answering dataset focused on cultural recognition of entities in multilingual contexts. \\

\textbf{For the question:} \\
1. Maintain the precise property being asked about in the original question (like location, category, administrative territory, awards, etc.) \\
2. Use natural, conversational phrasing in authentic \{language\_name\} that a native speaker would use \\
3. Do NOT reveal specific details about the entity in the question unless absolutely necessary \\
4. Ensure cultural sensitivity and respect for local naming conventions and terminology \\
5. Make the question grammatically correct, clear, and unambiguous \\
6. Phrase it as if someone is looking at an image of this entity and asking about it \\
7. Avoid awkward phrasing or other template artifacts \\

\textbf{For the answer:} \\
1. Ensure complete factual accuracy based on the provided information \\
2. Use natural language appropriate for \{language\_name\} with proper cultural context \\
3. Include key factual details from the original answer and leave out any unnecessary information \\
4. When appropriate, provide brief additional cultural context or significance of the entity \\
5. Make sure to include the full entity name and keep the answer around the property being asked about \\
6. Avoid vague phrases - be specific and informative. Ensure the answer is clear, concise, relevant, don't include unnecessary details. \\
7. It is best to avoid adding new information unless it is a well-known fact about the entity that enhances understanding. \\
8. We are grounding the model to cultural knowledge, so it is really important to be accurate and keep answers factually correct. \\
9. The region/country of the entity \{region\} is provided to you for context, so please don't confuse the entity with other regions or countries. \\

% \textbf{Examples based on real Wikidata entity questions and answers:} \\

% \textit{Example 1 (Location Question)} \\
% Original Question: "Which sovereign state does this entity belong to? (not for humans)" \\
% Original Answer: "Phủ Lý belongs to the sovereign state of Vietnam, country in Southeast Asia." \\
% Good Reformulation: \\
% Q: "In which country is this city located?" \\
% A: "Phủ Lý is located in Vietnam, a Southeast Asian nation. It serves as the provincial capital of Hà Nam Province in northern Vietnam." \\

% \textit{Example 2 (Category/Classification)} \\
% Original Question: "This entity is an example of which class or category?" \\
% Original Answer: "Roman Catholic Diocese of Da Lat is an instance of diocese of the Catholic Church, see under the supervision of a bishop of the Catholic church." \\
% Good Reformulation: \\
% Q: "What type of religious administrative division is shown in this image?" \\
% A: "This is the Roman Catholic Diocese of Đà Lạt, a diocese of the Catholic Church supervised by a bishop. It serves the Catholic community in Vietnam's Central Highlands region." \\

\textbf{Format your response exactly as:} \\
Q: [your reformulated question] \\
A: [your reformulated answer]

\end{tcolorbox}
}
\caption{Refining QA Prompt}
\label{fig:rephrasing_prompt}
\end{figure*}

\subsection{Multilingual MCQ and True/False Question Generation Pipeline}
\label{app:mcqs_pipeline}

To create culturally grounded VQA examples, we developed an entity-driven question generation pipeline that produces both multiple-choice and true/false questions across languages. Each culturally significant entity (identified via Wikidata) served as a seed for generating question–answer pairs in multiple languages. Using a multilingual large language model (Gemma-3-27B and Qwen-2.5-Instruct), we automatically generated questions based on each entity’s metadata (i.e., a known factual attribute of that entity). In particular, we prompted the model to generate a four-option multiple-choice question (MCQ) in the target language for each entity, and for format diversity we also created a yes/no true/false question for a subset of entities. This approach yielded roughly 8 million initial question–answer samples, with approximately 60\% in MCQ format and 40\% in true/false format, providing a balanced mix of question types. Moreover, each entity was covered in several languages (e.g., Hindi, Chinese, Arabic, Swahili), ensuring that the dataset captured culturally authentic knowledge across diverse linguistic contexts.

\noindent\textbf{Prompt Design and Multilingual Strategy.} We carefully crafted the generation prompts to preserve each question’s cultural context and authenticity. The system prompt instructed the model to act as a “cultural expert,” following detailed guidelines with in-context examples in the target language to demonstrate the desired question format. Specifically, the prompt emphasized using natural, engaging phrasing in the local language; respecting cultural sensitivities and naming conventions; avoiding revealing the answer in the question; ensuring correct grammar and clarity; varying the question style beyond simple identification; and aiming for a moderate difficulty level that tests deeper cultural knowledge. These measures kept the questions grounded in each entity’s cultural background and prevented trivial or overly direct queries. We also tailored the prompts for each target language by providing example Q\&A pairs in that language’s script and style. This multilingual prompting strategy enabled the model to produce fluent, culturally nuanced questions in every target language without losing authenticity.

\noindent\textbf{Multiple-Choice Question Generation.} For MCQs, the model was prompted to produce a question followed by four answer options (A, B, C, D), where option A was the correct answer\footnote{Option A is always the correct answer for output parsing consistency, during training we use shuffled options to ensure model does not overfit or prefer one option over the others.} (a true fact about the entity) and options B, C, and D were plausible but incorrect distractors drawn from the same cultural or topical domain. All answer choices were written to be similar in style and length and culturally sensible, so as to avoid any obvious giveaway of the correct option. In addition, the model was asked to provide a brief explanation justifying why option A is correct, incorporating relevant cultural or historical context. This process yielded rich multiple-choice items, for instance, a Hindi question about a historical dance might present one correct cultural tradition and three other plausible traditions as answer choices, all phrased naturally in Hindi script.

\noindent\textbf{True/False Question Generation.} We adopted a similar strategy for true/false items. The model generated a concise factual statement about the entity and labeled it as either true or false, effectively forming a yes/no question. To ensure these binary questions were non-trivial, we directed the model to leverage the entity’s attributes when crafting statements, for example, introducing a plausible but incorrect detail when a false statement was required so that answering the question would require specific knowledge of the entity’s background. Each true/false question was written in the target language with culturally appropriate wording (for instance, a Swahili prompt might ask whether a certain landmark is located in a particular region, with the answer marked “False” if that region is incorrect).

\noindent\textbf{Relevance Filtering.} All generated questions underwent rigorous filtering to ensure accuracy and cultural validity. We filtered out any outputs that were irrelevant and factually incorrect. From the roughly 8 million candidates initially generated, we retained only the best, non-trivial questions, resulting in a final \textsc{CulturalGround} dataset of about 6 million culturally MCQs samples. This final collection preserved a balanced mix of question formats and, coupled with our culturally-informed prompt design, ensured that the questions remained authentic to each culture while covering a broad range of entities and properties. Overall, this multilingual MCQ/true–false generation pipeline allowed us to inject diverse cultural knowledge into the VQA data in a high-quality manner, which was crucial for training the \textsc{CulturalPangea} model.

Finally, we provide the full prompt templates used for MCQ generation in \autoref{card: cultural-mcq-template} and True/False in \autoref{card:tf-prompt-template}.

\section{Cultural Domains and Long-Tail Entity Coverage in CulturalGround}

\begin{figure*}[!htbp]
  \centering
  \begin{minipage}[c]{1.0\linewidth}
    \includegraphics[width=\linewidth]{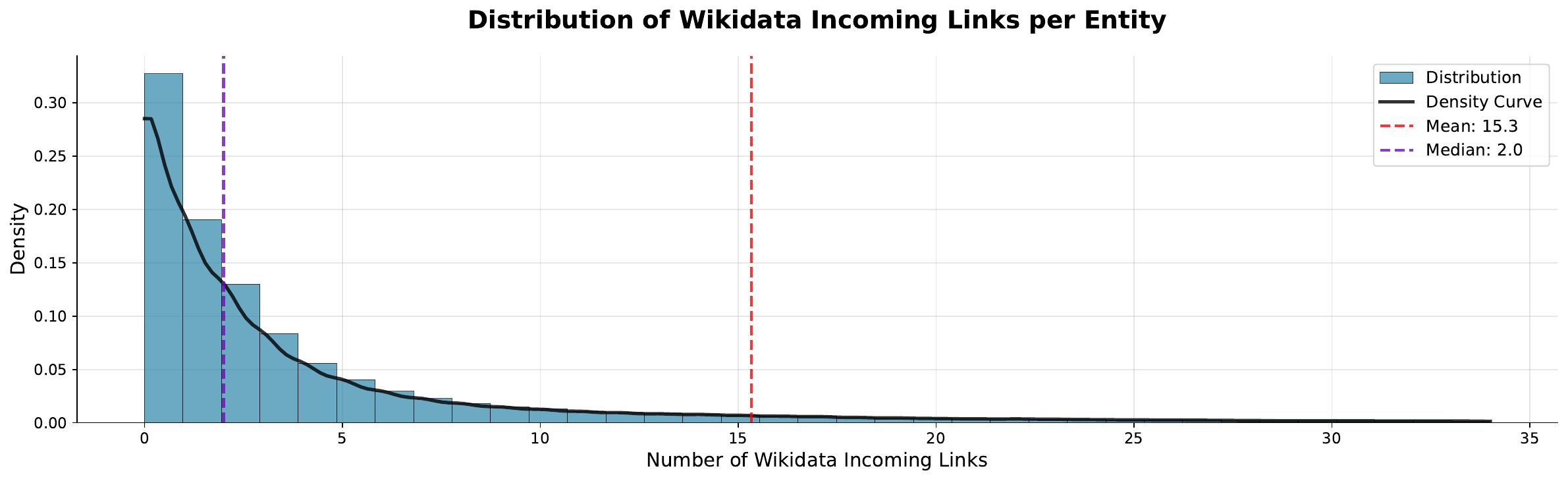}
  \end{minipage}
  
  \vspace{0.3cm}
  
  \begin{minipage}[c]{1.0\linewidth}
    \includegraphics[width=\linewidth]{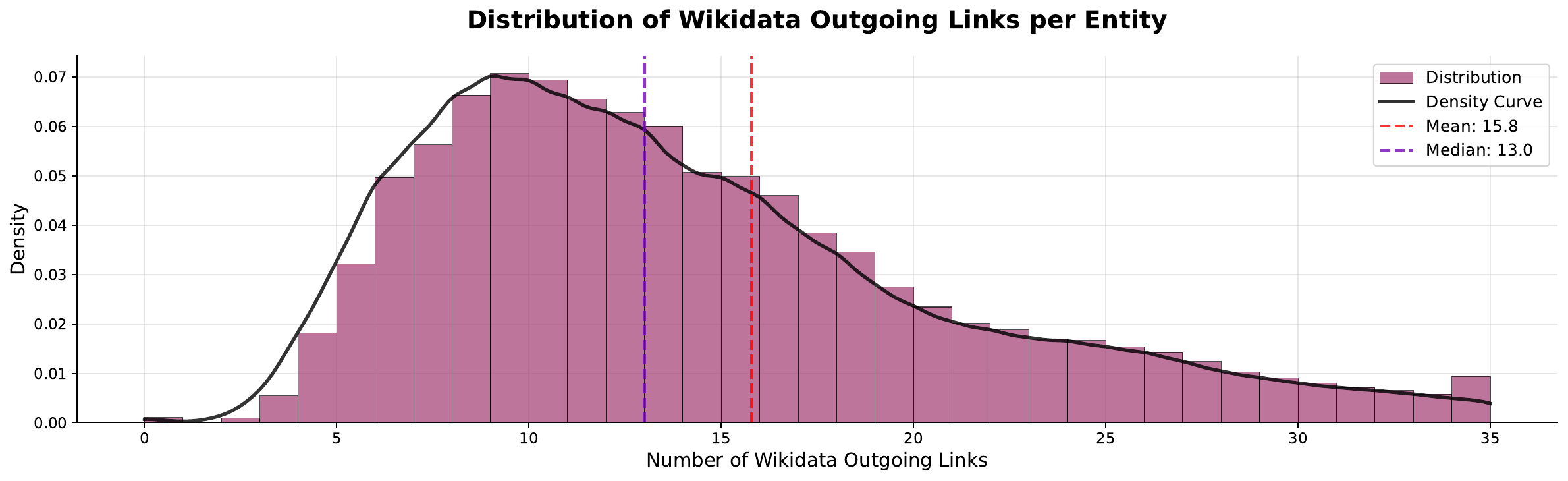}
  \end{minipage}
  
  \vspace{0.3cm}
  
  \begin{minipage}[c]{1.0\linewidth}
    \includegraphics[width=\linewidth]{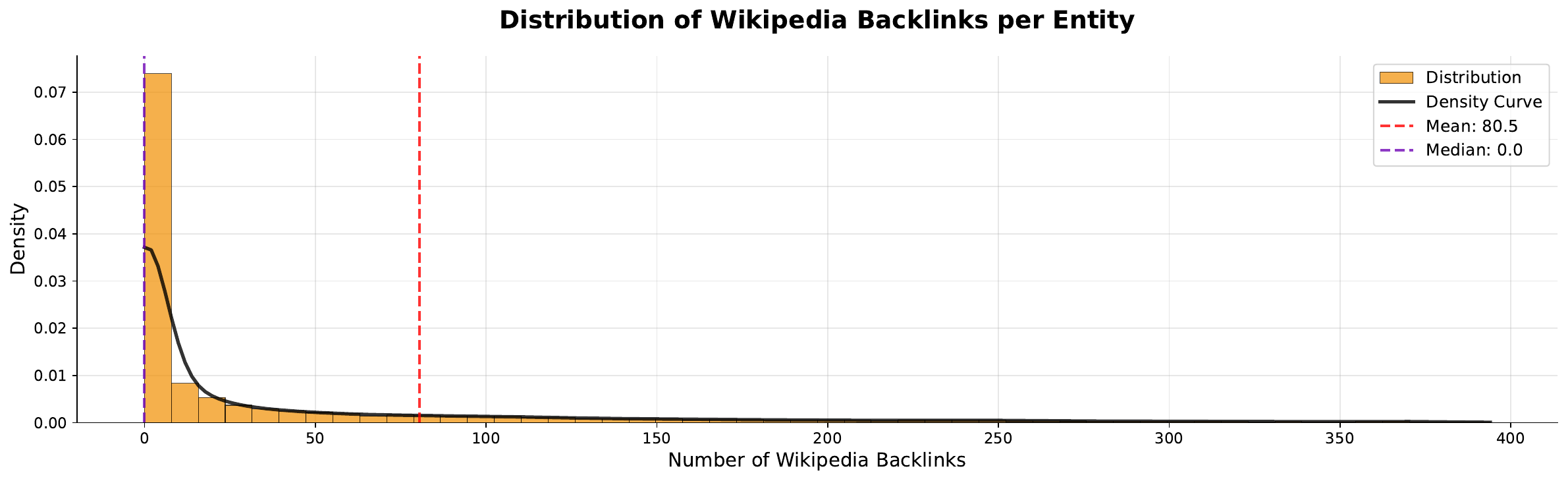}
  \end{minipage}
  
  \caption{Distribution of entity connectivity in \dataset, demonstrating our commitment to treating long-tail entities as first-class citizens. The plots show (top) incoming Wikidata links, (middle) outgoing Wikidata links, and (bottom) Wikipedia backlinks across all countries. Unlike previous datasets that focus primarily on highly popular entities, \dataset includes substantial representation of entities with few or no links, as evidenced by the significant density at lower links values.}
  \label{fig:entity_connectivity_distribution}
\end{figure*}

\begin{figure*}[h]
    \centering
    \includegraphics[width=\linewidth]{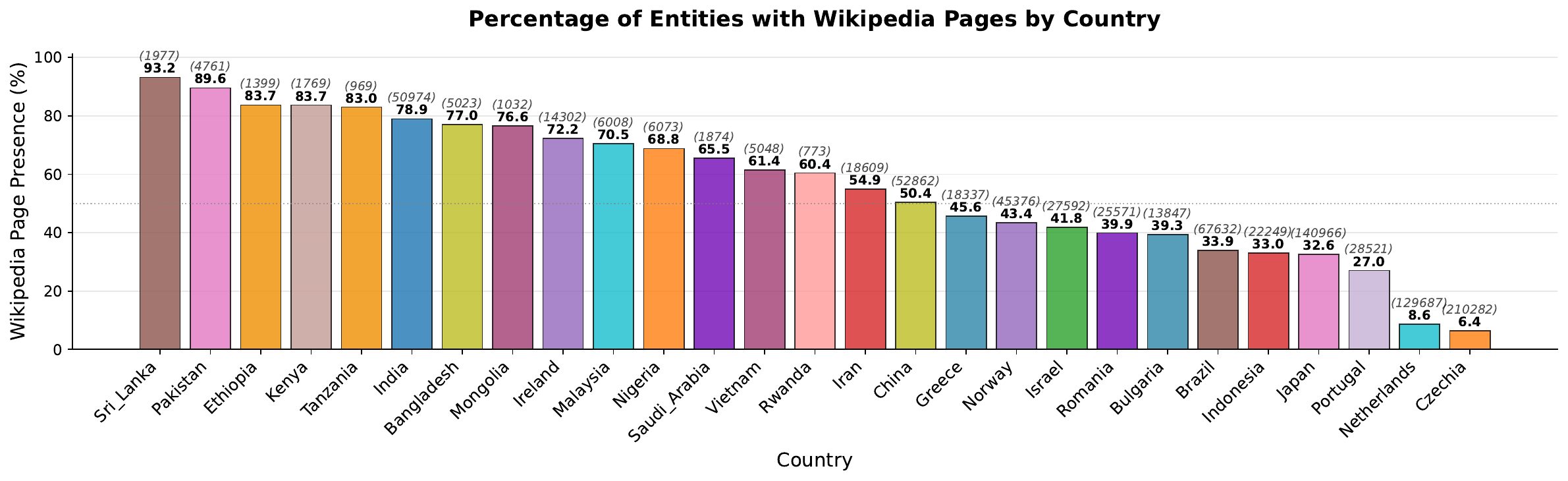}
    \caption{The proportion of entities that have Wikipedia page in some regions in \dataset. The majority of entities don't have Wikipedia presence, confirming that our data curation pipeline captures long-tail entities}
    \label{fig:wikipedia-presence}
\end{figure*}

\noindent\textbf{Long-tail entities emphasis.}
\dataset treats long-tail entities as the first-class citizen. As shown in \autoref{fig:entity_connectivity_distribution}, most entities in the dataset have only a small number of connections, such as very few incoming or outgoing links in the knowledge graph or minimal Wikipedia backlinks. Consequently, median link counts are extremely low, indicating that the typical entity in \textit{CulturalGround} is sparsely connected. This distribution is highly right-skewed, with a long tail: a handful of entities are linked to many others, but the vast majority are referenced only a few times. Such patterns underscore the dataset’s focus on culturally specific, niche entities that lie beyond the well-connected head of popular or globally known concepts. Notably, a substantial portion of the entities included have no dedicated Wikipedia page at all(see \autoref{fig:wikipedia-presence}. The prevalence of entries lacking Wikipedia articles further highlights the dataset’s extension into culturally important yet under-documented regions of knowledge, reinforcing the long-tail presence.

\noindent\textbf{Regional domain emphasis.}
Beyond its long-tail connectivity profile, \textit{CulturalGround} spans a broad array of domains and regions, surfacing localized cultural themes that vary systematically by country. 
% In India, heritage and settlement categories (e.g., villages, Hindu temples) are prominent; in Indonesia, religious architecture and local administrative units (e.g., mosques, \textit{desa}/\textit{kelurahan}) dominate; Portugal foregrounds ecclesiastical heritage and protected cultural assets (e.g., church buildings, cultural heritage, chapels); Israel highlights settlement types and archaeological sites (e.g., \textit{moshav}/\textit{kibbutz}, archaeological sites); Brazil exhibits a strong presence of cultural media and monuments (e.g., photographs, slides/negatives, historic sites); and Romania concentrates on rural settlements and ecclesiastical heritage (e.g., villages/communes, church and wooden-church categories). These region-specific concentrations, together with the heavy-tailed degree and backlink distributions, demonstrate that \textit{CulturalGround} intentionally emphasizes culturally grounded, long-tail entities across diverse domains.
For example, India’s entries are dominated by heritage and settlement categories (e.g., numerous Hindu temples and villages), reflecting the country’s rich corpus of historical sites and settlements; the Netherlands highlights vernacular architecture and art categories, aligning with its distinctive building traditions and artistic heritage; and Portugal’s distribution is led by ecclesiastical heritage categories, indicative of its deep‐rooted religious and historical legacy. Similarly, China’s data feature prominent infrastructure and administrative categories (such as railway stations and provincial divisions), echoing an emphasis on civic infrastructure and governance; Brazil is characterized by an abundance of cultural media and monument categories, mirroring its vibrant media culture and iconic landmarks; and Japan’s prominent categories include religious architecture (temples and shrines) and education (schools and universities), underscoring the country’s spiritual traditions and academic legacy. This diversity of region‐specific dominant categories underscores \textit{CulturalGround}'s broader goal of surfacing underrepresented cultural knowledge by grounding the dataset in regionally significant, long‐tail cultural entities.

\begin{figure*}[t]
\definecolor{highlightcolor}{RGB}{46, 134, 171}
\definecolor{framecolor}{RGB}{162, 59, 114}
\resizebox{\textwidth}{!}{%
\begin{tcolorbox}[
  colback=highlightcolor!15, 
  colframe=framecolor, 
  boxrule=1.0pt, 
 title=Cultural Entity Multiple-Choice Question Generation,
 label=cultural-mcq-template,
 enhanced,
 width=\textwidth
]

\textbf{System Prompt:} \\
You are a cultural expert who creates high-quality multiple-choice questions about entities while preserving cultural context, and authenticity. \\

\textbf{User Prompt:} \\
Given this entity and context in \{language\_name\}: \\
\textit{Entity:} \{label\} \\
\textit{Description:} \{description\} \\
\textit{Original Question:} \{question\} \\
\textit{Original Answer:} \{answer\} \\
\textit{Region/Country:} \{region\} \\

\textit{Task:} Create a multiple-choice question with four options (A, B, C, D) based on the cultural entity. \\

\textbf{For the multiple-choice question:} \\
1. Maintain the original topic but use natural, engaging phrasing in \{language\_name\} \\
2. NEVER reveal specific details about the entity in the question unless necessary \\
3. Respect cultural context and sensitivity \\
4. Make it grammatically correct, clear, and culturally relevant \\
5. Vary question formats beyond basic identification \\
6. Create questions with appropriate difficulty level \\
7. Aim for questions that test deeper cultural knowledge \\
8. Keep the entity as the grounding point \\

\textbf{For the options:} \\
1. Option A should ALWAYS be the correct answer \\
2. Create three plausible but incorrect options (B, C, D) \\
3. All options should be culturally accurate, sensible, and realistic \\
4. Ensure all options are similar in length and format \\
5. All incorrect options should be from the same general category \\
6. Options should represent meaningful distinctions but yet plausible and challenging within the cultural/regional context \\

\textbf{For the explanation:} \\
1. Briefly explain why option A is correct \\
2. Include relevant cultural or historical context \\
3. Keep explanation concise but informative (1-3 sentences) \\
4. Keep the entity as the grounding point \\

\textit{Example format will be provided based on language context.} \\

\textbf{Create a multiple-choice question for this entity in exactly this format:} \\
Q: [your multiple-choice question] \\
A) [correct answer] \\
B) [plausible incorrect option] \\
C) [plausible incorrect option] \\
D) [plausible incorrect option] \\
Correct: A \\
Explanation: [brief explanation with cultural context]

\end{tcolorbox}
}
\caption{Refining QA Prompt}
\label{card: cultural-mcq-template}
\end{figure*}
\begin{figure*}[t]
\definecolor{highlightcolor}{RGB}{46, 134, 171}
\definecolor{framecolor}{RGB}{162, 59, 114}
\resizebox{\textwidth}{!}{%
\begin{tcolorbox}[
 colback=highlightcolor!15, 
 colframe=framecolor, 
 boxrule=1.0pt, 
 title=Cultural Entity True/False Question Generation,
 label=cultural-tf-template,
 width=\textwidth,
]
\textbf{System Prompt:} \\
You are a cultural expert who creates clear and culturally-sensitive true/false statements/questions about entities that test understanding while preserving authenticity. \\

\textbf{User Prompt:} \\
Given this entity and context in \{language\_name\}: \\
\textit{Entity:} \{label\} \\
\textit{Description:} \{description\} \\
\textit{Original Question:} \{question\} \\
\textit{Original Answer:} \{answer\} \\
\textit{Region/Country:} \{region\} \\

\textit{Task:} Create a true/false statement based on the cultural entity. \\

\textbf{For the statement/question:} \\
1. Create either a clear statement OR a yes/no question about the entity in \{language\_name\} \\
2. Mix between statements and questions for variety \\
3. Make it unambiguous - clearly either true or false \\
4. Test meaningful cultural knowledge, not trivial details \\
5. Respect cultural sensitivity \\
6. Keep the entity as the central focus \\
7. Vary between true and false answers for diversity \\

\textbf{For the explanation:} \\
1. Briefly explain why the statement is true or false \\
2. Include relevant cultural or historical context \\
3. Keep it concise (1-2 sentences) \\

\textit{Example format will be provided based on language context.} \\

\textbf{Create a true/false item for this entity in exactly this format:} \\
Statement: [your true/false statement] \\
Answer: [True/False] \\
Explanation: [brief explanation] \\
\\
OR \\
\\
Question: [your true/false question] \\
Answer: [True/False] \\
Explanation: [brief explanation]
\end{tcolorbox}
}
\caption{True/False Question Generation Prompt Template}
\label{card:tf-prompt-template}
\end{figure*}
\begin{figure*}[t]
\definecolor{highlightcolor}{RGB}{46, 134, 171}
\definecolor{framecolor}{RGB}{162, 59, 114}
\resizebox{\textwidth}{!}{%
\begin{tcolorbox}[
  colback=highlightcolor!15, 
  colframe=framecolor, 
  boxrule=1.0pt, 
  title=Prompt for Evaluating VQA Sample Quality and Image-Entity Alignment,
  label=qwen-vqa-evaluation-template,
  breakable,
  width=\textwidth
]

\textbf{System Prompt:} \\
You are an expert at evaluating whether images match with cultural entities and their descriptions. You assess alignment between visual content and textual information. \\

\textbf{User Prompt:} \\
Evaluate this VQA sample for quality and alignment. \\

\textbf{Entity Information:} \\
\textit{Label:} \{label\} \\
\textit{Description:} \{description\} \\
\textit{Region/Country:} \{region\} \\
\textit{Language:} \{language\} \\
\textit{Question:} \{question\} \\
\textit{Answer:} \{answer\} \\

\textbf{Your task is to determine:} \\
1. Does the image show or reasonably represent the entity described? \\
2. Are there any quality issues with this sample? \\

\textbf{Common issues to check for:} \\
1. Image completely unrelated to the entity (e.g., entity is about a person, but image is of animal. Or entity is about park but image show city, or entity is about a person but image is of a building) \\
2. Mixed languages in question or answer \\
3. Obvious factual errors in the answer that you can confirm and very sure about \\
4. Question and answer mismatch \\
5. Corrupted or incomplete answer \\

\textbf{If you are not sure about the answer:} \\
1. Treat sample as match and no issue \\
2. We are mostly concerned with the image being completely irrelevant to the entity and we understand some models may not know some long-tail entities \\
3. So unless there is a clear mismatch or quality issue in rephrased question/answer, treat it as match \\

\textbf{Other considerations:} \\
1. If the question asks about education or birth place or other entity properties, treat it as a match if the image is related to the entity, even if it does not show the specific property \\
2. If the image is not provided, treat it as match unless the answer is clearly unrelated to the entity or has problematic issues mentioned above \\
3. I repeat, if you are not sure about your answer and can not confirm it which might happen alot with long-tail entities, treat it as match and no issue \\

\textbf{Format your response exactly as (Notice and keep the line breaks):} \\
MATCH: [True/False] \\
ISSUE: [None/ImageMismatch/MixedLanguage/FactualError/QAMismatch/Unclear] \\
EXPLANATION: [Brief explanation of your assessment]

\end{tcolorbox}
}
\caption{VQA Quality Evaluation Prompt}
\end{figure*}
\begin{figure*}[t]
\definecolor{highlightcolor}{RGB}{46, 134, 171}
\definecolor{framecolor}{RGB}{162, 59, 114}
\resizebox{\textwidth}{!}{%
\begin{tcolorbox}[
  colback=highlightcolor!15, 
  colframe=framecolor, 
  boxrule=1.0pt, 
  title=Prompt for Evaluating MCQ Quality and Cultural Alignment,
  label=qwen-mcq-evaluation-template,
  breakable,
  width=\textwidth
]

\textbf{System Prompt:} \\
You are an expert at evaluating MCQ quality and cultural alignment. You assess whether questions match with cultural entities and check for quality issues. \\

\textbf{User Prompt:} \\
Evaluate this MCQ sample for quality and alignment. \\

\textbf{Entity Information:} \\
\textit{Label:} \{label\} \\
\textit{Description:} \{description\} \\
\textit{Region/Country:} \{region\} \\
\textit{Language:} \{language\} \\
\textit{Question Type:} \{question\_type\} \\
\textit{Question:} \{question\} \\
\textit{Options:} \{options\_text\} \\
\textit{Correct Answer:} \{correct\_answer\} \\
\textit{Explanation:} \{explanation\} \\

\textbf{Your task is to determine:} \\
1. Does the image (if present) reasonably represent the entity described? \\
2. Is the question culturally relevant to the specified region? \\
3. Are there any quality issues with this MCQ? \\

\textbf{Common issues to check for:} \\
1. Image completely unrelated to the entity or question \\
2. Question not relevant to the cultural context or region \\
3. Incorrect answer or poor explanation \\
4. Mixed languages in question, options, or explanation \\
5. Poorly formed question or confusing options \\
6. Factual errors you can confirm \\

\textbf{Guidelines:} \\
1. If you are not sure about cultural relevance or correctness, treat it as acceptable \\
2. Focus on obvious mismatches and clear quality issues \\
3. For questions without images, focus on cultural relevance and question quality \\
4. Consider regional context when evaluating cultural appropriateness \\

\textbf{Format your response exactly as:} \\
MATCH: [True/False] \\
CULTURALLY\_RELEVANT: [True/False] \\
ISSUE: [None/ImageMismatch/CulturalMismatch/IncorrectAnswer/MixedLanguage/
PoorQuestion/FactualError/Other] \\
EXPLANATION: [Brief explanation of your assessment]

\end{tcolorbox}
}
\caption{MCQ Quality Evaluation Prompt}
\end{figure*}

\section{Balancing Regions and Languages With Hybrid Sampling}

\label{app:sampling}
\begin{figure*}[t]
    \centering
    \includegraphics[width=\linewidth]{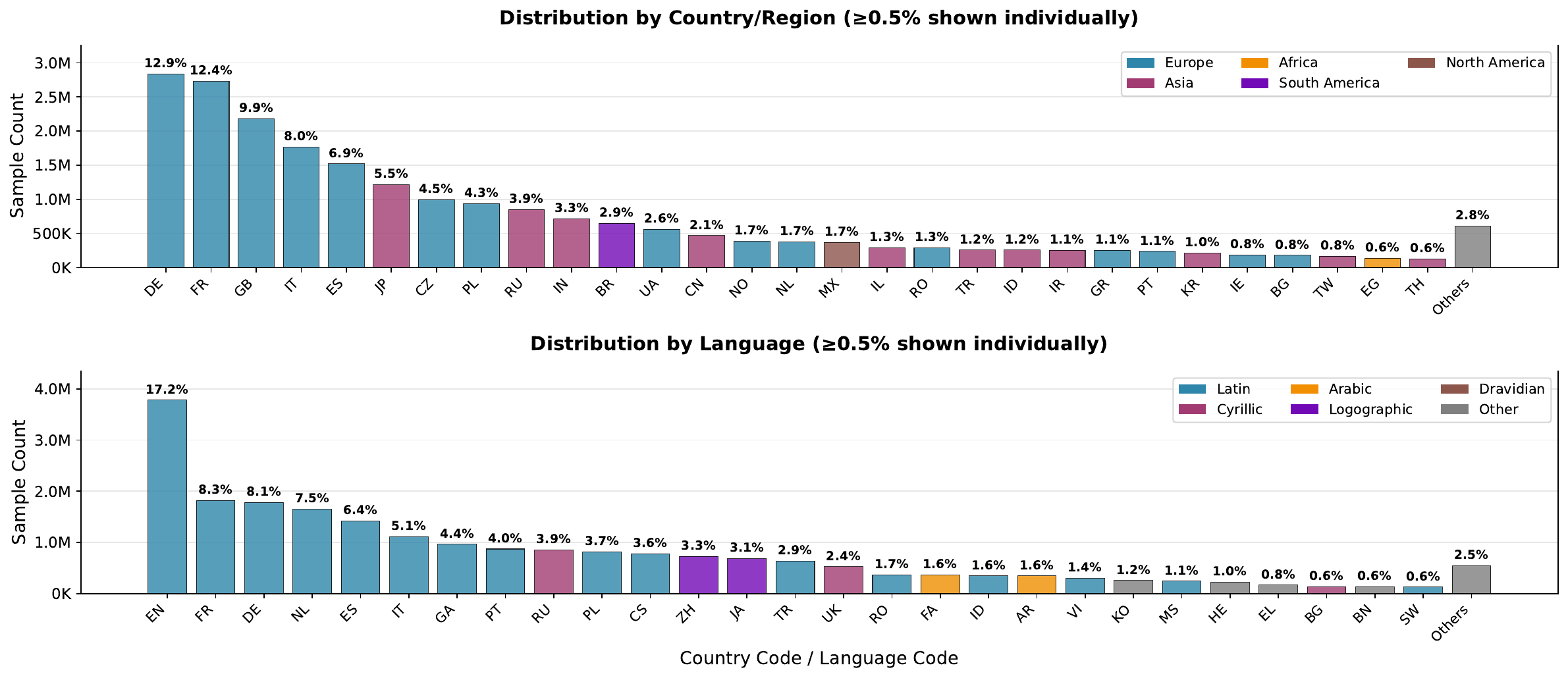}
    \caption{The distribution of languages and countries in \dataset}
    \label{fig:countries-langs}
\end{figure*}

\autoref{fig:countries-langs}, \autoref{tab:dataset_stats}, and \autoref{tab:lang_data_stats} provide comprehensive statistics on the training data across regions and languages. While preparing the training data, to mitigate the imbalance between high-resourceful and low-resourceful regions and languages, we employ hybrid temperature-based sampling~\citep{arivazhagan2019massively}, where we first sample data by regions to cap high-resourceful regions without affecting smaller regions, and then finally apply smooth language sampling to balance languages. We use temperature of 4.0 for region sampling and 1.5 for language sampling. We use small temperature in language sampling to keep cross-lingual associations between entities.

% including both raw extraction counts and the remaining samples after applying temperature sampling~\citep{arivazhagan2019massively} for regional balance. For the full unfiltered data, we employed hybrid sampling with temperature values of 4 for regional sampling and 1.5 for language sampling to achieve balanced representation across diverse cultural contexts.

% \autoref{tab:region_sampling_filtered} further details the data refinement process, showing sample counts both before and after the quality filtering stage and subsequent region sampling procedures. For the filtered data, we applied more conservative temperature sampling parameters of 3 for regional sampling and 1.5 for language sampling to maintain data quality while preserving cultural diversity. Finally, \autoref{fig:language_dist_plot_filtered} illustrates the final distribution of samples across languages following the complete language sampling process.

\section{M3LS Dataset Creation}
\label{app:m3ls}

M3LS is a large-scale multi-lingual, multi-modal summarization dataset introduced by~\citep{verma2023large}. It contains news articles with images in 20 languages (sourced from BBC News over a decade), paired with professionally annotated summaries. We leverage this dataset to automatically create an \textbf{entity-centric} data source for improving cross-lingual entity recognition.

\textbf{Automated Entity Extraction Pipeline:} For each news article in five high-diversity languages (Hindi, Tamil, Indonesian, Japanese, Vietnamese) plus English, we use a specialized LLM prompt to identify the article's most central entity. The prompt provides the article's title, summary, first paragraph, and any image captions, and asks the model to return: (1) the main entity mention (in the original language), (2) its corresponding English Wikipedia page title, and (3) a brief justification. This process is fully automated and yields a structured JSON output per article containing the extracted entity and its English Wikipedia reference. By querying an LLM in this way, we effectively perform cross-lingual entity linking: mapping non-English entity mentions to a canonical English Wikipedia title.

\textbf{Data Curation and Filtering:} The raw generation produced approximately 203K candidate Q\&A pairs (image + question-answer) across the six languages. Each instance uses the article's image and asks a question like ``What is the Wikipedia page title that corresponds to the entity `X' mentioned in the news article `Y'?'', with the answer being the English Wikipedia title for entity X. We then removed 4,575 instances that overlapped with the MERLIN dataset (to avoid test contamination) and reduced the exceedingly large English portion to 50K examples. This resulted in a final training set of $\sim$90K high-quality multi-modal instances, which is about 45\% of the initially generated data. Each example thereby provides a multilingual visual context and a linked entity label, serving as valuable weak supervision for entity recognition and linking. This M3LS-derived resource augments our training data and helps the model learn to recognize entities in multiple languages and modalities.

The full prompt used for generating entity data is shown in \autoref{fig:m3ls_prompt}.

\begin{figure*}[t]
\definecolor{highlightcolor}{RGB}{46, 134, 171}
\definecolor{framecolor}{RGB}{162, 59, 114}
\resizebox{\textwidth}{!}{%
\begin{tcolorbox}[
  colback=highlightcolor!15, 
  colframe=framecolor, 
  boxrule=1.0pt, 
  title=Prompt for Entity Extraction from Multilingual News Articles,
  label=entity-extraction-template,
  breakable,
  width=\textwidth
]

\textbf{System Message:} \\
 You are a specialized entity extraction AI assistant that identifies the most important entities in news articles across multiple languages. Your task is to analyze news content and extract the central entity, providing both its name in the original language and its standard English Wikipedia title. \\

\textbf{User Prompt:} \\
Below is a news article in \{LANGUAGE\}. Analyze it and identify the most relevant entity mentioned. \\

\textbf{Article Information:} \\
\textit{Article Title:} \{title\} \\
\textit{Article Summary:} \{summary\} \\
\textit{Article First Paragraph:} \{first\_paragraph\} \\
\textit{Image Captions:} \{image\_captions\} (if available) \\
\textit{Keywords:} \{keywords\} (if available) \\

\textbf{Please extract the following information:} \\
1. The most relevant entity mentioned in the article (in \{LANGUAGE\}) \\
2. The Wikipedia page title for this entity (in English) \\
3. A brief justification for why this is the most relevant entity \\

\textbf{Format your response as JSON with the following keys:} \\
- entity\_mention: The entity name in \{LANGUAGE\} \\
- wikipedia\_title: The English Wikipedia title for this entity \\
- justification: Your explanation \\

\textbf{Examples of entity extraction from different languages:} \\
\textit{Example 1 (Vietnamese):} \\
Article Title: Amabie, 'bua yem' chong Covid-19 cua nguoi Nhat \\
Entity Name: Covid-19 \\
English Wikipedia Title: COVID-19 \\

[MORE EXAMPLES...]

\textbf{Additional Guidelines:} \\
If there are multiple important entities, choose the most central one to the article.

\end{tcolorbox}
}
\caption{Entity Extraction Prompt}
\label{fig:m3ls_prompt}
\end{figure*}

\section{Breakdown Results}

\subsection{MaRVL}
We show the performance of different models on the MaRVL benchmark in \autoref{tab:breakdown:MaRVL}.
% \begin{table*}[t]
% \centering
% \small
% \begin{tabular}{@{}l|ccccccc@{}}
% \toprule
% \textbf{Model} & \textbf{English} & \textbf{Multi} & \textbf{Indonesian} & \textbf{Swahili} & \textbf{Tamil} & \textbf{Turkish} & \textbf{Chinese} \\
% \midrule
% Llava-Next-7B      & 62.8 & 50.9 & 52.2 & 50.6 & 50.5 & 50.4 & 50.6 \\
% Molmo-7B-D         & 65.3 & 54.9 & 61.1 & 49.6 & 49.6 & 52.2 & 62.2 \\
% Llama3.2-11B       & 64.5 & 58.1 & 62.7 & 52.4 & 54.0 & 61.6 & 59.5 \\
% PaliGemma-3B       & 56.5 & 52.2 & 53.4 & 49.6 & 50.5 & 56.3 & 51.3 \\
% mBLIP mT0-XL       & 67.3 & 66.7 & 64.9 & 64.8 & 69.7 & 68.1 & 65.9 \\
% AyaVision-8B       & 73.8 & 64.5 & 66.6 & 53.1 & 59.7 & 73.7 & 69.3 \\
% Pangea-7B          & 87.0 & 78.7 & 80.8 & 74.2 & 69.8 & 84.6 & 84.1 \\
% \model   & 89.0 & 80.3 & 83.6 & 74.6 & 73.2 & 85.4 & 84.5 \\
% \bottomrule
% \end{tabular}
% \caption{Performance of selected models on the MaRVL benchmark across different languages.}
% \label{tab:breakdown:MaRVL}
% \end{table*}

\begin{table*}[t]
\centering
\small
\begin{adjustbox}{width=\textwidth}
\begin{tabular}{l@{}*{7}{>{\centering\arraybackslash}p{1.8cm}@{}}}
\toprule
\textbf{Model} & \textbf{English} & \textbf{Multi} & \textbf{Indonesian} & \textbf{Swahili} & \textbf{Tamil} & \textbf{Turkish} & \textbf{Chinese} \\
\midrule
Llava-Next-7B      & 62.8 & 50.9 & 52.2 & 50.6 & 50.5 & 50.4 & 50.6 \\
Molmo-7B-D         & 65.3 & 54.9 & 61.1 & 49.6 & 49.6 & 52.2 & 62.2 \\
Llama3.2-11B       & 64.5 & 58.1 & 62.7 & 52.4 & 54.0 & 61.6 & 59.5 \\
PaliGemma-3B       & 56.5 & 52.2 & 53.4 & 49.6 & 50.5 & 56.3 & 51.3 \\
mBLIP mT0-XL       & 67.3 & 66.7 & 64.9 & 64.8 & 69.7 & 68.1 & 65.9 \\
AyaVision-8B       & 73.8 & 64.5 & 66.6 & 53.1 & 59.7 & 73.7 & 69.3 \\
Pangea-7B          & 87.0 & 78.7 & 80.8 & 74.2 & 69.8 & 84.6 & 84.1 \\
\textbf{CulturalPangea-7B}   & \textbf{89.0} & \textbf{80.3} & \textbf{83.6} & \textbf{74.6} & \textbf{73.2} & \textbf{85.4} & \textbf{84.5} \\
\bottomrule
\end{tabular}
\end{adjustbox}
\caption{Performance of selected models on the MaRVL benchmark across different languages.}
\label{tab:breakdown:MaRVL}
\end{table*}
\subsection{M3Exam}
We show the performance of different models on the M3Exam benchmark in \autoref{tab:breakdown:m3exam}.
% \begin{table*}[t]
% \centering
% % \resizebox{\linewidth}{!}{%
% \small
% \begin{tabular}{@{}l|cccccccc@{}}
% \toprule
% \textbf{Model}         & \textbf{English} & \textbf{Multi} & \textbf{Afrikaans} & \textbf{Chinese} & \textbf{Italian} & \textbf{Portuguese} & \textbf{Thai} & \textbf{Vietnamese} \\
% \midrule
% Llava-Next-7B          & 36.5 & 28.4 & 28.2 & 25.4 & 37.8 & 27.0 & 23.7 & 28.4 \\
% Molmo-7B-D             & 57.1 & 39.1 & 35.6 & 56.4 & 49.4 & 40.2 & 27.4 & 25.9 \\
% Llama3.2-11B           & 51.8 & 36.6 & 42.3 & 46.4 & 45.8 & 28.4 & 26.4 & 30.2 \\
% PaliGemma-3B           & 36.0 & 25.6 & 26.4 & 24.7 & 32.2 & 24.3 & 27.2 & 19.0 \\
% mBLIP mT0-XL           & 22.8 & 25.0 & 16.0 & 25.6 & 33.7 & 21.2 & 22.4 & 31.0 \\
% AyaVision-8B           & 56.2 & 41.7 & 47.2 & 49.2 & 55.4 & 37.8 & 30.4 & 30.2 \\
% Pangea-7B              & 61.4 & 42.0 & 52.1 & 49.2 & 54.9 & 43.3 & 32.9 & 19.8 \\
% \model       & 58.0 & 46.7 & 52.8 & 55.2 & 57.9 & 46.0 & 34.9 & 33.6 \\
% \bottomrule
% \end{tabular}%
% % }

% \caption{Performance of selected models on the M3Exam dataset across different languages.}
% \label{tab:breakdown:m3exam}
% \end{table*}

\begin{table*}[t]
\centering
\small
\begin{adjustbox}{width=\textwidth}
\begin{tabular}{l@{}*{8}{>{\centering\arraybackslash}p{1.8cm}@{}}}
\toprule
\textbf{Model}         & \textbf{English} & \textbf{Multi} & \textbf{Afrikaans} & \textbf{Chinese} & \textbf{Italian} & \textbf{Portuguese} & \textbf{Thai} & \textbf{Vietnamese} \\
\midrule
Llava-Next-7B          & 36.5 & 28.4 & 28.2 & 25.4 & 37.8 & 27.0 & 23.7 & 28.4 \\
Molmo-7B-D             & 57.1 & 39.1 & 35.6 & 56.4 & 49.4 & 40.2 & 27.4 & 25.9 \\
Llama3.2-11B           & 51.8 & 36.6 & 42.3 & 46.4 & 45.8 & 28.4 & 26.4 & 30.2 \\
PaliGemma-3B           & 36.0 & 25.6 & 26.4 & 24.7 & 32.2 & 24.3 & 27.2 & 19.0 \\
mBLIP mT0-XL           & 22.8 & 25.0 & 16.0 & 25.6 & 33.7 & 21.2 & 22.4 & 31.0 \\
AyaVision-8B           & 56.2 & 41.7 & 47.2 & 49.2 & 55.4 & 37.8 & 30.4 & 30.2 \\
Pangea-7B              & 61.4 & 42.0 & 52.1 & 49.2 & 54.9 & 43.3 & 32.9 & 19.8 \\
\textbf{CulturalPangea-7B}      & \textbf{58.0} & \textbf{46.7} & \textbf{52.8} & \textbf{55.2} & \textbf{57.9} & \textbf{46.0} & \textbf{34.9} & \textbf{33.6} \\
\bottomrule
\end{tabular}
\end{adjustbox}
\caption{Performance of selected models on the M3Exam dataset across different languages.}
\label{tab:breakdown:m3exam}
\end{table*}
\subsection{MAXM}
% We show the performance of different models on the MAXM benchmark in \autoref{tab:breakdown:MAXM}.
% \begin{table*}[t]
% \centering
% \small
% \begin{tabular}{@{}l|cccccccc@{}}
% \toprule
% \textbf{Model}         & \textbf{English} & \textbf{Multi} & \textbf{French} & \textbf{Hindi} & \textbf{Hebrew} & \textbf{Romanian} & \textbf{Thai} & \textbf{Chinese} \\
% \midrule
% Llava-Next-7B          & 54.9 & 21.4 & 33.7 & 16.2 & 10.7 & 15.5 & 18.3 & 33.9 \\
% Molmo-7B-D             & 52.9 & 37.5 & 45.5 & 33.5 & 30.7 & 28.9 & 46.3 & 40.4 \\
% Llama3.2-11B           & 55.3 & 43.9 & 48.1 & 50.4 & 41.8 & 36.6 & 56.7 & 30.0 \\
% PaliGemma-3B           & 47.9 & 19.9 &  8.0 & 36.5 & 19.3 & 13.4 & 31.3 & 10.8 \\
% mBLIP mT0-XL           & 44.7 & 36.8 & 36.0 & 42.7 & 28.9 & 30.3 & 56.3 & 26.4 \\
% AyaVision-8B           & 49.4 & 52.1 & 56.1 & 63.8 & 57.5 & 50.7 & 34.7 & 49.8 \\
% Pangea-7B              & 53.5 & 53.3 & 43.9 & 53.5 & 59.3 & 45.8 & 67.2 & 50.2 \\
% \model       & 55.3 & 53.9 & 45.5 & 50.4 & 62.9 & 48.6 & 68.3 & 48.0 \\
% \bottomrule
% \end{tabular}
% \caption{Performance of selected models on the MAXM dataset across different languages.}
% \label{tab:breakdown:MAXM}
% \end{table*}
We show the performance of different models on the MAXM benchmark in \autoref{tab:breakdown:MAXM}.
\begin{table*}[t]
\centering
\small
\begin{adjustbox}{width=\textwidth}
\begin{tabular}{l@{}*{8}{>{\centering\arraybackslash}p{1.8cm}@{}}}
\toprule
\textbf{Model}         & \textbf{English} & \textbf{Multi} & \textbf{French} & \textbf{Hindi} & \textbf{Hebrew} & \textbf{Romanian} & \textbf{Thai} & \textbf{Chinese} \\
\midrule
Llava-Next-7B          & 54.9 & 21.4 & 33.7 & 16.2 & 10.7 & 15.5 & 18.3 & 33.9 \\
Molmo-7B-D             & 52.9 & 37.5 & 45.5 & 33.5 & 30.7 & 28.9 & 46.3 & 40.4 \\
Llama3.2-11B           & 55.3 & 43.9 & 48.1 & 50.4 & 41.8 & 36.6 & 56.7 & 30.0 \\
PaliGemma-3B           & 47.9 & 19.9 &  8.0 & 36.5 & 19.3 & 13.4 & 31.3 & 10.8 \\
mBLIP mT0-XL           & 44.7 & 36.8 & 36.0 & 42.7 & 28.9 & 30.3 & 56.3 & 26.4 \\
AyaVision-8B           & 49.4 & 52.1 & 56.1 & 63.8 & 57.5 & 50.7 & 34.7 & 49.8 \\
Pangea-7B              & 53.5 & 53.3 & 43.9 & 53.5 & 59.3 & 45.8 & 67.2 & 50.2 \\
\textbf{CulturalPangea-7B}       & \textbf{55.3} & \textbf{53.9} & \textbf{45.5} & \textbf{50.4} & \textbf{62.9} & \textbf{48.6} & \textbf{68.3} & \textbf{48.0} \\
\bottomrule
\end{tabular}
\end{adjustbox}
\caption{Performance of selected models on the MAXM dataset across different languages.}
\label{tab:breakdown:MAXM}
\end{table*}
\subsection{MERLIN}
We show the performance of different models on the MERLIN benchmark in \autoref{tab:all_merlin_results}.

\begin{table*}[!t]
  \centering
  \scriptsize
  \setlength{\tabcolsep}{3pt}
  \begin{adjustbox}{width=\textwidth}
    \begin{tabular}{l@{}*{6}{>{\centering\arraybackslash}p{1.8cm}@{}}}
      \toprule
      Models & Multi & Hindi & Tamil & Indonesian & Japanese & Vietnamese \\
      \midrule
      
      \multicolumn{7}{c}{\textbf{Exact Match}} \\
      \midrule
      Llava-Next-7B & 34.1 & 30.1 & 13.0 & 42.4 & 41.9 & 43.1 \\
      Molmo-7B-D & 42.9 & 30.8 & 15.8 & 54.7 & 53.8 & 59.5 \\
      PaliGemma-3B & 13.1 & 12.6 & 2.5 & 22.0 & 13.7 & 15.0 \\
      mBLIP-mT0-XL & 15.8 & 10.8 & 11.0 & 19.2 & 14.9 & 23.3 \\
      AyaVision-8B & 55.3 & 55.4 & 40.5 & 55.3 & 62.6 & 62.9 \\
      Pangea-7B & 66.5 & 67.6 & 55.2 & 68.8 & 73.1 & 67.6 \\
      \textbf{CulturalPangea-7B} & \textbf{81.1} & \textbf{77.0} & \textbf{76.2} & \textbf{80.6} & \textbf{85.8} & \textbf{85.7} \\
      \bottomrule
    \end{tabular}
  \end{adjustbox}
  
  \vspace{0.5cm}
  
  \begin{adjustbox}{width=\textwidth}
    \begin{tabular}{l@{}*{6}{>{\centering\arraybackslash}p{1.8cm}@{}}}
      \toprule
      Models & Multi & Hindi & Tamil & Indonesian & Japanese & Vietnamese \\
      \midrule
      
      \multicolumn{7}{c}{\textbf{Exact Match + Alias}} \\
      \midrule
      Llava-Next-7B & 38.8 & 33.0 & 14.7 & 50.0 & 46.3 & 50.1 \\
      Molmo-7B-D & 47.8 & 33.2 & 17.2 & 63.3 & 59.3 & 66.0 \\
      PaliGemma-3B & 19.1 & 15.0 & 2.6 & 31.9 & 20.4 & 25.8 \\
      mBLIP-mT0-XL & 20.1 & 14.2 & 12.9 & 26.5 & 20.9 & 26.0 \\
      AyaVision-8B & 57.5 & 56.9 & 41.3 & 59.0 & 65.1 & 65.0 \\
      Pangea-7B & 71.6 & 71.2 & 58.9 & 77.4 & 79.5 & 71.2 \\
      \textbf{CulturalPangea-7B} & \textbf{84.8} & \textbf{79.3} & \textbf{78.9} & \textbf{86.6} & \textbf{90.1} & \textbf{89.3} \\
      \bottomrule
    \end{tabular}
  \end{adjustbox}
  
  \vspace{0.5cm}
  
  \begin{adjustbox}{width=\textwidth}
    \begin{tabular}{l@{}*{6}{>{\centering\arraybackslash}p{1.8cm}@{}}}
      \toprule
      Models & Multi & Hindi & Tamil & Indonesian & Japanese & Vietnamese \\
      \midrule
      
      \multicolumn{7}{c}{\textbf{Exact Match + Target in Prediction}} \\
      \midrule
      Llava-Next-7B & 50.7 & 48.5 & 22.3 & 67.8 & 57.2 & 57.7 \\
      Molmo-7B-D & 56.5 & 40.6 & 23.4 & 73.5 & 69.2 & 75.9 \\
      PaliGemma-3B & 16.1 & 17.8 & 4.3 & 25.3 & 15.4 & 17.8 \\
      mBLIP-mT0-XL & 17.5 & 12.5 & 12.7 & 20.9 & 16.8 & 24.5 \\
      AyaVision-8B & 69.9 & 66.3 & 58.9 & 73.3 & 73.9 & 77.3 \\
      Pangea-7B & 73.1 & 73.3 & 61.1 & 79.1 & 78.7 & 73.3 \\
      \textbf{CulturalPangea-7B} & \textbf{83.5} & \textbf{81.2} & \textbf{78.9} & \textbf{83.7} & \textbf{87.0} & \textbf{86.9} \\
      \bottomrule
    \end{tabular}
  \end{adjustbox}
  
  \vspace{0.5cm}
  
  \begin{adjustbox}{width=\textwidth}
    \begin{tabular}{l@{}*{6}{>{\centering\arraybackslash}p{1.8cm}@{}}}
      \toprule
      Models & Multi & Hindi & Tamil & Indonesian & Japanese & Vietnamese \\
      \midrule
      
      \multicolumn{7}{c}{\textbf{Exact Match + Alias + Target in Prediction}} \\
      \midrule
      Llava-Next-7B & 60.4 & 57.0 & 31.8 & 77.6 & 66.1 & 69.3 \\
      Molmo-7B-D & 66.5 & 52.3 & 32.9 & 85.2 & 77.6 & 84.7 \\
      PaliGemma-3B & 25.1 & 23.4 & 5.7 & 40.3 & 25.1 & 31.1 \\
      mBLIP-mT0-XL & 23.4 & 17.1 & 16.4 & 30.0 & 24.5 & 28.9 \\
      AyaVision-8B & 78.3 & 73.4 & 68.2 & 83.4 & 82.1 & 84.3 \\
      Pangea-7B & 80.8 & 80.1 & 68.6 & 88.7 & 86.3 & 80.1 \\
      \textbf{CulturalPangea-7B} & \textbf{88.4} & \textbf{84.0} & \textbf{84.4} & \textbf{91.0} & \textbf{91.7} & \textbf{91.0} \\
      \bottomrule
    \end{tabular}
  \end{adjustbox}
  
\caption{Performance results across all models and evaluation metrics on cultural vision tasks across five languages. Four evaluation methods are used with increasing compassion: Exact Match requires predictions to exactly match the target title; +Alias also accepts predictions matching any alias of the entity(alias names from Wikidata); +Target also accepts cases where the target appears anywhere in the prediction; and All Methods combines exact match, alias matching, and target-in-prediction. Results demonstrate that all models benefit significantly from more lenient evaluation criteria, as exact matching requires precise title recall while alias and target-in-prediction methods better capture semantic understanding. CulturalPangea-7B consistently achieves the highest performance across all metrics and languages.}
  \label{tab:all_merlin_results}
\end{table*}
% \input{latex/tables/merlin}
% ========================================
% OPTION 1: Simple separate tables with spacing (RECOMMENDED)
% ========================================

\subsection{XM100}
We show the performance of different models on the XM100 benchmark in \autoref{tab:breakdown:XM100}.

\begin{table*}[!t]
  \centering
  \scriptsize

  % Languages 1–12
  \setlength{\tabcolsep}{3pt}
  \begin{adjustbox}{width=\textwidth}
    \begin{tabular}{l@{}*{12}{>{\centering\arraybackslash}p{1.2cm}@{}}}
      \toprule
      Models & \rotatebox{90}{English} & \rotatebox{90}{Multi} & \rotatebox{90}{Arabic} & \rotatebox{90}{Bengali} & \rotatebox{90}{Czech} & \rotatebox{90}{Danish} & \rotatebox{90}{German} & \rotatebox{90}{Greek} & \rotatebox{90}{Spanish} & \rotatebox{90}{Persian} & \rotatebox{90}{Finnish} & \rotatebox{90}{Filipino} \\
      \midrule
      Llava-Next-7B    & 92.4 & 15.5 & 5.0 & 0.0 & 27.3 & 22.6 & 23.8 & 2.4 & 59.6 & 4.8 & 9.3 & 6.2 \\
      Molmo-7B-D       & 4.7  & 6.0  & 5.5 & 2.7 & 2.8  & 7.6  & 9.5  & 2.6 & 8.3  & 7.2 & 2.0 & 2.4 \\
      Llama3.2-11B     & 23.1 & 5.8  & 0.0 & 0.0 & 0.2  & 15.3 & 4.4  & 0.5 & 14.9 & 0.0 & 2.3 & 11.0 \\
      PaliGemma-3B     & 24.7 & 0.6  & 0.0 & 0.0 & 0.6  & 1.0  & 3.1  & 0.0 & 0.5  & 0.0 & 0.0 & 0.1 \\
      mBLIP mT0-XL     & 101.0& 6.8  & 7.7 & 2.7 & 11.2 & 6.9  & 5.8  & 7.7 & 19.0 & 11.5& 4.7 & 5.9 \\
      AyaVision-8B     & 10.3 & 10.0 & 7.8 & 7.7 & 9.3  & 7.9  & 11.7 & 6.0 & 14.5 & 15.5& 2.0 & 2.5 \\
      Pangea-7B        & 93.2 & 29.7 & 39.6& 25.1& 27.5 & 32.4 & 42.7 & 11.9& 96.5 & 34.7& 8.6 & 13.8 \\
      \model & 90.8 & 36.9 & 41.2& 29.6& 39.4 & 30.9 & 51.7 & 21.3& 82.9 & 38.5& 3.5 & 10.0 \\
      \bottomrule
    \end{tabular}
  \end{adjustbox}
  
  \vspace{1.5em} % Space between table parts
  
  % Languages 13–25
  \setlength{\tabcolsep}{2.5pt}
  \begin{adjustbox}{width=\textwidth}
    \begin{tabular}{l@{}*{13}{>{\centering\arraybackslash}p{1.15cm}@{}}}
      \toprule
      Models & \rotatebox{90}{French} & \rotatebox{90}{Hebrew} & \rotatebox{90}{Hindi} & \rotatebox{90}{Croatian} & \rotatebox{90}{Hungarian} & \rotatebox{90}{Indonesian} & \rotatebox{90}{Italian} & \rotatebox{90}{Japanese} & \rotatebox{90}{Korean} & \rotatebox{90}{Maori} & \rotatebox{90}{Dutch} & \rotatebox{90}{Norwegian} & \rotatebox{90}{Polish} \\
      \midrule
      Llava-Next-7B    & 61.9 & 3.0  & 3.9  & 7.4  & 16.7  & 20.5      & 36.3    & 1.1      & 10.1   & 2.4     & 50.7   & 27.2      & 21.8       \\
      Molmo-7B-D       & 18.5 & 10.6 & 2.5  & 2.2  & 1.5   & 25.8      & 12.5    & 1.2      & 4.5    & 1.0     & 8.5    & 5.6       & 3.7        \\
      Llama3.2-11B     & 20.0 & 0.0  & 0.0  & 0.6  & 18.3  & 1.4       & 22.1    & 0.0      & 0.0    & 2.3     & 35.6   & 0.6       & 1.1        \\
      PaliGemma-3B     & 1.3  & 0.0  & 0.0  & 0.3  & 0.4   & 0.3       & 0.4     & 0.0      & 0.0    & 2.3     & 3.1    & 0.5       & 0.6        \\
      mBLIP mT0-XL     & 10.9 & 9.4  & 2.0  & 2.2  & 7.6   & 8.9       & 4.8     & 1.1      & 4.5    & 4.2     & 11.1   & 7.2       & 11.2       \\
      AyaVision-8B     & 15.1 & 20.4 & 3.5  & 4.3  & 2.5   & 27.0      & 19.0    & 3.6      & 11.4   & 0.2     & 12.0   & 5.9       & 13.2       \\
      Pangea-7B        & 71.0 & 33.3 & 30.4 & 13.9 & 9.6   & 64.6      & 54.4    & 0.5      & 20.4   & 0.6     & 56.3   & 53.3      & 33.8       \\
      \model & 77.7 & 42.5 & 28.4 & 14.2 & 6.1   & 57.7      & 58.3    & 0.7      & 23.1   & 0.6     & 63.6   & 58.3      & 48.0       \\
      \bottomrule
    \end{tabular}
  \end{adjustbox}
  
  \vspace{1.5em} % Space between table parts
  
  % Languages 26–37
  \setlength{\tabcolsep}{3pt}
  \begin{adjustbox}{width=\textwidth}
    \begin{tabular}{l@{}*{12}{>{\centering\arraybackslash}p{1.2cm}@{}}}
      \toprule
      Models & \rotatebox{90}{Portuguese} & \rotatebox{90}{Quechua} & \rotatebox{90}{Romanian} & \rotatebox{90}{Russian} & \rotatebox{90}{Swedish} & \rotatebox{90}{Swahili} & \rotatebox{90}{Telugu} & \rotatebox{90}{Thai} & \rotatebox{90}{Turkish} & \rotatebox{90}{Ukrainian} & \rotatebox{90}{Vietnamese} & \rotatebox{90}{Chinese} \\
      \midrule
      Llava-Next-7B    & 48.2 & 0.0  & 14.9 & 20.2 & 31.9 & 0.6  & 0.0  & 0.0 & 0.0  & 0.1 & 0.0  & 1.1  \\
      Molmo-7B-D       & 8.5  & 0.2  & 5.4  & 12.5 & 4.9  & 0.0  & 0.4  & 0.0 & 3.8  & 3.8 & 20.5 & 0.0  \\
      Llama3.2-11B     & 24.5 & 0.0  & 10.5 & 0.5  & 9.1  & 6.5  & 0.0  & 0.0 & 0.0  & 0.0 & 0.0  & 0.0  \\
      PaliGemma-3B     & 2.8  & 0.1  & 1.1  & 0.0  & 2.0  & 0.0  & 0.0  & 0.9 & 0.0  & 0.0 & 0.4  & 0.0  \\
      mBLIP mT0-XL     & 10.8 & 0.6  & 4.3  & 7.9  & 11.0 & 7.5  & 7.4  & 0.0 & 9.9  & 4.0 & 6.6  & 0.0  \\
      AyaVision-8B     & 11.5 & 0.2  & 23.1 & 21.6 & 6.0  & 1.4  & 3.6  & 0.0 & 14.7 & 14.9& 25.2 & 4.0  \\
      Pangea-7B        & 82.6 & 0.0  & 41.7 & 62.0 & 30.8 & 45.8 & 0.0  & 0.0 & 0.0  & 0.0 & 0.0  & 0.9  \\
      \model & 87.5 & 0.8  & 43.4 & 66.1 & 23.0 & 46.8 & 27.6 & 0.0 & 44.3 & 39.9& 84.1 & 0.5  \\
      \bottomrule
    \end{tabular}
  \end{adjustbox}

  \caption{XM100 benchmark performance (\%) of all multimodal models across 37 languages.}
  \label{tab:breakdown:XM100}
\end{table*}

\subsection{CVQA}
We show the performance of different models on the CVQA benchmark in \autoref{tab:breakdown:CVQA}.

\begin{table*}[!t]
  \centering
  \scriptsize
  \setlength{\tabcolsep}{2pt}
  \renewcommand{\arraystretch}{1}

  % First table: Average + first 10 language-country pairs
  \centering
  \begin{adjustbox}{width=\textwidth}
    \begin{tabular}{l@{}*{11}{>{\centering\arraybackslash}p{1.2cm}@{}}}
      \toprule
      Models & \rotatebox{90}{Average} & \rotatebox{90}{Brazil-Portuguese} & \rotatebox{90}{Bulgaria-Bulgarian} & \rotatebox{90}{China-Chinese} & \rotatebox{90}{Ethiopia-Amharic} & \rotatebox{90}{India-Bengali} & \rotatebox{90}{India-Hindi} & \rotatebox{90}{India-Tamil} & \rotatebox{90}{India-Telugu} & \rotatebox{90}{India-Urdu} & \rotatebox{90}{Indonesia-Indonesian} \\
      \midrule
      Llava-Next-7B    & 41.3 & 62.3 & 41.5 & 51.1 & 29.5 & 31.1 & N/A & 28.8 & 28.0 & N/A & 42.2 \\
      Molmo-7B-D       & 58.8 & 69.0 & 54.9 & 66.2 & 58.1 & 61.9 & 51.7 & 61.2 & 58.5 & 50.5 & 52.9 \\
      Llama3.2-11B     & 69.6 & 74.6 & 64.2 & 73.6 & 68.4 & 76.9 & 68.2 & 80.4 & 80.5 & 54.6 & 65.8 \\
      PaliGemma-3B     & 43.3 & 53.9 & 39.1 & 53.7 & 24.8 & 46.2 & N/A & 46.0 & 43.5 & N/A & 45.4 \\
      mBLIP mT0-XL     & 37.7 & 44.4 & 38.0 & 39.9 & 35.9 & 36.4 & N/A & 44.2 & 39.0 & N/A & 37.4 \\
      Aya-Vision-8B    & 51.8 & 66.8 & 44.7 & 65.2 & 29.4 & 49.8 & 62.7 & 44.4 & 47.5 & 47.3 & 56.7 \\
      Pangea-7B        & 57.5 & 72.9 & 53.6 & 74.0 & 35.9 & 59.4 & 74.6 & 51.9 & 54.5 & 59.5 & 62.1 \\
      \model & 59.1 & 71.5 & 56.6 & 70.7 & 39.7 & 59.1 & 74.6 & 54.2 & 58.5 & 66.4 & 61.2 \\
      \bottomrule
    \end{tabular}
  \end{adjustbox}

  \vspace{1.5em}

  % Second table: next 11 language-country pairs
  \centering
  \begin{adjustbox}{width=\textwidth}
    \begin{tabular}{l@{}*{11}{>{\centering\arraybackslash}p{1.2cm}@{}}}
      \toprule
      Models & \rotatebox{90}{Indonesia-Javanese} & \rotatebox{90}{Indonesia-Sundanese} & \rotatebox{90}{Ireland-Irish} & \rotatebox{90}{Japan-Japanese} & \rotatebox{90}{Kenya-Swahili} & \rotatebox{90}{Malaysia-Malay} & \rotatebox{90}{Mexico-Spanish} & \rotatebox{90}{Mongolia-Mongolian} & \rotatebox{90}{Nigeria-Igbo} & \rotatebox{90}{Norway-Norwegian} & \rotatebox{90}{Pakistan-Urdu} \\
      \midrule
      Llava-Next-7B    & 38.7 & 35.5 & 42.6 & 32.5 & 46.2 & 45.7 & 51.4 & 33.3 & 35.0 & 56.9 & 36.6 \\
      Molmo-7B-D       & 53.9 & 55.0 & 64.4 & 42.9 & 73.3 & 54.6 & 53.6 & 51.9 & 53.0 & 54.8 & 67.1 \\
      Llama3.2-11B     & 60.6 & 64.0 & 76.4 & 54.2 & 79.1 & 72.1 & 66.6 & 54.5 & 61.5 & 66.9 & 78.7 \\
      PaliGemma-3B     & 41.4 & 33.0 & 34.4 & 43.3 & 44.0 & 44.1 & 47.4 & 29.2 & 32.0 & 52.2 & 44.9 \\
      mBLIP mT0-XL     & 37.4 & 31.0 & 35.3 & 30.0 & 45.1 & 40.6 & 44.9 & 29.2 & 30.5 & 42.8 & 40.3 \\
      Aya-Vision-8B    & 48.2 & 46.5 & 47.2 & 48.3 & 55.0 & 57.0 & 57.6 & 28.5 & 34.7 & 53.2 & 50.9 \\
      Pangea-7B        & 49.2 & 53.0 & 56.4 & 48.3 & 64.1 & 59.7 & 61.9 & 42.0 & 46.5 & 64.2 & 66.2 \\
      \model & 52.9 & 58.0 & 54.9 & 52.2 & 67.0 & 58.4 & 56.3 & 42.6 & 43.5 & 62.9 & 71.3 \\
      \bottomrule
    \end{tabular}
  \end{adjustbox}

  \vspace{1.5em}

  % Third table: remaining 10 language-country pairs
  \centering
  \begin{adjustbox}{width=\textwidth}
    \begin{tabular}{l@{}*{10}{>{\centering\arraybackslash}p{1.2cm}@{}}}
      \toprule
      Models & \rotatebox{90}{Romania-Romanian} & \rotatebox{90}{Russia-Russian} & \rotatebox{90}{Singapore-Chinese} & \rotatebox{90}{South Korea-Korean} & \rotatebox{90}{Spain-Spanish} & \rotatebox{90}{Sri\_Lanka-Sinhala} & \rotatebox{90}{Indonesia-Minangkabau} & \rotatebox{90}{Egypt-Egyptian Arabic} & \rotatebox{90}{France-Breton} & \rotatebox{90}{India-Marathi} \\
      \midrule
      Llava-Next-7B    & 52.3 & 53.5 & 44.8 & 43.4 & 63.5 & 29.8 & 40.2 & 33.5 & 27.4 & N/A \\
      Molmo-7B-D       & 63.6 & 61.5 & 69.3 & 65.2 & 70.1 & 68.0 & 54.6 & 56.7 & 44.2 & N/A \\
      Llama3.2-11B     & 76.8 & 74.5 & 80.7 & 73.8 & 81.4 & 72.4 & 68.9 & 68.5 & 49.4 & N/A \\
      PaliGemma-3B     & 50.3 & 53.5 & 48.6 & 61.0 & 60.1 & 31.6 & 39.8 & 40.4 & 29.9 & N/A \\
      mBLIP mT0-XL     & 43.7 & 42.0 & 36.8 & 38.3 & 53.5 & 31.1 & 34.7 & 31.0 & 23.5 & N/A \\
      Aya-Vision-8B    & 62.8 & 66.3 & 66.8 & 74.4 & 74.5 & 28.9 & 52.4 & 51.5 & 34.4 & 50.8 \\
      Pangea-7B        & 64.2 & 74.0 & 66.0 & 70.3 & 72.6 & 39.1 & 47.4 & 49.3 & 34.3 & 56.4 \\
      \model & 69.2 & 70.0 & 74.1 & 69.0 & 73.6 & 39.1 & 53.8 & 52.7 & 37.5 & 59.4 \\
      \bottomrule
    \end{tabular}
  \end{adjustbox}

  \caption{CVQA benchmark performance (\%) of multimodal models across 31 language-country pairs in local-languages.}
  \label{tab:breakdown:CVQA}
\end{table*}

\begin{table*}[]
  \centering
  \scriptsize
  \setlength{\tabcolsep}{2pt}
  \renewcommand{\arraystretch}{1}

  % First table: Average + first 10 language-country pairs
  \centering
  \begin{adjustbox}{width=\textwidth}
    \begin{tabular}{l@{}*{11}{>{\centering\arraybackslash}p{1.2cm}@{}}}
      \toprule
      Models & \rotatebox{90}{Average} & \rotatebox{90}{Brazil-Portuguese} & \rotatebox{90}{Bulgaria-Bulgarian} & \rotatebox{90}{China-Chinese} & \rotatebox{90}{Ethiopia-Amharic} & \rotatebox{90}{India-Bengali} & \rotatebox{90}{India-Hindi} & \rotatebox{90}{India-Tamil} & \rotatebox{90}{India-Telugu} & \rotatebox{90}{India-Urdu} & \rotatebox{90}{Indonesia-Indonesian} \\
      \midrule
      Llava-Next-7B    & 56.1 & 61.3 & 50.9 & 58.8 & 52.9 & 60.8 & N/A  & 61.4 & 60.5 & N/A  & 48.5 \\
      Molmo-7B-D       & 47.7 & 65.8 & 45.6 & 68.5 & 31.6 & 47.9 & N/A  & 36.4 & 41.5 & N/A  & 50.5 \\
      Llama3.2-11B     & 61.0 & 72.9 & 54.4 & 72.0 & 41.9 & 62.9 & N/A  & 66.4 & 66.5 & N/A  & 63.6 \\
      PaliGemma-3B     & 54.2 & 59.5 & 49.3 & 54.9 & 52.6 & 59.1 & N/A  & 66.0 & 62.5 & N/A  & 49.3 \\
      mBLIP mT0-XL     & 40.8 & 45.4 & 39.1 & 43.7 & 34.2 & 43.0 & N/A  & 46.0 & 41.0 & N/A  & 38.1 \\
      Aya-Vision-8B    & 61.8 & 68.7 & 54.7 & 63.0 & 55.6 & 62.6 & 73.1 & 69.2 & 69.0 & 69.1 & 56.3 \\
      Pangea-7B        & 64.9 & 72.9 & 60.1 & 67.2 & 60.7 & 67.1 & 76.1 & 71.0 & 68.0 & 72.3 & 60.4 \\
      \model & 66.2 & 71.1 & 60.4 & 62.7 & 63.2 & 69.9 & 76.1 & 72.0 & 70.5 & 73.2 & 65.0 \\
      \bottomrule
    \end{tabular}
  \end{adjustbox}

  \vspace{1.5em}

  % Second table: next 11 language-country pairs
  \centering
  \begin{adjustbox}{width=\textwidth}
    \begin{tabular}{l@{}*{11}{>{\centering\arraybackslash}p{1.2cm}@{}}}
      \toprule
      Models & \rotatebox{90}{Indonesia-Javanese} & \rotatebox{90}{Indonesia-Sundanese} & \rotatebox{90}{Ireland-Irish} & \rotatebox{90}{Japan-Japanese} & \rotatebox{90}{Kenya-Swahili} & \rotatebox{90}{Malaysia-Malay} & \rotatebox{90}{Mexico-Spanish} & \rotatebox{90}{Mongolia-Mongolian} & \rotatebox{90}{Nigeria-Igbo} & \rotatebox{90}{Norway-Norwegian} & \rotatebox{90}{Pakistan-Urdu} \\
      \midrule
      Llava-Next-7B    & 48.1 & 49.0 & 66.6 & 40.9 & 71.1 & 54.9 & 51.1 & 44.2 & 53.0 & 57.2 & 67.1 \\
      Molmo-7B-D       & 45.1 & 39.5 & 43.6 & 44.8 & 47.6 & 51.7 & 55.1 & 35.9 & 36.0 & 49.2 & 46.8 \\
      Llama3.2-11B     & 48.8 & 54.0 & 57.4 & 58.1 & 61.5 & 69.2 & 64.7 & 41.0 & 39.5 & 65.9 & 65.7 \\
      PaliGemma-3B     & 48.1 & 46.0 & 58.3 & 44.8 & 59.7 & 54.9 & 51.7 & 43.4 & 46.0 & 55.2 & 67.6 \\
      mBLIP mT0-XL     & 39.1 & 32.5 & 37.4 & 34.0 & 50.2 & 41.6 & 34.7 & 33.9 & 39.5 & 43.1 & 45.4 \\
      Aya-Vision-8B    & 54.2 & 53.5 & 62.9 & 50.2 & 72.5 & 61.3 & 57.6 & 47.8 & 57.0 & 57.9 & 70.8 \\
      Pangea-7B        & 57.2 & 56.0 & 72.7 & 45.8 & 77.2 & 62.5 & 61.6 & 52.9 & 59.5 & 64.9 & 72.2 \\
      \model & 57.2 & 61.0 & 72.7 & 50.2 & 80.2 & 65.4 & 60.7 & 52.6 & 57.0 & 64.2 & 75.0 \\
      \bottomrule
    \end{tabular}
  \end{adjustbox}

  \vspace{1.5em}

  % Third table: remaining 10 language-country pairs
  \centering
  \begin{adjustbox}{width=\textwidth}
    \begin{tabular}{l@{}*{10}{>{\centering\arraybackslash}p{1.2cm}@{}}}
      \toprule
      Models & \rotatebox{90}{Romania-Romanian} & \rotatebox{90}{Russia-Russian} & \rotatebox{90}{Singapore-Chinese} & \rotatebox{90}{South Korea-Korean} & \rotatebox{90}{Spain-Spanish} & \rotatebox{90}{Sri\_Lanka-Sinhala} & \rotatebox{90}{Indonesia-Minangkabau} & \rotatebox{90}{Egypt-Egyptian Arabic} & \rotatebox{90}{France-Breton} & \rotatebox{90}{India-Marathi} \\
      \midrule
      Llava-Next-7B    & 62.6 & 58.5 & 62.3 & 60.0 & 67.6 & 59.1 & 51.4 & 54.7 & 37.5 & N/A  \\
      Molmo-7B-D       & 52.0 & 63.5 & 66.0 & 56.9 & 66.7 & 31.6 & 43.4 & 43.8 & 29.6 & N/A  \\
      Llama3.2-11B     & 75.5 & 74.5 & 73.6 & 73.1 & 83.3 & 51.1 & 58.2 & 56.7 & 36.3 & N/A  \\
      PaliGemma-3B     & 60.9 & 56.0 & 59.4 & 58.3 & 61.0 & 62.2 & 43.4 & 51.2 & 37.3 & N/A  \\
      mBLIP mT0-XL     & 43.7 & 41.0 & 43.9 & 41.4 & 51.9 & 48.0 & 38.6 & 42.9 & 30.4 & N/A  \\
      Aya-Vision-8B    & 63.9 & 66.0 & 70.3 & 67.2 & 69.5 & 64.4 & 58.2 & 57.6 & 43.7 & 68.8 \\
      Pangea-7B        & 71.9 & 68.5 & 71.7 & 66.6 & 75.2 & 70.6 & 56.9 & 59.1 & 45.2 & 69.3 \\
      \model & 69.9 & 66.5 & 75.0 & 73.4 & 73.9 & 69.8 & 60.2 & 63.1 & 50.1 & 70.8 \\
      \bottomrule
    \end{tabular}
  \end{adjustbox}

  \caption{CVQA benchmark performance (\%) of multimodal models across 31 language-country pairs in English.}
  \label{tab:breakdown_CVQA_english}
\end{table*}
\subsection{ALMBench}
We show the performance of different models on the ALMBench benchmark in \autoref{tab:breakdown:ALMBench}.

\begin{table*}[t]
  \centering
  \scriptsize
  \captionsetup[subtable]{skip=1ex}

  % First table: English, Multi + first 8 languages (10 columns total)
  \centering
  \setlength{\tabcolsep}{3pt}
  \begin{adjustbox}{width=\textwidth}
    \begin{tabular}{l@{}*{10}{>{\centering\arraybackslash}p{1.2cm}@{}}}
      \toprule
      Models               & \rotatebox{90}{English} & \rotatebox{90}{Multi} & \rotatebox{90}{Amharic} & \rotatebox{90}{Bengali} & \rotatebox{90}{Bulgarian} & \rotatebox{90}{Chinese (Simpl.)} & \rotatebox{90}{Chinese (Trad.)} & \rotatebox{90}{Czech} & \rotatebox{90}{Dutch} & \rotatebox{90}{French} \\
      \midrule
      Llava-Next-7B        & 68.0   &   42.4    & 14.0   & 18.0   & 50.0   & 56.0   & 52.0    & 51.0   & 56.0   & 59.0    \\
      Molmo-7B-D           & 77.0   &   49.1    & 32.0   & 41.0   & 50.0   & 61.0   & 62.0    & 54.0   & 56.0   & 64.0    \\
      Llama3.2-11B         & 78.5   &   56.6    & 17.4   & 57.1   & 58.0   & 64.5   & 53.8    & 63.5   & 65.2   & 72.5    \\
      PaliGemma-3B         & 50.0   &   35.3    & 16.3   & 28.4   & 34.5   & 41.4   & 36.1    & 39.8   & 38.3   & 41.3    \\
      mBLIP mT0-XL         & 72.0   &   36.9    & 21.0   & 22.0   & 39.0   & 49.0   & 40.0    & 35.0   & 43.0   & 65.0    \\
      AyaVision-8B         & 71.3   &   55.1    &  8.8   & 34.3   & 53.2   & 64.8   & 56.9    & 66.4   & 70.7   & 74.3    \\
      Pangea-7B            & 71.5   &  59.9     & 25.7   & 53.2   & 59.8   & 67.3   & 60.9    & 63.2   & 64.0   & 75.9    \\
      \model     & 75.9   &   63.5    & 31.3   & 58.4   & 63.5   & 70.0   & 63.2    & 65.4   & 62.9   & 74.8    \\
      \bottomrule
    \end{tabular}
  \end{adjustbox}

  \vspace{1.5em}

  % Second table: next 10 languages
  \centering
  \setlength{\tabcolsep}{3pt}
  \begin{adjustbox}{width=\textwidth}
    \begin{tabular}{l@{}*{10}{>{\centering\arraybackslash}p{1.2cm}@{}}}
      \toprule
      Models               & \rotatebox{90}{German} & \rotatebox{90}{Greek} & \rotatebox{90}{Hebrew} & \rotatebox{90}{Hindi} & \rotatebox{90}{Igbo} & \rotatebox{90}{Indonesian} & \rotatebox{90}{Irish} & \rotatebox{90}{Italian} & \rotatebox{90}{Japanese} & \rotatebox{90}{Javanese} \\
      \midrule
      Llava-Next-7B        & 60.0    & 28.0   & 32.0    & 30.0   & 37.0   & 57.0      & 21.0   & 64.0   & 47.0   & 47.0    \\
      Molmo-7B-D           & 58.0    & 39.0   & 37.0    & 44.0   & 34.0   & 57.0      & 45.0   & 58.0   & 61.0   & 53.0    \\
      Llama3.2-11B         & 67.4    & 62.9   & 57.0    & 66.6   & 28.8   & 71.8      & 35.6   & 78.5   & 59.0   & 44.9    \\
      PaliGemma-3B         & 45.6    & 40.2   & 30.1    & 34.7   & 14.1   & 42.6      & 20.6   & 45.4   & 39.7   & 34.8    \\
      mBLIP mT0-XL         & 58.0    & 29.0   & 28.0    & 29.0   & 32.0   & 40.0      & 24.0   & 59.0   & 39.0   & 34.0    \\
      AyaVision-8B         & 69.3    & 69.6   & 64.1    & 64.6   & 22.6   & 70.9      & 18.9   & 74.9   & 68.7   & 61.6    \\
      Pangea-7B            & 71.4    & 57.9   & 55.3    & 58.1   & 47.9   & 67.6      & 45.2   & 66.2   & 69.3   & 67.9    \\
      \model     & 73.1    & 59.5   & 66.2    & 62.4   & 51.9   & 70.2      & 54.3   & 76.6   & 71.1   & 69.1    \\
      \bottomrule
    \end{tabular}
  \end{adjustbox}

  \vspace{1.5em}

  % Third table: next 10 languages
  \centering
  \setlength{\tabcolsep}{2.5pt}
  \begin{adjustbox}{width=\textwidth}
    \begin{tabular}{l@{}*{10}{>{\centering\arraybackslash}p{1.15cm}@{}}}
      \toprule
      Models               & \rotatebox{90}{Korean} & \rotatebox{90}{Malay} & \rotatebox{90}{Mongolian} & \rotatebox{90}{Norwegian} & \rotatebox{90}{Persian} & \rotatebox{90}{Polish} & \rotatebox{90}{Portuguese} & \rotatebox{90}{Romanian} & \rotatebox{90}{Russian} & \rotatebox{90}{Sinhala} \\
      \midrule
      Llava-Next-7B        & 41.0   & 57.0   & 28.0    & 57.0     & 33.0     & 57.0    & 62.0       & 52.0   & 51.0   & 18.0    \\
      Molmo-7B-D           & 52.0   & 57.0   & 36.0    & 50.0     & 42.0     & 58.0    & 54.0       & 50.0   & 61.0   & 39.0    \\
      Llama3.2-11B         & 54.4   & 70.8   & 24.8    & 66.2     & 59.0     & 66.3    & 66.3       & 66.7   & 62.2   & 37.0    \\
      PaliGemma-3B         & 41.1   & 39.3   & 16.9    & 41.0     & 37.6     & 38.0    & 41.4       & 40.4   & 32.6   & 13.8    \\
      mBLIP mT0-XL         & 30.0   & 44.0   & 25.0    & 34.0     & 28.0     & 33.0    & 60.0       & 34.0   & 44.0   & 21.0    \\
      AyaVision-8B         & 65.0   & 70.8   & 22.3    & 53.9     & 64.3     & 65.7    & 66.8       & 66.0   & 65.1   & 17.9    \\
      Pangea-7B            & 60.8   & 69.1   & 52.4    & 65.5     & 56.1     & 66.0    & 61.7       & 65.7   & 63.3   & 21.0    \\
      \model     & 63.4   & 72.0   & 57.7    & 66.2     & 62.4     & 67.6    & 64.8       & 65.4   & 64.8   & 36.2    \\
      \bottomrule
    \end{tabular}
  \end{adjustbox}

  \vspace{1.5em}

  % Fourth table: final 10 languages
  \centering
  \setlength{\tabcolsep}{3pt}
  \begin{adjustbox}{width=\textwidth}
    \begin{tabular}{l@{}*{10}{>{\centering\arraybackslash}p{1.2cm}@{}}}
      \toprule
      Models               & \rotatebox{90}{Spanish} & \rotatebox{90}{Sundanese} & \rotatebox{90}{Swahili} & \rotatebox{90}{Tamil} & \rotatebox{90}{Telugu} & \rotatebox{90}{Thai} & \rotatebox{90}{Turkish} & \rotatebox{90}{Ukrainian} & \rotatebox{90}{Urdu} & \rotatebox{90}{Vietnamese} \\
      \midrule
      Llava-Next-7B        & 59.0    & 37.0     & 23.0     & 18.0   & 18.0    & 23.0    & 43.0     & 55.0      & 25.0  & 48.0       \\
      Molmo-7B-D           & 58.0    & 50.0     & 41.0     & 23.0   & 24.0    & 40.0    & 51.0     & 50.0      & 44.0  & 50.0       \\
      Llama3.2-11B         & 66.6    & 39.4     & 41.0     & 44.1   & 51.8    & 59.2    & 64.0     & 63.8      & 54.5  & 64.5       \\
      PaliGemma-3B         & 38.3    & 32.2     & 37.9     & 25.8   & 32.7    & 42.5    & 46.6     & 47.1      & 31.2  & 36.5       \\
      mBLIP mT0-XL         & 60.0    & 38.0     & 33.0     & 26.0   & 30.0    & 26.0    & 31.0     & 28.0      & 28.0  & 27.0       \\
      AyaVision-8B         & 65.0    & 56.1     & 26.4     & 36.6   & 34.4    & 40.6    & 71.9     & 74.0      & 39.9  & 69.2       \\
      Pangea-7B            & 65.4    & 65.8     & 63.1     & 38.2   & 50.9    & 62.3    & 67.2     & 68.3      & 62.0  & 67.3       \\
      \model     & 69.4    & 67.1     & 65.2     & 44.5   & 55.8    & 66.2    & 73.1     & 70.3      & 59.9  & 64.8       \\
      \bottomrule
    \end{tabular}
  \end{adjustbox}

  \caption{ALMBench benchmark performance (\%) of all multimodal models across 39 languages.}
  \label{tab:breakdown:ALMBench}
\end{table*}

\section{CultureGround Examples}
\label{section:examples}

\begin{figure*}[t]
\definecolor{highlightcolor}{RGB}{255, 120, 84}
\definecolor{framecolor}{RGB}{162, 59, 114}

\resizebox{\textwidth}{!}{%
\begin{tcolorbox}[
  colback=highlightcolor!15,
  colframe=framecolor,
  boxrule=1.0pt,
  title=CulturalGround(French),
]

  \adjustbox{valign=t}{
  \begin{minipage}[t]{0.45\textwidth}
    \includegraphics[width=0.8\linewidth]{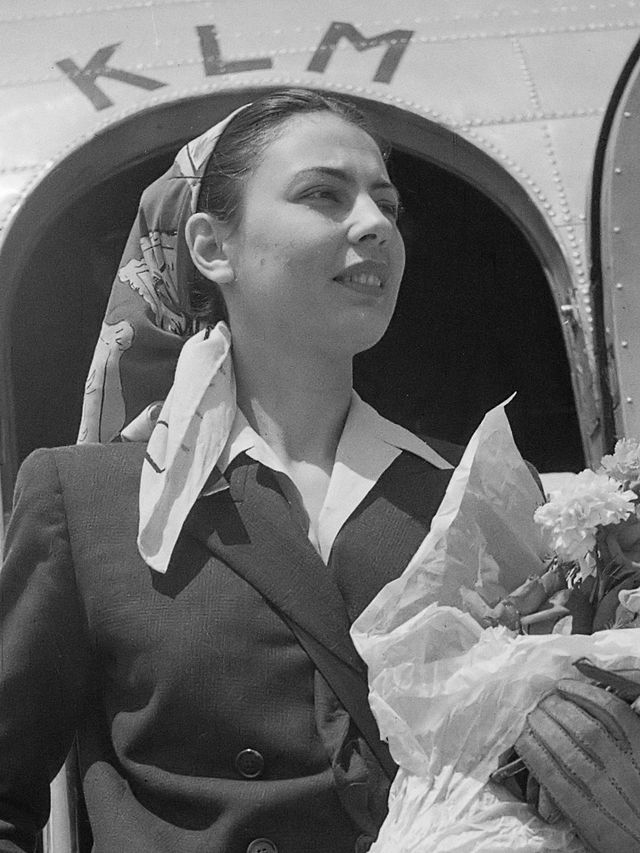} \\
    \footnotesize{Marie Déa}
  \end{minipage}
  }
  \adjustbox{valign=t}{
  \begin{minipage}[t]{0.5\textwidth}
    \textcolor{blue}{\textbf{Entity Info}} \\
    Language: French \\
    Entity name: Marie Déa \\
    Country: France \\
    Wikidata ID: Q3292505 \\
    Question type: Property-level question \\
    Property: P1412 (languages spoken / written) \\
    Template question: Quelle(s) langue(s) cette entité parle-t-elle ou écrit-elle ?(Which language(s) does this entity speak or write) \\
    Template answer: Marie Déa parle ou écrit français, langue romane.(Marie Déa speaks or writes French, a Romance language)
  \end{minipage}
  } \\

  \hrulefill

  \textcolor{blue}{\textbf{Question}} \\
  Quelle langue parle ou écrit l'actrice française que vous voyez sur cette photo ? (Which language does the French actress you see in this photo speak or write?)\\

  \textcolor{blue}{\textbf{Answer}} \\
  Marie Déa parle et écrit en français, la langue romane de France. (Marie Déa speaks and writes French, the Romance language of France.)
\end{tcolorbox}
}
\caption{Sample from in CulturalGround. Template question and answer are created from our curation pipeline and further }
\label{fig:culturalground_samples}
\end{figure*}

\begin{figure*}[t]
\definecolor{highlightcolor}{RGB}{255, 120, 84}
\definecolor{framecolor}{RGB}{162, 59, 114}

\resizebox{\textwidth}{!}{%
\begin{tcolorbox}[
  colback=highlightcolor!15,
  colframe=framecolor,
  boxrule=1.0pt,
  title=CulturalGround(Indonesian),
]

  \adjustbox{valign=t}{
  \begin{minipage}[t]{0.45\textwidth}
    \includegraphics[width=0.8\linewidth]{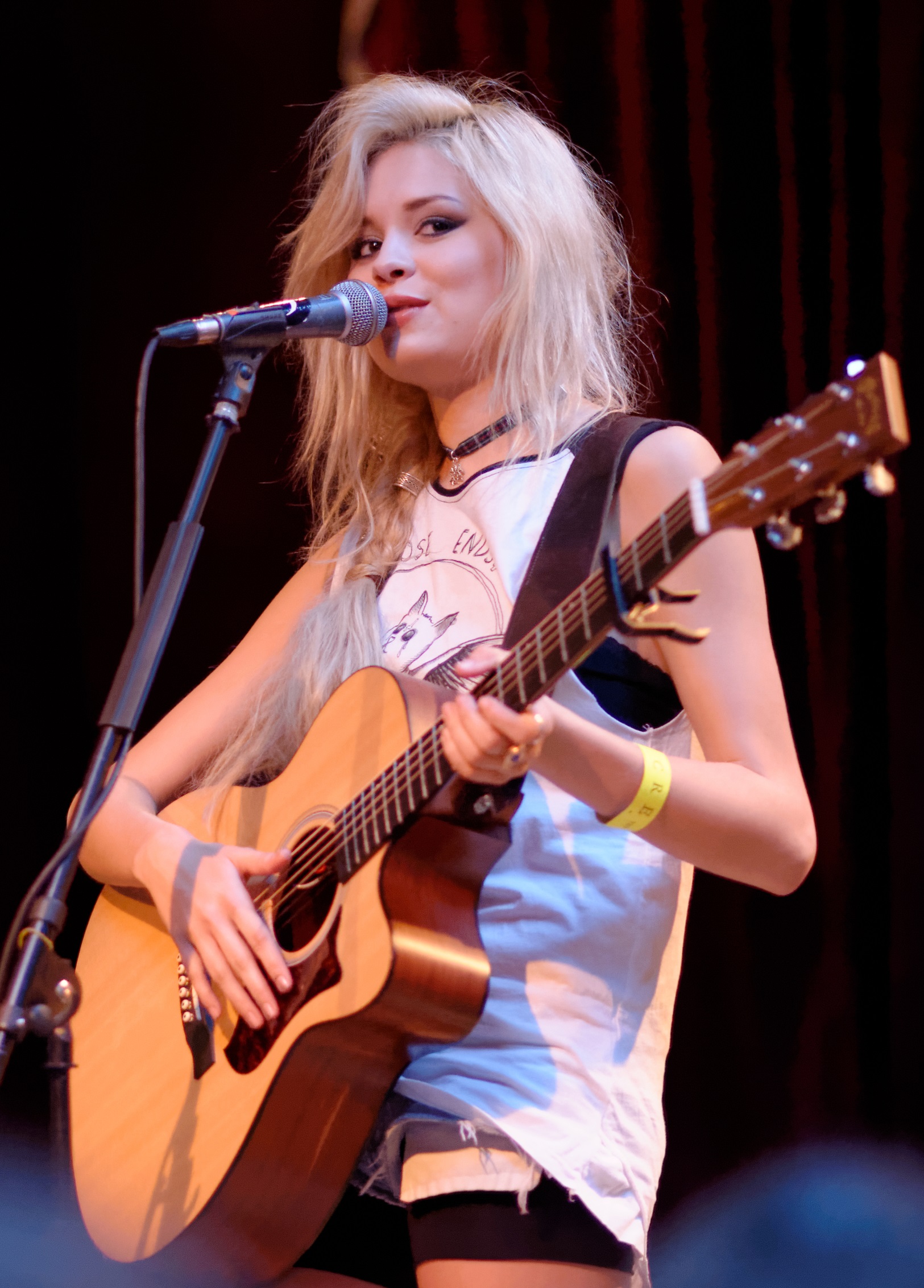} \\
    \footnotesize{Nina Nesbitt}
  \end{minipage}
  }
  \adjustbox{valign=t}{
  \begin{minipage}[t]{0.5\textwidth}
    \textcolor{blue}{\textbf{Entity Info}} \\
    Language: Indonesian \\
    Entity name: Nina Nesbitt \\
    Country: United Kingdom \\
    Wikidata ID: Q510361 \\
    Question type: Property-level question \\
    Property: P27 (country of citizenship) \\
    Template question: Negara manakah yang mengakui entitas ini sebagai warganya? \\
    \hspace*{1em}(Which country recognizes this entity as its citizen?) \\
    Template answer: Nina Nesbitt diakui sebagai warga negara Britania Raya, negara di Eropa Barat Laut. \\
    \hspace*{1em}(Nina Nesbitt is recognized as a citizen of United Kingdom, country in north-west Europe.)
  \end{minipage}
  } \\

  \hrulefill

  \textcolor{blue}{\textbf{Question}} \\
  Kalau melihat gambar ini, Nina Nesbitt berasal dari negara mana ya? \\
  (Where is Nina Nesbitt from?) \\

  \textcolor{blue}{\textbf{Answer}} \\
  Nina Nesbitt adalah warga negara Britania Raya, sebuah negara di Eropa Barat Laut. \\
  Britania Raya dikenal dengan sejarah dan budayanya yang kaya, serta merupakan rumah bagi banyak musisi dan seniman berbakat seperti Nina Nesbitt. \\
  (Nina Nesbitt is a singer-songwriter from the United Kingdom, specifically Scotland. She is known for her indie pop music and has gained recognition as a Scottish musical artist.)
\end{tcolorbox}
}
\end{figure*}

\begin{figure*}[t]
\definecolor{highlightcolor}{RGB}{255, 120, 84}
\definecolor{framecolor}{RGB}{162, 59, 114}

\resizebox{\textwidth}{!}{%
\begin{tcolorbox}[
  colback=highlightcolor!15,
  colframe=framecolor,
  boxrule=1.0pt,
  title=CulturalGround(Spanish),
]

  \adjustbox{valign=t}{
  \begin{minipage}[t]{0.45\textwidth}
    \includegraphics[width=0.8\linewidth]{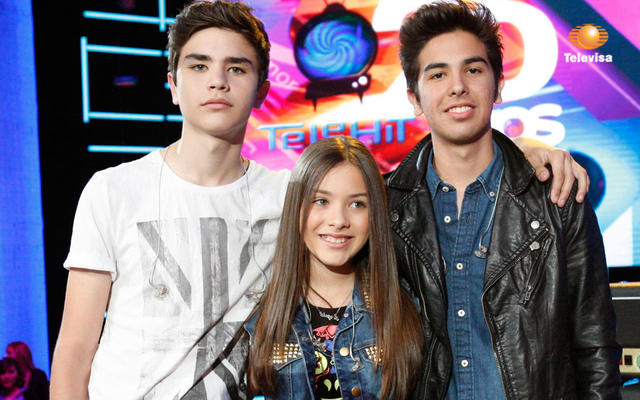} \\
    \footnotesize{Vázquez Sounds}
  \end{minipage}
  }
  \adjustbox{valign=t}{
  \begin{minipage}[t]{0.5\textwidth}
    \textcolor{blue}{\textbf{Entity Info}} \\
    Language: Spanish \\
    Entity name: Vázquez Sounds \\
    Country: Mexico \\
    Wikidata ID: Q2485066 \\
    Question type: Property-level question \\
    Property: P495 (country of origin) \\
    Template question: ¿Cuál es el país de origen de esta entidad?(What is the country of origin of this entity?) \\
    Template answer: Vázquez Sounds se originó en México, país de América del Norte.(Vázquez Sounds originated in Mexico, a country in North America.)
  \end{minipage}
  } \\

  \hrulefill

  \textcolor{blue}{\textbf{Question}} \\
  ¿En qué país se originó este grupo musical? \\
  (In which country did this music group originate?) \\

  \textcolor{blue}{\textbf{Answer}} \\
  Vázquez Sounds se originó en México, un país de América del Norte. \\
  (Vázquez Sounds originated in Mexico, a country in North America.)
\end{tcolorbox}
}
\end{figure*}

\begin{figure*}[t]
\definecolor{highlightcolor}{RGB}{255, 120, 84}
\definecolor{framecolor}{RGB}{162, 59, 114}

\resizebox{\textwidth}{!}{%
\begin{tcolorbox}[
  colback=highlightcolor!15,
  colframe=framecolor,
  boxrule=1.0pt,
  title=CulturalGround(Spanish),
]

  \adjustbox{valign=t}{
  \begin{minipage}[t]{0.45\textwidth}
    \includegraphics[width=0.8\linewidth]{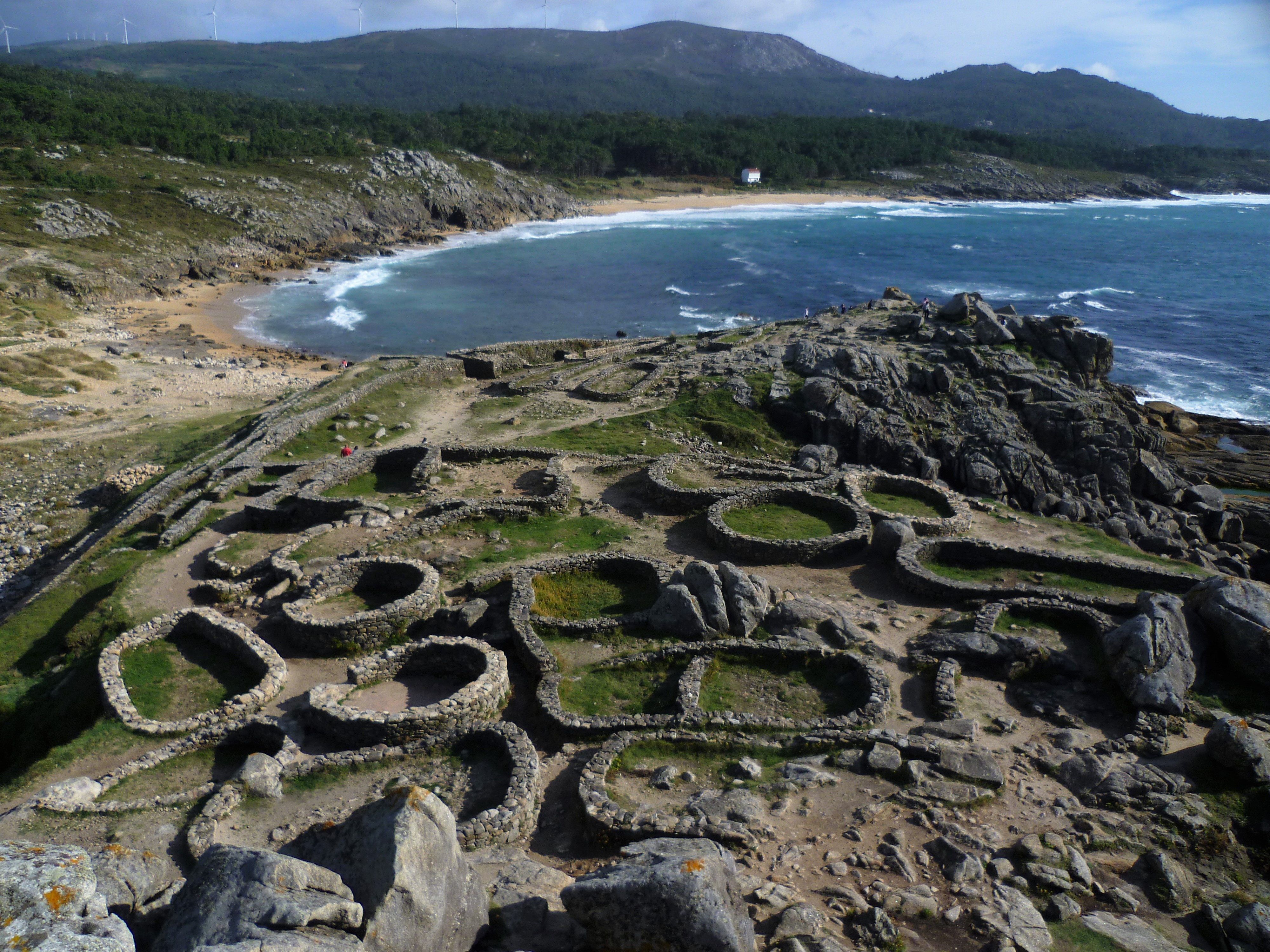} \\
    \footnotesize{castro}
  \end{minipage}
  }
  \adjustbox{valign=t}{
  \begin{minipage}[t]{0.5\textwidth}
    \textcolor{blue}{\textbf{Entity Info}} \\
    Language: Spanish \\
    Entity name: castro \\
    Country: Spain \\
    Wikidata ID: Q5050823 \\
    Question type: Property-level question \\
    Property: P2596 (culture) \\
    Template question: ¿Con qué cultura está asociada esta entidad? \\
    \hspace*{1em}(Which culture is this entity associated with?) \\
    Template answer: El \textit{castro} está asociado a la cultura castreña, una cultura arqueológica ibérica. \\
    \hspace*{1em}(The \textit{castro} is associated with the Castro culture, an Iberian archaeological culture.)
  \end{minipage}
  } \\

  \hrulefill

  \textcolor{blue}{\textbf{Question}} \\
  ¿A qué cultura o tradición arqueológica se asocia normalmente este tipo de asentamiento? \\
  (What culture built these kinds of settlements?) \\

  \textcolor{blue}{\textbf{Answer}} \\
  El \textit{castro} está asociado a la cultura castreña, una cultura arqueológica propia del noroeste de España. Estos asentamientos fortificados son característicos de la época prerromana y reflejan las formas de vida de las comunidades locales antes de la llegada del Imperio Romano. \\
  (This \textit{castro} is associated with the Castro culture, an archaeological culture primarily found in northern Spain. These fortified settlements were often constructed before the Roman period and represent a significant part of the region’s pre-Roman history.)
\end{tcolorbox}
}
\end{figure*}

\begin{figure*}[t]
\definecolor{highlightcolor}{RGB}{255, 120, 84}
\definecolor{framecolor}{RGB}{162, 59, 114}

\resizebox{\textwidth}{!}{%
\begin{tcolorbox}[
  colback=highlightcolor!15,
  colframe=framecolor,
  boxrule=1.0pt,
  title=CulturalGround(English),
]

  \adjustbox{valign=t}{
  \begin{minipage}[t]{0.45\textwidth}
    \includegraphics[width=0.8\linewidth]{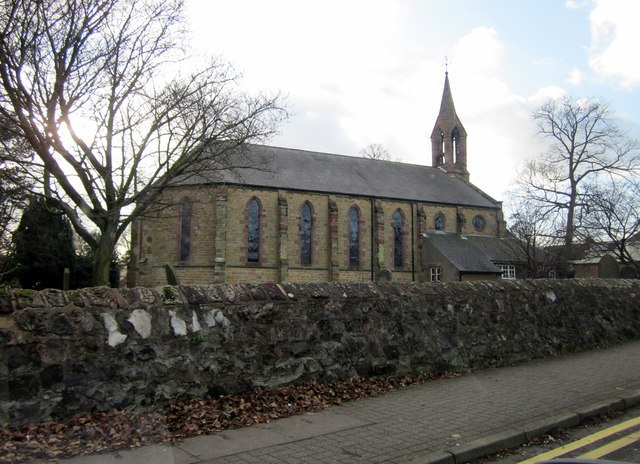} \\
    \footnotesize{Christ Church}
  \end{minipage}
  }
  \adjustbox{valign=t}{
  \begin{minipage}[t]{0.5\textwidth}
    \textcolor{blue}{\textbf{Entity Info}} \\
    Language: English \\
    Entity name: Christ Church \\
    Country: United Kingdom \\
    Wikidata ID: Q19362878 \\
    Question type: Property-level question \\
    Property: P140 (religion or worldview) \\
    Template question: Which religion is associated with this entity? \\
    Template answer: Christ Church is associated with Anglicanism, a Christian tradition developing out of the practices, liturgy and identity of the Church of England.
  \end{minipage}
  } \\

  \hrulefill

  \textcolor{blue}{\textbf{Question}} \\
  What Christian tradition does this church belong to? \\

  \textcolor{blue}{\textbf{Answer}} \\
  Christ Church belongs to the Anglican tradition, which developed from the practices and identity of the Church of England. As part of the Anglican faith, Christ Church observes a blend of Catholic and Protestant traditions and is common throughout the United Kingdom.
\end{tcolorbox}
}
\end{figure*}

\begin{figure*}[t]
\definecolor{highlightcolor}{RGB}{255, 120, 84}
\definecolor{framecolor}{RGB}{162, 59, 114}

\resizebox{\textwidth}{!}{%
\begin{tcolorbox}[
  colback=highlightcolor!15,
  colframe=framecolor,
  boxrule=1.0pt,
  title=CulturalGround(German),
]

  \adjustbox{valign=t}{
  \begin{minipage}[t]{0.45\textwidth}
    \includegraphics[width=0.8\linewidth]{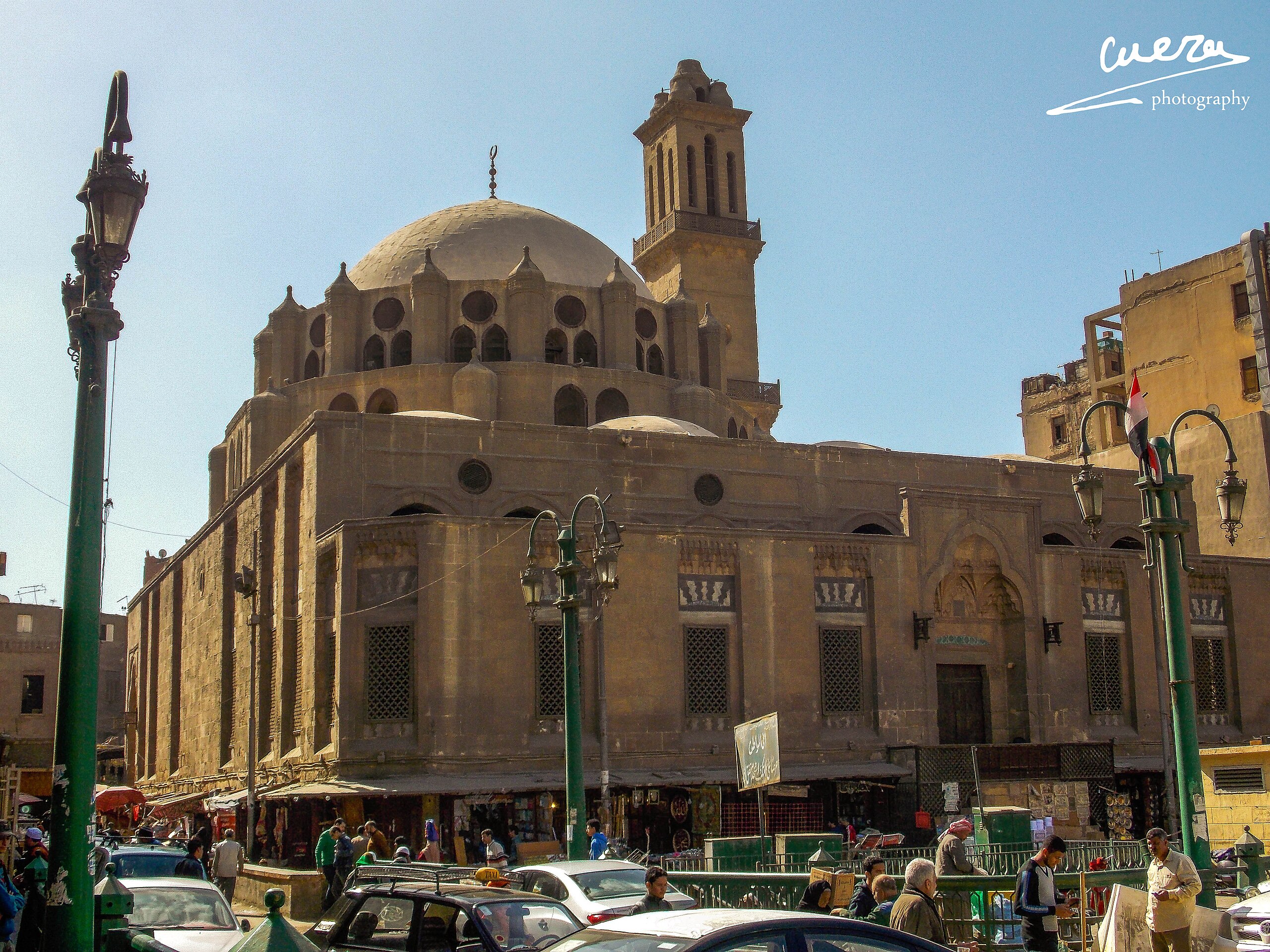} \\
    \footnotesize{Muhammad-Bey-Abu-adh-Dhahab-Moschee}
  \end{minipage}
  }
  \adjustbox{valign=t}{
  \begin{minipage}[t]{0.5\textwidth}
    \textcolor{blue}{\textbf{Entity Info}} \\
    Language: German \\
    Entity name: Muhammad-Bey-Abu-adh-Dhahab-Moschee \\
    Country: Egypt \\
    Wikidata ID: Q6916436 \\
    Question type: Property-level question \\
    Property: P1435 (heritage designation) \\
    Template question: Welchen Denkmalstatus hat diese Entität? \\
    \hspace*{1em}(What heritage designation does this entity have?) \\
    Template answer: Muhammad-Bey-Abu-adh-Dhahab-Moschee hat den Denkmalstatus Teil einer UNESCO-Welterbestätte, Element einer UNESCO-Welterbestätte. \\
    \hspace*{1em}(The Muhammad Bek Abu El Dahab Complex has the heritage designation of being part of a UNESCO World Heritage Site, a component forming part of a UNESCO World Heritage Site.)
  \end{minipage}
  } \\

  \hrulefill

  \textcolor{blue}{\textbf{Question}} \\
  Welchen Denkmalstatus hat diese Moschee? \\
  (What special cultural recognition has this mosque received?) \\

  \textcolor{blue}{\textbf{Answer}} \\
  Die Muhammad-Bey-Abu-adh-Dhahab-Moschee hat den Denkmalstatus als Teil einer UNESCO-Welterbestätte. \\
  (The Muhammad Bek Abu El Dahab Complex has been recognized as part of a UNESCO World Heritage Site, highlighting its significant cultural and historical importance in Egypt.)
\end{tcolorbox}
}
\end{figure*}

\begin{figure*}[t]
\definecolor{highlightcolor}{RGB}{255, 120, 84}
\definecolor{framecolor}{RGB}{162, 59, 114}

\resizebox{\textwidth}{!}{%
\begin{tcolorbox}[
  colback=highlightcolor!15, 
  colframe=framecolor, 
  boxrule=1.0pt, 
  title=CulturalGround(Turkish),
]

  \adjustbox{valign=t}{
  \begin{minipage}[t]{0.45\textwidth}
    \includegraphics[width=0.8\linewidth]{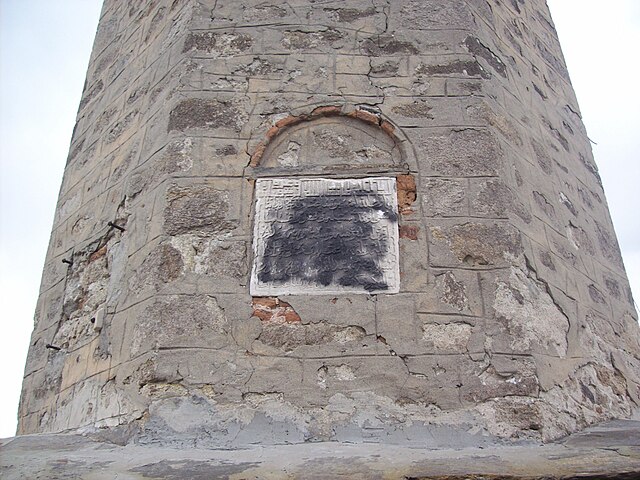} \\
    \footnotesize{Filibe Saat Kulesi(Clock Tower of Plovdiv)}
  \end{minipage}
  }
  \adjustbox{valign=t}{
  \begin{minipage}[t]{0.5\textwidth}
    \textcolor{blue}{\textbf{Entity Info}} \\
    Language: Turkish \\
    Entity name: Filibe Saat Kulesi(Clock Tower of Plovdiv) \\
    Country: Bulgaria \\
    Wikidata ID: Q37939027 \\
    Question type: Entity-level question \\
    Template question: Bu görüntüde gösterilen varlık nedir?(What is the entity shown in this image?) \\
    Template answer: Bu görüntüde gösterilen varlık Filibe Saat Kulesi'dir, .(The entity shown in the image is Clock Tower of Plovdiv, .)
  \end{minipage}
  } \\

  \hrulefill

  \textcolor{blue}{\textbf{Question}} \\
  Bu fotoğraftaki yapı ne olarak bilinir?(What is the structure in this photo known as?) \\

  \textcolor{blue}{\textbf{Answer}} \\
  Bu fotoğraftaki yapı Filibe Saat Kulesi'dir. Bulgaristan'ın Filibe şehrinde yer alan bu kule, şehrin sembol yapılarından biridir ve Osmanlı döneminden kalmıştır. (The structure in this photo is the Plovdiv Clock Tower. It is located in Plovdiv, Bulgaria, and is one of the city’s landmark buildings, dating back to the Ottoman period.)
\end{tcolorbox}
}
\end{figure*}

\begin{figure*}[t]
\definecolor{highlightcolor}{RGB}{255, 120, 84}
\definecolor{framecolor}{RGB}{162, 59, 114}

\resizebox{\textwidth}{!}{%
\begin{tcolorbox}[
  colback=highlightcolor!15, 
  colframe=framecolor, 
  boxrule=1.0pt, 
  title=CulturalGround(Japanese),
]

  \adjustbox{valign=t}{
  \begin{minipage}[t]{0.45\textwidth}
    \includegraphics[width=0.8\linewidth]{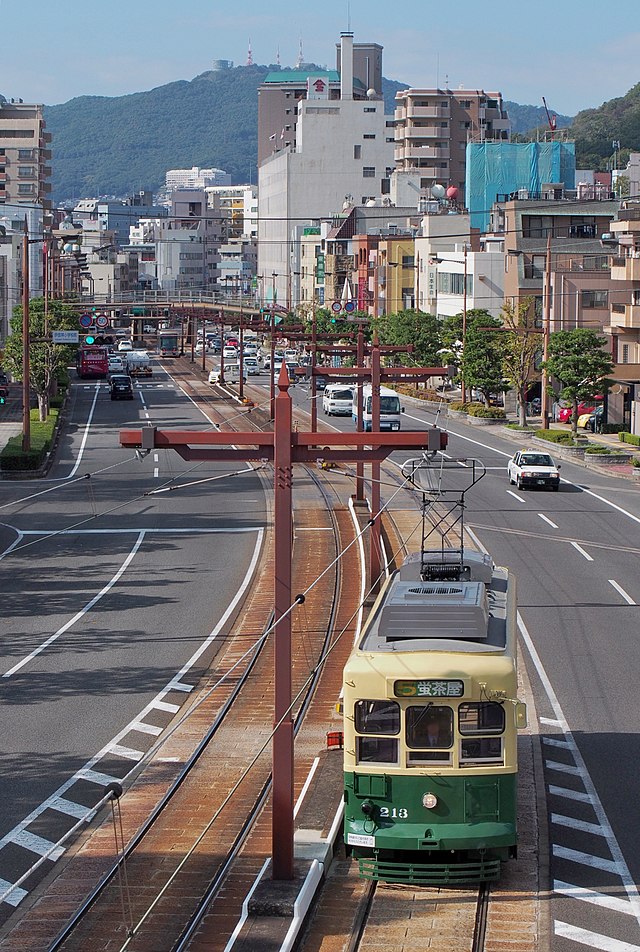} \\
    \footnotesize{Shinnakagawa-machi Station}
  \end{minipage}
  }
  \adjustbox{valign=t}{
  \begin{minipage}[t]{0.5\textwidth}
    \textcolor{blue}{\textbf{Entity Info}} \\
    Language: Japanese \\
    Entity name: \jp{新中川町停留場}(Shinnakagawa-machi Station) \\
    Country: Japan \\
    Wikidata ID: Q11501118 \\
    Question type: Entity-level question \\
    Template question: \jp{この画像に表示されているものは何ですか？}?(What is the entity shown in this image?) \\
    Template answer: \jp{この画像に表示されているのは新中川町停留場です。長崎県長崎市にある長崎電気軌道の路面電車停留場.} (The entity shown in the image is Shin Nakagawa-Machi Station, tram station in Nagasaki, Nagasaki prefecture, Japan.)
  \end{minipage}
  } \\

  \hrulefill

  \textcolor{blue}{\textbf{Question}} \\
 \jp{画像に映っている停留場はどこにありますか} (What type of transportation facility is shown in this image?)\\

  \textcolor{blue}{\textbf{Answer}} \\
  \jp{画像に映っているのは新中川町停留場で、長崎県長崎市にある長崎電気軌道の路面電車停留場です。} (Shin Nakagawa-Machi Station is a tram station located in Nagasaki, Nagasaki prefecture, Japan.)
\end{tcolorbox}
}
\end{figure*}

\begin{figure*}[t]
  \centering
  \includegraphics[width=\linewidth]{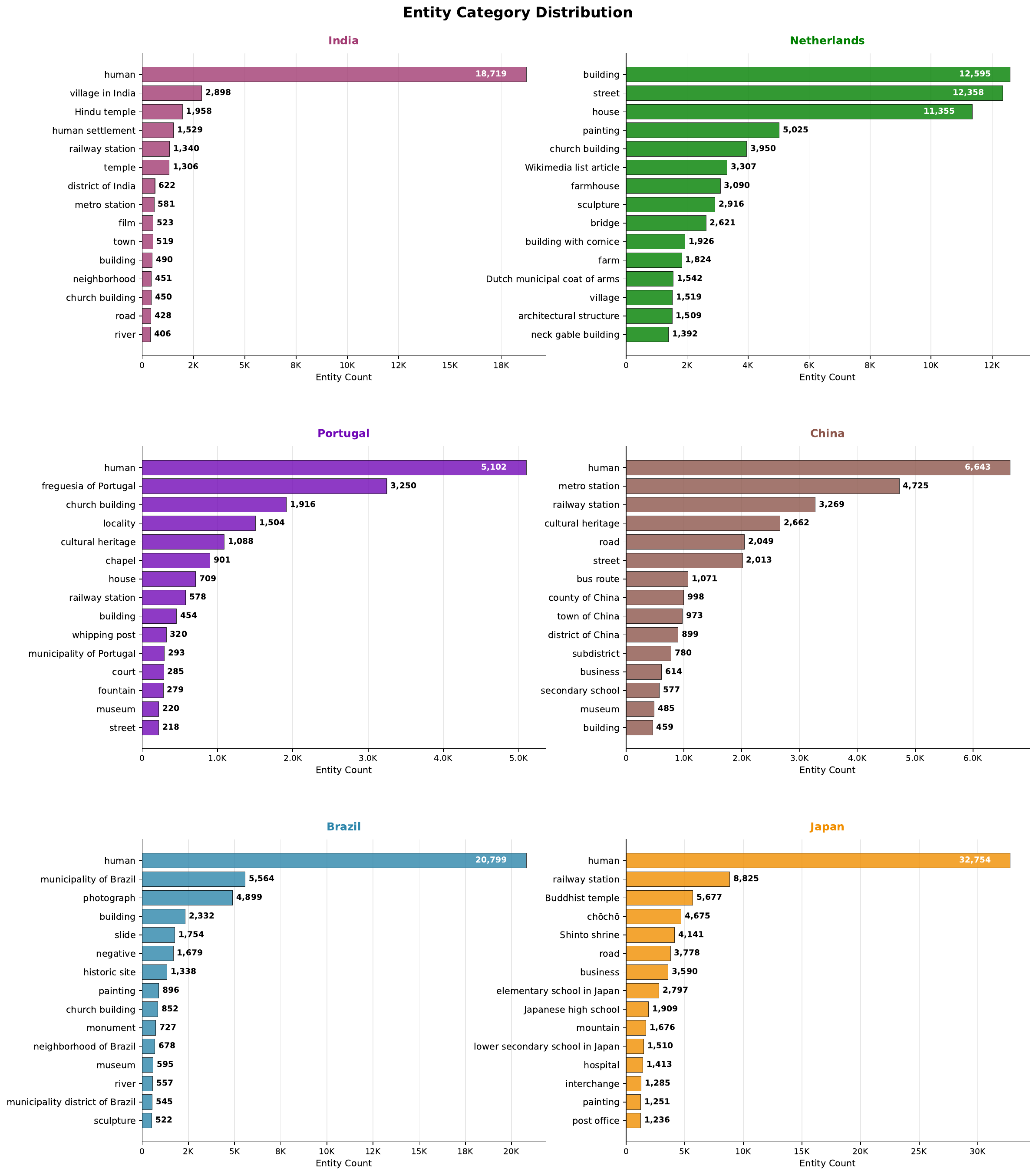}
  \caption{Entity–category distributions for six representative countries reveal distinct regional cultural emphases. India focuses on heritage and settlements (e.g., Hindu temples, villages); the Netherlands highlights vernacular architecture and art (e.g., historic buildings, streets, paintings); Portugal surfaces ecclesiastical heritage and protected cultural assets (e.g., church buildings, chapels, registered heritage sites); China emphasizes infrastructure and administrative categories (e.g., metro and railway stations, county‐ and town‐level entities); Brazil shows strong cultural media and monument presence (e.g., photographs, negatives, historic sites); and Japan combines religious architecture with education‐related categories (e.g., Buddhist temples, Shinto shrines, schools). Bar length shows entity counts, with longer bars indicating more frequent categories within each country.}
  \label{fig:entity_categories}
\end{figure*}

%language diversity: germany, brazil, italy, egypt, india
\begin{figure*}[!htbp]
  \centering
  \begin{minipage}[c]{1.0\linewidth}
    \includegraphics[width=\linewidth]{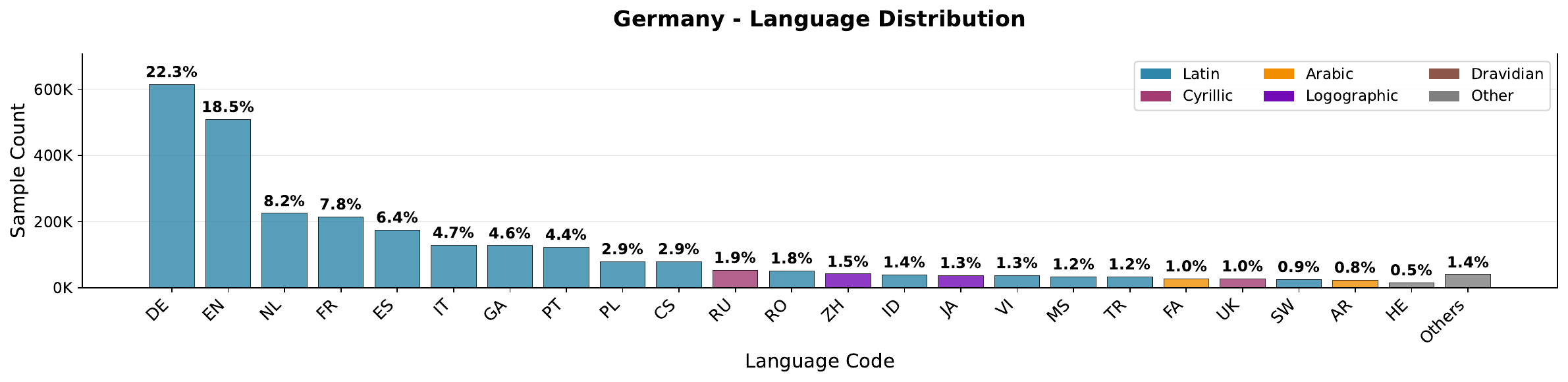}
  \end{minipage}
  
  \vspace{0.3cm}
  
  \begin{minipage}[c]{1.0\linewidth}
    \includegraphics[width=\linewidth]{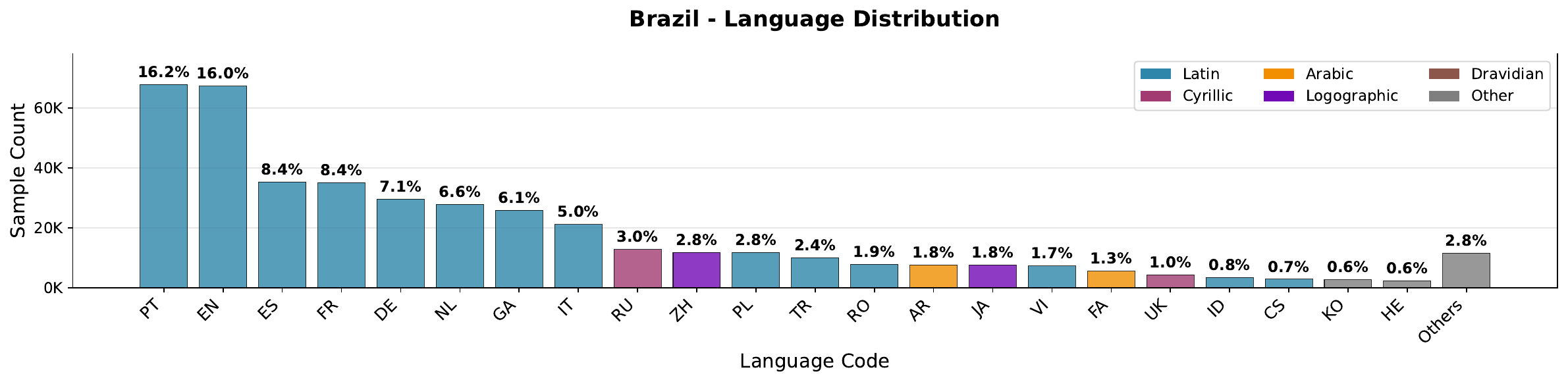}
  \end{minipage}
  
  \vspace{0.3cm}
  
  \begin{minipage}[c]{1.0\linewidth}
    \includegraphics[width=\linewidth]{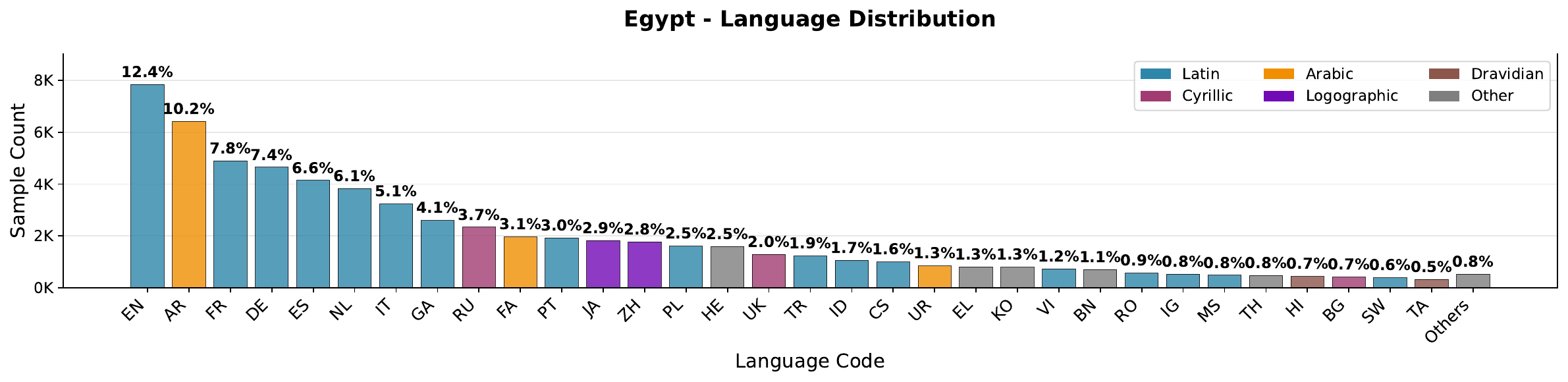}
  \end{minipage}
  
  \vspace{0.3cm}
  
  \begin{minipage}[c]{1.0\linewidth}
    \includegraphics[width=\linewidth]{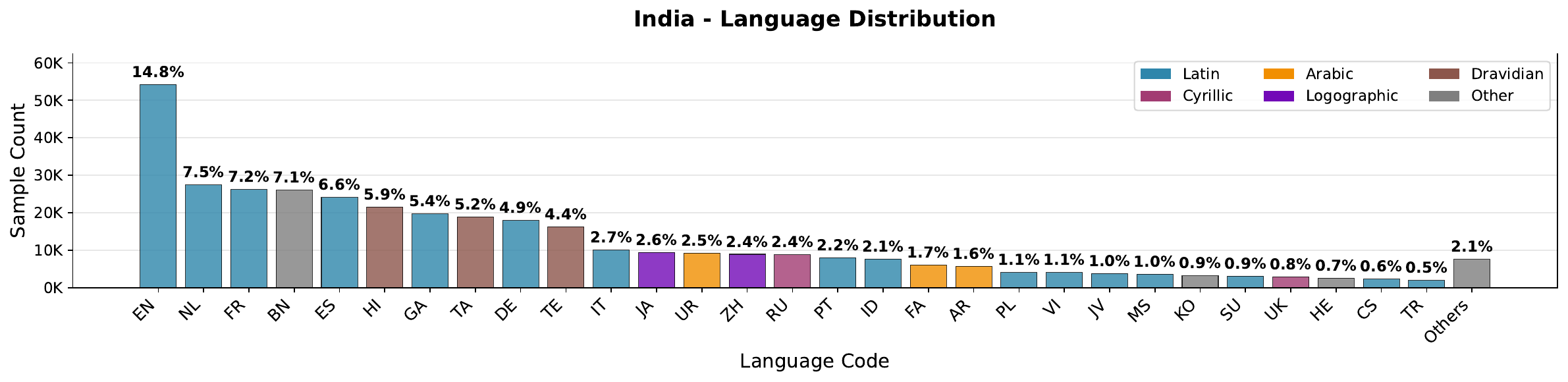}
  \end{minipage}
  
  \vspace{0.3cm}
  
  \begin{minipage}[c]{1.0\linewidth}
    \includegraphics[width=\linewidth]{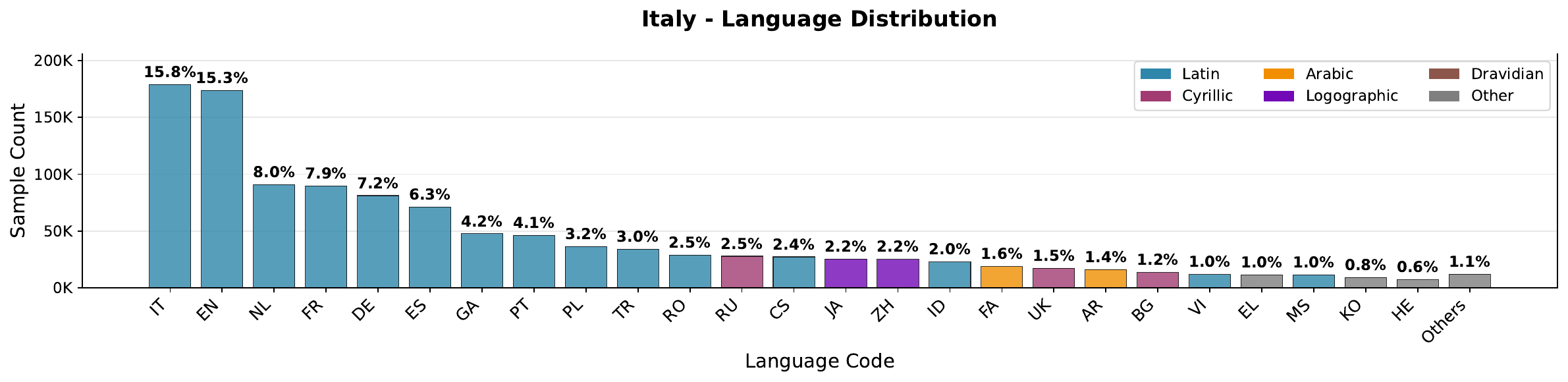}
  \end{minipage}
  
  \caption{Language distribution across five countries showing the diversity of multilingual content in \dataset. Each plot displays languages with less than 0.5\% representation, with colors indicating script families. The percentages above each bar indicate the proportion of samples in each language within the respective country's subset.}
  \label{fig:language_diversity_distribution}
\end{figure*}

% force a page break and switch to one-column
% \clearpage
% \onecolumn

% \tiny
% \renewcommand{\arraystretch}{0.9}
% \setlength{\tabcolsep}{3pt}

% \begin{longtable}{@{}  
%     l            % Relation
%     l            % Label
%     p{3cm}       % Description
%     p{4.5cm}     % Template question
%     p{4.5cm}     % Answer template
%   @{}}
% \toprule
% \textbf{Relation} & \textbf{Label} & \textbf{Description} 
%   & \textbf{Template question} & \textbf{Answer template} \\
% \midrule
% \endfirsthead

% \toprule
% \textbf{Relation} & \textbf{Label} & \textbf{Description} 
%   & \textbf{Template question} & \textbf{Answer template} \\
% \midrule
% \endhead

% \midrule \multicolumn{5}{r}{\emph{(continued)}} \\ \midrule
% \endfoot

% \bottomrule
% \endlastfoot

% P17   & country                        & sovereign state of this item; don’t use on humans  
%       & Which sovereign state does this entity belong to?  
%       & \{entity\_label\} belongs to the sovereign state of \{property\_value\}. \\

% P2596 & culture                        & culture associated with this entity  
%       & Which culture is this entity associated with?  
%       & \{entity\_label\} is associated with \{property\_value\} culture. \\

% % … paste in all remaining relation rows here …

% \end{longtable}

% % switch back to two-column for whatever follows
% \twocolumn

% force a page break and switch to one-column
\clearpage
\onecolumn

\tiny
\renewcommand{\arraystretch}{0.99}
\setlength{\tabcolsep}{3pt}

\begin{longtable}{@{}  
    l            % Relation
    p{2.5cm}            % Label
    p{3cm}       % Description
    p{4.2cm}     % Template question
    p{4.5cm}     % Answer template
  @{}}
\toprule
\textbf{Property} & \textbf{Label} & \textbf{Description} 
  & \textbf{Template question} & \textbf{Answer template} \\
\midrule
\endfirsthead

\toprule
\textbf{Property} & \textbf{Label} & \textbf{Description} 
  & \textbf{Template question} & \textbf{Answer template} \\
\midrule
\endhead

\midrule \multicolumn{5}{r}{\emph{(continued)}} \\ \midrule
\endfoot

\bottomrule
\endlastfoot

P17    & country                        & sovereign state of this item; don’t use on humans  
       & Which sovereign state does this entity belong to?  
       & \{entity\_label\} belongs to the sovereign state of \{property\_value\}. \\

P2596  & culture                        & culture associated with this entity  
       & Which culture is this entity associated with?  
       & \{entity\_label\} is associated with \{property\_value\} culture. \\

P172   & ethnic group                   & ethnic group of which an entity is a member  
       & Which ethnic group is this entity a member of?  
       & \{entity\_label\} is a member of the \{property\_value\} ethnic group. \\

P140   & religion                       & religion of a person, organization or religious building, or associated with this subject  
       & Which religion is associated with this entity?  
       & \{entity\_label\} is associated with the \{property\_value\} religion. \\

P1559  & name in native language        & name of the entity in its native or original language  
       & What is the name of this entity in its native language?  
       & The name of \{entity\_label\} in its native language is \{property\_value\}. \\

P37    & official language              & language designated as official by this item  
       & Which language is officially designated by this entity?  
       & \{entity\_label\} officially designates the language \{property\_value\}. \\

P103   & native language                & language or languages a person has learned from early childhood  
       & Which language did this entity learn from early childhood?  
       & \{entity\_label\} learned \{property\_value\} from early childhood. \\

P825   & dedicated to                   & subject is dedicated to this entity  
       & What is this entity dedicated to?  
       & \{entity\_label\} is dedicated to \{property\_value\}. \\

P149   & architectural style            & architectural style of a building or structure  
       & What is the architectural style of this entity?  
       & \{entity\_label\} is built in the \{property\_value\} architectural style. \\

P1435  & heritage designation           & heritage designation of a cultural site or monument  
       & What heritage designation does this entity have?  
       & \{entity\_label\} has the heritage designation of \{property\_value\}. \\

P1268  & represents                     & entity, concept or subject represented by this item  
       & What does this entity represent?  
       & \{entity\_label\} represents \{property\_value\}. \\

P837   & day in year for periodic occurrence  
       & date in the year when a periodic event occurs  
       & On which day of the year does this event occur?  
       & \{entity\_label\} occurs on \{property\_value\}. \\

P495   & country of origin              & country of origin of this item (creative work, food, phrase, product, etc.)  
       & Which country is the origin of this entity?  
       & \{entity\_label\} originated in \{property\_value\}. \\

P407   & language of work or name       & language associated with this creative work  
       & Which language is associated with this entity?  
       & \{entity\_label\} is associated with the \{property\_value\} language. \\

P1412  & languages spoken, written or signed  
       & language(s) that a person speaks or writes, including the native language(s)  
       & Which language(s) does this entity speak or write?  
       & \{entity\_label\} speaks or writes \{property\_value\}. \\

P2936  & language used                  & language widely used in this place or by this organization  
       & Which language is used by this entity?  
       & \{entity\_label\} uses the language \{property\_value\}. \\

P131   & located in the administrative territorial entity  
       & the item is located on the territory of the following administrative entity  
       & Within which administrative territorial entity is this entity located?  
       & \{entity\_label\} is located in \{property\_value\}. \\

P276   & location                       & location of the item, physical object or event  
       & Where is this entity located?  
       & \{entity\_label\} is located in \{property\_value\}. \\

P625   & coordinate location            & geocoordinates of the subject  
       & What are the coordinates of this entity?  
       & \{entity\_label\} is located at coordinates \{property\_value\}. \\

P706   & located in/on physical feature & located on the given landform or body of water  
       & On which physical feature is this entity located?  
       & \{entity\_label\} is located on \{property\_value\}. \\

P206   & located in or next to body of water  
       & body of water on or next to which a place is located  
       & Which body of water is this entity located in or next to?  
       & \{entity\_label\} is located in or next to \{property\_value\}. \\

P30    & continent                      & continent of which the subject is a part  
       & On which continent is this entity located?  
       & \{entity\_label\} is located on the continent of \{property\_value\}. \\

P170   & creator                        & maker of this creative work or other object  
       & Who is the creator of this entity?  
       & \{entity\_label\} was created by \{property\_value\}. \\

P86    & composer                       & person(s) who wrote the music for this song, musical work, or opera  
       & Who composed the music for this entity?  
       & The music for \{entity\_label\} was composed by \{property\_value\}. \\

P162   & producer                       & person(s) who produced the film, musical work, theatrical production, etc.  
       & Who produced this entity?  
       & \{entity\_label\} was produced by \{property\_value\}. \\

P136   & genre                          & creative work’s genre or an artist’s field of work  
       & What genre is associated with this entity?  
       & \{entity\_label\} is associated with the \{property\_value\} genre. \\

P571   & inception                      & time when an entity begins to exist; for date of official opening use P1619  
       & When was this entity established or founded?  
       & \{entity\_label\} was established in \{property\_value\}. \\

P585   & point in time                  & date or point in time when an event occurred  
       & When did this event occur?  
       & \{entity\_label\} occurred on \{property\_value\}. \\

P1269  & facet of                       & topic of which this item is an aspect  
       & This entity is a facet of which broader topic?  
       & \{entity\_label\} is a facet of \{property\_value\}. \\

P19    & place of birth                 & most specific known birth location of a person, animal or fictional character  
       & Where was this entity born?  
       & \{entity\_label\} was born in \{property\_value\}. \\

P20    & place of death                 & most specific known death location of a person, animal or fictional character  
       & Where did this entity die?  
       & \{entity\_label\} died in \{property\_value\}. \\

P27    & country of citizenship         & the object is a country that recognizes the subject as its citizen  
       & Which country recognizes this entity as its citizen?  
       & \{entity\_label\} is recognized as a citizen of \{property\_value\}. \\

P569   & date of birth                  & date on which the subject was born  
       & When was this entity born?  
       & \{entity\_label\} was born on \{property\_value\}. \\

P570   & date of death                  & date on which the subject died  
       & When did this entity die?  
       & \{entity\_label\} died on \{property\_value\}. \\

P36    & capital                        & primary city of a country, state or other type of administrative territorial entity  
       & What is the capital of this entity?  
       & The capital of \{entity\_label\} is \{property\_value\}. \\

P1376  & capital of                     & administrative division of which the municipality is the governmental seat  
       & Of which administrative division is this entity the capital?  
       & \{entity\_label\} is the capital of \{property\_value\}. \\

P47    & shares border with             & countries or administrative subdivisions that this item borders  
       & Which entity does this share a border with?  
       & \{entity\_label\} shares a border with \{property\_value\}. \\

P106   & occupation                     & occupation of a person  
       & What is the occupation of this entity?  
       & \{entity\_label\}’s occupation is \{property\_value\}. \\

P39    & position held                  & position or public office currently or formerly held  
       & Which position does this entity hold?  
       & \{entity\_label\} holds the position of \{property\_value\}. \\

P102   & member of political party      & the political party of which this politician is or has been a member  
       & Which political party is this entity a member of?  
       & \{entity\_label\} is a member of the \{property\_value\} party. \\

P166   & award received                 & award or recognition received by a person, organisation or creative work  
       & Which award did this entity receive?  
       & \{entity\_label\} received the following award(s): \{property\_value\}. \\

P800   & notable work                   & notable scientific, artistic or literary work among subject’s works  
       & What is a notable work by this entity?  
       & A notable work by \{entity\_label\} is \{property\_value\}. \\

P1303  & instrument                     & musical instrument that a person plays  
       & Which musical instrument does this entity play?  
       & \{entity\_label\} plays the \{property\_value\}. \\

P641   & sport                          & sport that the subject participates in or is associated with  
       & In which sport does this entity participate?  
       & \{entity\_label\} participates in \{property\_value\}. \\

P54    & member of sports team          & sports teams that the subject represents or represented  
       & Which sports team does this entity represent?  
       & \{entity\_label\} represents the sports team \{property\_value\}. \\

P69    & educated at                    & educational institution attended by subject  
       & Which educational institution did this entity attend?  
       & \{entity\_label\} studied at \{property\_value\}. \\

P512   & academic degree                & academic degree that the person holds  
       & Which academic degree does this entity hold?  
       & \{entity\_label\} holds the academic degree of \{property\_value\}. \\

P101   & field of work                  & specialization of a person or organization  
       & In which field does this entity work?  
       & \{entity\_label\} works in the field of \{property\_value\}. \\

P108   & employer                       & person or organization for which the subject works  
       & Who employs this entity?  
       & \{entity\_label\} is employed by \{property\_value\}. \\

P937   & work location                  & location where persons were active  
       & In which location was this entity active?  
       & \{entity\_label\} was active in \{property\_value\}. \\

P31    & instance of                    & class of which this subject is a particular example  
       & This entity is an example of which class?  
       & \{entity\_label\} is an instance of \{property\_value\}. \\

P279   & subclass of                    & this item is a class (subset) of that item  
       & Of which broader class is this entity a subclass?  
       & \{entity\_label\} is a subclass of \{property\_value\}. \\

P361   & part of                        & object of which the subject is a part  
       & Which larger entity is this entity part of?  
       & \{entity\_label\} is part of \{property\_value\}. \\

P527   & has part                       & part of this subject  
       & Which parts does this entity include?  
       & \{entity\_label\} consists of \{property\_value\}. \\

P138   & named after                    & entity or event that inspired the subject’s name  
       & What is this entity named after?  
       & \{entity\_label\} was named after \{property\_value\}. \\

P577   & publication date               & date when a work was first published or released  
       & When was this entity published?  
       & \{entity\_label\} was published on \{property\_value\}. \\

P1619  & date of official opening        & date when a place or organization officially opened  
       & When was this entity officially opened?  
       & \{entity\_label\} was officially opened on \{property\_value\}. \\

P740   & location of formation          & location where a group or organization was formed  
       & Where was this entity formed?  
       & \{entity\_label\} was formed in \{property\_value\}. \\

P159   & headquarters location          & location where an organization's headquarters is situated  
       & Where are the headquarters of this entity located?  
       & The headquarters of \{entity\_label\} is located in \{property\_value\}. \\

P793   & significant event              & significant or notable events associated with the subject  
       & What significant events are associated with this entity?  
       & Notable events associated with \{entity\_label\} include \{property\_value\}. \\

P463   & member of                      & organization or club to which the subject belongs  
       & Which organization is this entity a member of?  
       & \{entity\_label\} is a member of \{property\_value\}. \\

P190   & twinned administrative body    & twin towns or sister cities  
       & Which city is twinned with this entity?  
       & \{entity\_label\} is twinned with \{property\_value\}. \\

P530   & diplomatic relation            & diplomatic relations of the country  
       & With which countries does this entity maintain diplomatic relations?  
       & \{entity\_label\} maintains diplomatic relations with \{property\_value\}. \\

P176   & manufacturer                   & manufacturer or producer of this product  
       & Who manufactures this entity?  
       & \{entity\_label\} is produced by \{property\_value\}. \\

P178   & developer                      & organisation or person that developed the item  
       & Who developed this entity?  
       & \{entity\_label\} was developed by \{property\_value\}. \\

P127   & owned by                       & owner of the subject  
       & Who owns this entity?  
       & \{entity\_label\} is owned by \{property\_value\}. \\

P137   & operator                       & entity that operates the equipment, facility, or service  
       & Who operates this entity?  
       & \{entity\_label\} is operated by \{property\_value\}. \\

P449   & original network               & network the radio or television show was originally aired on  
       & On which network was this show originally aired?  
       & \{entity\_label\} was originally aired on the \{property\_value\} network. \\

P264   & record label                   & brand associated with the marketing of music recordings  
       & Under which record label is this entity’s music released?  
       & \{entity\_label\}’s music is released under \{property\_value\}. \\

P364   & original language of film or TV show  
       & language in which a film or performance work was originally created  
       & In which language was this entity originally created?  
       & \{entity\_label\} was originally created in \{property\_value\}. \\

P180   & depicts                        & entity visually depicted in an image or work  
       & What does this entity depict?  
       & This entity depicts \{property\_value\}. \\

P921   & main subject                   & primary topic of a work  
       & What is the main subject of this entity?  
       & The main subject of \{entity\_label\} is \{property\_value\}. \\

P1433  & published in                   & larger work that a given work was published in  
       & In which larger work was this entity published?  
       & \{entity\_label\} was published in \{property\_value\}. \\

P413   & position played on team / speciality  
       & position or specialism of a player on a team  
       & What position does this entity play?  
       & \{entity\_label\} plays in the position \{property\_value\}. \\

P1923  & participating team             & teams that participated in an event  
       & Which teams participated in this event?  
       & Teams participating in \{entity\_label\} include \{property\_value\}. \\

P1001  & applies to jurisdiction        & territorial jurisdiction that an institution or law applies to  
       & Under which jurisdiction does this entity operate?  
       & \{entity\_label\} operates under the jurisdiction of \{property\_value\}. \\

   \caption{Culturally relevant Wikidata properties with corresponding template questions and answers used for multilingual VQA generation.}
   \label{tab:properties_list}
\end{longtable}
% switch back to two-column for whatever follows
\twocolumn

\begin{table*}[t]
\centering
\small
% \resizebox{\textwidth}{!}{%
\begin{tabular}{@{}lrrrrrrr@{}}
\toprule
\textbf{Country/Region} & \textbf{Entities} & \textbf{Images} & \textbf{Template QA} & \textbf{Open-Ended} & \textbf{MCQ} & \textbf{Open-Ended$_F$} & \textbf{MCQ$_F$} \\
\midrule
Germany & 332,650 & 350,828 & 2,752,048 & 2,835,679 & 965,541 & 1,506,438 & 426,272 \\
France & 268,298 & 276,983 & 2,676,838 & 2,729,262 & 941,466 & 1,435,627 & 528,449 \\
United Kingdom & 175,486 & 328,906 & 1,355,577 & 2,183,466 & 891,282 & 1,319,135 & 469,302 \\
Italy & 128,821 & 222,351 & 1,133,463 & 1,763,658 & 745,977 & 1,323,626 & 653,884 \\
Spain & 124,280 & 216,019 & 985,241 & 1,519,295 & 616,304 & 906,943 & 545,056 \\
Japan & 82,690 & 145,843 & 793,759 & 1,214,762 & 483,233 & 799,963 & 431,739 \\
Czechia & 110,384 & 198,223 & 636,978 & 994,864 & 401,437 & 679,115 & 380,160 \\
Poland & 98,577 & 131,155 & 753,750 & 936,799 & 361,028 & 529,669 & 328,143 \\
Russia & 119,158 & 180,253 & 613,822 & 848,540 & 343,834 & 628,558 & 311,416 \\
India & 29,574 & 72,683 & 365,804 & 717,067 & 218,854 & 542,516 & 270,301 \\
Brazil & 38,575 & 68,775 & 419,684 & 648,164 & 257,966 & 479,162 & 236,749 \\
Ukraine & 57,665 & 100,367 & 367,819 & 562,770 & 224,044 & 421,096 & 207,434 \\
China & 38,435 & 68,858 & 288,524 & 468,916 & 200,950 & 365,277 & 187,660 \\
Norway & 27,632 & 47,615 & 255,226 & 382,264 & 146,757 & 273,697 & 118,463 \\
Netherlands & 72,709 & 72,709 & 375,078 & 375,020 & 119,563 & 225,651 & 114,602 \\
Mexico & 12,224 & 29,724 & 184,998 & 370,152 & 113,682 & 271,408 & 122,758 \\
Israel & 19,689 & 33,731 & 183,099 & 289,430 & 124,912 & 233,556 & 105,840 \\
Romania & 15,408 & 26,451 & 196,705 & 287,122 & 109,326 & 194,952 & 104,126 \\
Indonesia & 9,026 & 22,060 & 145,832 & 256,309 & 66,731 & 148,594 & 79,859 \\
Turkey & 13,610 & 23,876 & 163,963 & 256,350 & 107,366 & 183,648 & 99,250 \\
Iran & 12,930 & 32,496 & 114,996 & 252,235 & 80,307 & 194,867 & 103,478 \\
Greece & 9,975 & 24,887 & 125,163 & 250,048 & 76,779 & 172,912 & 95,873 \\
Portugal & 19,733 & 35,229 & 155,542 & 237,166 & 94,069 & 162,184 & 93,708 \\
South Korea & 8,809 & 15,175 & 149,796 & 209,911 & 71,649 & 123,550 & 65,233 \\
Ireland & 9,115 & 22,856 & 86,838 & 185,033 & 58,225 & 146,654 & 72,337 \\
Bulgaria & 7,167 & 17,315 & 94,452 & 177,989 & 54,002 & 129,713 & 64,048 \\
Taiwan & 12,644 & 33,410 & 71,483 & 166,306 & 54,930 & 142,712 & 70,085 \\
Egypt & 3,920 & 9,596 & 63,237 & 136,891 & 43,655 & 104,816 & 48,698 \\
Thailand & 5,837 & 15,037 & 58,397 & 125,292 & 39,345 & 101,078 & 49,959 \\
Pakistan & 2,851 & 6,973 & 38,005 & 76,927 & 24,085 & 59,778 & 29,507 \\
Malaysia & 3,858 & 9,788 & 38,208 & 79,684 & 24,666 & 63,484 & 31,065 \\
Nigeria & 2,519 & 6,368 & 42,080 & 77,164 & 21,339 & 53,213 & 25,948 \\
Bangladesh & 3,659 & 9,236 & 29,253 & 62,700 & 20,382 & 51,071 & 25,715 \\
Vietnam & 3,230 & 5,744 & 37,035 & 58,513 & 24,297 & 43,626 & 21,855 \\
Singapore & 1,752 & 4,298 & 23,619 & 54,281 & 17,059 & 41,825 & 19,176 \\
Saudi Arabia & 948 & 2,292 & 17,759 & 35,046 & 10,772 & 26,547 & 13,087 \\
Kenya & 1,120 & 2,763 & 17,251 & 36,337 & 11,412 & 29,164 & 14,657 \\
Ethiopia & 880 & 2,163 & 14,244 & 29,976 & 9,551 & 23,713 & 10,955 \\
Sri Lanka & 1,066 & 2,651 & 14,643 & 29,484 & 8,861 & 22,177 & 10,913 \\
Tanzania & 592 & 1,454 & 11,966 & 26,332 & 8,451 & 17,689 & 11,589 \\
Mongolia & 542 & 1,306 & 12,482 & 23,604 & 6,900 & 16,429 & 8,765 \\
Rwanda & 572 & 1,393 & 7,332 & 15,693 & 5,157 & 11,850 & 5,821 \\
\midrule
\textbf{Total} & \textbf{1,888,610} & \textbf{2,879,840} & \textbf{15,871,989} & \textbf{21,986,501} & \textbf{8,206,146} & \textbf{14,207,683} & \textbf{6,613,935} \\
\bottomrule
\end{tabular}%
% }
\caption{Dataset statistics by country/region. The dataset contains culturally significant entities from Wikidata with 1-3 images per entity and questions generated from 76 cultural properties. Unique entities shown are those with images, text-only entities excluded. F in last two columns means filtered data.}
\label{tab:dataset_stats}
\end{table*}
\begin{table*}[t]
\centering
\small % Use a slightly smaller font for better fit and consistency
\begin{tabular}{@{}lrrrr@{}}
\toprule
\textbf{Language} & \textbf{Open-Ended} & \textbf{MCQ} & \textbf{Open-Ended$_F$} & \textbf{MCQ$_F$} \\
\midrule
\textbf{en} (English) & 3,778,963 & 1,369,758 & 2,501,144 & 1,152,830 \\
\textbf{fr} (French) & 1,822,466 & 668,153 & 1,181,935 & 530,004 \\
\textbf{de} (German) & 1,782,256 & 626,116 & 1,083,314 & 469,522 \\
\textbf{nl} (Dutch) & 1,648,445 & 602,869 & 1,053,835 & 487,091 \\
\textbf{es} (Spanish) & 1,415,511 & 508,136 & 878,913 & 412,530 \\
\textbf{it} (Italian) & 1,114,458 & 430,928 & 745,316 & 347,233 \\
\textbf{ga} (Irish) & 964,614 & 357,266 & 615,712 & 282,814 \\
\textbf{pl} (Polish) & 818,624 & 312,878 & 511,913 & 245,297 \\
\textbf{ru} (Russian) & 849,610 & 336,357 & 553,662 & 277,540 \\
\textbf{pt} (Portuguese) & 872,402 & 324,938 & 542,464 & 244,671 \\
\textbf{cs} (Czech) & 781,353 & 285,846 & 480,799 & 233,627 \\
\textbf{ja} (Japanese) & 685,032 & 267,259 & 441,822 & 215,680 \\
\textbf{zh} (Chinese) & 728,825 & 286,369 & 491,016 & 236,206 \\
\textbf{tr} (Turkish) & 640,652 & 246,485 & 415,126 & 194,963 \\
\textbf{uk} (Ukrainian) & 526,988 & 208,179 & 346,493 & 172,357 \\
\textbf{ro} (Romanian) & 366,781 & 141,767 & 242,138 & 105,055 \\
\textbf{fa} (Persian) & 362,570 & 145,847 & 241,236 & 115,227 \\
\textbf{id} (Indonesian) & 347,249 & 130,057 & 223,098 & 100,871 \\
\textbf{ar} (Arabic) & 346,263 & 134,798 & 229,576 & 110,000 \\
\textbf{vi} (Vietnamese) & 298,369 & 118,273 & 199,562 & 87,990 \\
\textbf{ko} (Korean) & 256,574 & 104,499 & 172,769 & 84,691 \\
\textbf{he} (Hebrew) & 221,549 & 91,434 & 150,173 & 71,602 \\
\textbf{ms} (Malay) & 243,026 & 93,024 & 161,397 & 69,462 \\
\textbf{el} (Greek) & 166,436 & 64,092 & 102,493 & 50,733 \\
\textbf{bg} (Bulgarian) & 139,184 & 55,340 & 92,766 & 45,780 \\
\textbf{bn} (Bengali) & 137,984 & 48,763 & 95,023 & 46,212 \\
\textbf{ur} (Urdu) & 97,025 & 37,085 & 65,567 & 33,166 \\
\textbf{hi} (Hindi) & 77,997 & 27,260 & 57,202 & 29,295 \\
\textbf{sw} (Swahili) & 128,935 & 46,746 & 77,641 & 32,988 \\
\textbf{ta} (Tamil) & 75,908 & 27,264 & 53,259 & 26,670 \\
\textbf{th} (Thai) & 85,927 & 33,369 & 58,433 & 30,558 \\
\textbf{te} (Telugu) & 55,477 & 20,068 & 38,837 & 20,015 \\
\textbf{jv} (Javanese) & 58,164 & 21,218 & 39,747 & 19,933 \\
\textbf{su} (Sundanese) & 30,857 & 10,840 & 21,238 & 10,583 \\
\textbf{ig} (Igbo) & 23,854 & 8,278 & 16,154 & 7,729 \\
\textbf{si} (Sinhala) & 16,828 & 6,687 & 12,407 & 6,306 \\
\textbf{mn} (Mongolian) & 13,495 & 5,605 & 9,650 & 4,682 \\
\textbf{am} (Amharic) & 3,975 & 1,627 & 2,704 & 1,483 \\
\textbf{no} (Norwegian) & 1,875 & 668 & 1,149 & 539 \\
\midrule
\textbf{TOTAL} & \textbf{21,986,501} & \textbf{8,206,146} & \textbf{14,207,683} & \textbf{6,613,935} \\
\bottomrule
\end{tabular}
\caption{Data distribution statistics by language, showing unfiltered and filtered counts for Open-Ended and Multiple-Choice Question (MCQ) items.}
\label{tab:lang_data_stats}
\end{table*}

\end{document}